\documentclass[10pt,journal,compsoc,twoside]{IEEEtran}

\ifCLASSOPTIONcompsoc
\usepackage[nocompress]{cite}
\else
\usepackage{cite}
\fi
\ifCLASSINFOpdf
\else
\fi
\usepackage{times}
\usepackage{epsfig}
\usepackage{graphicx}
\usepackage{amsmath}
\usepackage{amssymb}
\usepackage{algorithm,algpseudocode}

\graphicspath{{Figures/}}

\usepackage{makecell}

\usepackage{amsmath,amssymb,amsfonts} 
\usepackage{subcaption}
\usepackage{fixltx2e}
\usepackage{multirow}
\usepackage{booktabs}
\usepackage{amsmath}

\usepackage{algorithm,algpseudocode}

\algrenewcommand\algorithmicforall{\textbf{foreach}}

\algdef{SE}[SUBALG]{SubAlgorithm}{EndSubAlgorithm}{\algorithmicsubalgorithm}{\algorithmicend\ \algorithmicsubalgorithm}%

\algtext*{SubAlgorithm}
\algtext*{EndSubAlgorithm}

\algdef{SE}[SUBALG]{Indent}{EndIndent}{}{\algorithmicend\ }%
\algtext*{Indent}
\algtext*{EndIndent}

\usepackage{threeparttable}

\usepackage{arydshln}

\usepackage{amsmath,amsfonts,bm}

\def\eqref#1{equation~\ref{#1}}

\def\1{\bm{1}}

\DeclareMathAlphabet{\mathsfit}{\encodingdefault}{\sfdefault}{m}{sl}
\SetMathAlphabet{\mathsfit}{bold}{\encodingdefault}{\sfdefault}{bx}{n}

\def\emB{{B}}

\DeclareMathOperator*{\argmax}{arg\,max}

\newcommand{\specialcell}[2][c]{%
	\begin{tabular}[#1]{@{}c@{}}#2\end{tabular}}
\usepackage[T1]{fontenc}
\usepackage{wrapfig}
\usepackage{etoolbox}

\usepackage{xcolor}
\usepackage{soul}

\definecolor{mycolor}{rgb}{0.75, 0.75, 0.75}
\definecolor{lavendergray}{rgb}{0.77, 0.76, 0.82}
\definecolor{lightgray}{rgb}{0.83, 0.83, 0.83}
\newcommand{\highlightbox}[1]{\colorbox{mycolor}{$#1$}}

\usepackage{colortbl}

\definecolor{Gray}{gray}{0.38}
\definecolor{LightCyan}{rgb}{0.48,0.78,1}
\definecolor{LightColor1}{rgb}{0.68,0.88,1.2}

\definecolor{LightGray}{gray}{.8}
\newcolumntype{C}[1]{>{\centering\arraybackslash}p{#1}}

\newcolumntype{y}{>{\columncolor{yellow}}c}
\newcolumntype{a}{>{\columncolor{lightgray}}c}
\newcolumntype{g}{>{\columncolor{pink}}c}
\newcolumntype{o}{>{\columncolor{LightColor1}}c}
\newcolumntype{u}{>{\columncolor{LightCyan}}c}

\usepackage{threeparttable}

\usepackage[numbers,sort&compress]{natbib}

\usepackage[fleqn]{nccmath}
\usepackage{framed}
\newenvironment{mathframed}{\framed%
	\allowdisplaybreaks
	\vspace*{-\abovedisplayskip}\noindent}{%
	\vspace*{-\dimexpr\baselineskip+\topsep}\endframed}

\newcounter{eqnthing}

\usepackage{scalerel}
\usepackage{tikz}
\usetikzlibrary{svg.path}
\definecolor{orcidlogocol}{HTML}{A6CE39}
\tikzset{
	orcidlogo/.pic={
		\fill[orcidlogocol] svg{M256,128c0,70.7-57.3,128-128,128C57.3,256,0,198.7,0,128C0,57.3,57.3,0,128,0C198.7,0,256,57.3,256,128z};
		\fill[white] svg{M86.3,186.2H70.9V79.1h15.4v48.4V186.2z}
		svg{M108.9,79.1h41.6c39.6,0,57,28.3,57,53.6c0,27.5-21.5,53.6-56.8,53.6h-41.8V79.1z M124.3,172.4h24.5c34.9,0,42.9-26.5,42.9-39.7c0-21.5-13.7-39.7-43.7-39.7h-23.7V172.4z}
		svg{M88.7,56.8c0,5.5-4.5,10.1-10.1,10.1c-5.6,0-10.1-4.6-10.1-10.1c0-5.6,4.5-10.1,10.1-10.1C84.2,46.7,88.7,51.3,88.7,56.8z};
	}
}

\newcommand\orcidicon[1]{\href{https://orcid.org/#1}{\mbox{\scalerel*{
				\begin{tikzpicture}[yscale=-1,transform shape]
					\pic{orcidlogo};
				\end{tikzpicture}
			}{|}}}}

\usepackage{stmaryrd}

\usepackage{enumitem}
\setlist[itemize,1]{label=$\bullet$}
\setlist[itemize,2]{label=$\checkmark$}

\usepackage[pagebackref=true,breaklinks=true,letterpaper=true,colorlinks,bookmarks=false]{hyperref}
\usepackage{url}
\hypersetup{
	colorlinks=true,
}

\begin{document}
	
	\onecolumn  
	\thispagestyle{empty}
	\noindent
	\begin{center}
		\large
		Summary of changes
	\end{center}
	
	~\\
	A preliminary version, entitled “ProSelfLC: Progressive Self Label Correction for Training Robust Deep Neural Networks”, was published in the IEEE Conference on Computer Vision and Pattern Recognition (CVPR) 2021. Compared with our CVPR 2021 version, {this manuscript is a substantial revision. There is more than 50\% improvement and extension}.
	
	~\\
	Concretely, the new extension and improvements are listed as follows:
	\begin{enumerate}
		
		\item There exist three well-accepted findings: (a) deep models easily fit random noise [92]; (b)
		deep networks learn simple semantic patterns before fitting noise
		[4]; (c) modern deep neural works tend to be over-confident [20,
		52]. \\
		In this paper, we disclose \textbf{a new insightful one}, which complements them:
		\begin{center}
			\textit{Deep neural networks become less confident of learning
				semantic patterns before fitting noise when the label noise rises.}
		\end{center}  
		
		\item To illustrate the new finding \textbf{mathematically and visually} in Section 4: 
		\begin{itemize}[label={\checkmark}]
			\item We \textbf{add a technical subsection 4.1 about calibration error}, where the Generic coarse Signed Calibration Error (GSCE) is proposed. 
			\item We \textbf{add an empirical analysis subsection 4.2 to visualize} the miscalibration.  
		\end{itemize}
		
		\item \textbf{A new technical proposal},  inspired by the new finding and miscalibration analysis, is introduced to decrease the entropy of self knowledge. Concretely, we propose to use an Annealed Temperature and learn towards a revised low-temperature entropy state.~\\

		\item Other new technical details are added, including: 
		\begin{itemize}[label={\checkmark}]
			\item In the beginning of Section 2, we clarify \textbf{the definition of knowledge confidence, and two confidence metrics}. 
			
			\item In Section 2.3, we \textbf{add a new Proposition 2 to discuss the equivalence property of cross entropy, KL divergence, and relative entropy} when a target probability distribution is fixed in training.  
			\item In Section 6.3, we \textbf{discuss sample selection using the small-loss criterion}, to give a clearer explanation of ProSelfLC's superiority.  
		\end{itemize}

		\item Much more experiments are added with an updated state-of-the-art:
		\begin{itemize}[label={\checkmark}]
			\item \textbf{One more vision dataset} (Food-101N), which is commonly used to test a method’s effectiveness for training robust models against real-world label noise.
			\item \textbf{A very diverse data domain, i.e., a protein dataset}, by which we demonstrate the general applicability of our method. 
			\item \textbf{An updated state-of-the-art results} on two real-world noisy datasets Clothing1M and Food-101N under a fair comparison with the relevant methods from ICLR 2021, ICML 2021, CVPR 2021, ICLR 2022, CVPR 2022. Some of them are either concurrent with our CVPR 2021 publication or publicly available after our first arXiv version (May, 2020).
			\item \textbf{More thorough experimental analysis on key factors (E.g., with/without AT, Hyper-parameters space of $B$ and $T$, Self trust scheme variants)} is presented. 
		\end{itemize}

		\item For easily reproducing our results and promoting future research, we will \textbf{release and maintain our source code} at \url{https://github.com/XinshaoAmos
			Wang/ProSelfLC-AT}.
	\end{enumerate}
	
	\twocolumn
	\setcounter{page}{1}

	%
	\title{
		ProSelfLC: Progressive Self Label Correction \\ 
		Towards A Low-Temperature Entropy State  
	}
	%
	%
	%
	%
	
	\author{Xinshao~Wang$^{1,2~\textsuperscript{\orcidicon{0000-0001-8907-8258}}}$,
		~Yang~Hua$^{3~\textsuperscript{\orcidicon{0000-0001-5536-503X}}}$, 
		~Elyor~Kodirov$^{1}$,
		~Sankha Subhra Mukherjee$^{1}$, 
		~David~A.~Clifton$^{2}$,
		~and~Neil~M.~Robertson$^{1}$
		\IEEEcompsocitemizethanks{
			\IEEEcompsocthanksitem 
			$^1$Zenith Ai, UK.
			$^2$Institute of Biomedical Engineering, Department of Engineering Science, University of Oxford, UK.
			$^3$Queen's University Belfast, UK.
			\IEEEcompsocthanksitem 
			Emails: \{xinshao, elyor, rick, neil\}@zenithai.co.uk, \{xinshao.wang,
			david.clifton\}@eng.ox.ac.uk,
			\{y.hua\}@qub.ac.uk. 
			\IEEEcompsocthanksitem
			Prof. David A. Clifton was supported by the National Institute for Health Research (NIHR) Oxford Biomedical Research Centre (BRC).
			
			\IEEEcompsocthanksitem 
			Corresponding authors: Xinshao Wang, Neil~M.~Robertson. 
		}
		\\
	}

	%
	%

	\markboth{
		TPAMI Submission
	}{~}
	%



	\IEEEtitleabstractindextext{%
		\begin{abstract}
			


			
			There is a family of label modification approaches including self and non-self label correction (LC), and output regularisation.  
			They are widely used for training robust deep neural networks (DNNs), but have not been mathematically and thoroughly analysed together.  
			%
			We study them and discover three key issues:  
			(1) We are more interested in adopting Self LC as it leverages its own knowledge and requires no auxiliary models.
			However, it is unclear how to adaptively trust a learner as the training proceeds.     
			(2) Some methods penalise while the others reward low-entropy (i.e., high-confidence) predictions, prompting us to ask which one is better.
			(3) Using the standard training setting, a learned model becomes less confident when severe noise exists. The model is uncalibrated with a conflict between knowledge accuracy and confidence. 
			Self LC using high-entropy knowledge would generate high-entropy targets.      
			%
			
			~\\
			To resolve the issue (1), inspired by a well-accepted finding, i.e., deep neural networks learn meaningful patterns before fitting noise, we propose a novel end-to-end method named ProSelfLC, which is designed according to the learning time and prediction entropy. 
			Concretely, for any data point, we progressively and adaptively trust its predicted probability distribution versus its annotated one if a network has been trained for a relatively long time and the prediction is of low entropy.
			%
			%
			For the issue (2), the effectiveness of ProSelfLC defends entropy minimisation. By ProSelfLC, we empirically prove that it is more effective to redefine a semantic low-entropy state and optimise the learner toward it.   
			To address the issue (3), we decrease the entropy of self knowledge using a low temperature before exploiting it to correct labels, so that the revised labels redefine low-entropy target probability distributions. 
			%
			%
			
			~\\
			We demonstrate the effectiveness of ProSelfLC through extensive experiments in both clean and noisy settings, and on both image and protein datasets.   
			Furthermore, our source code is available at 
			\url{https://github.com/XinshaoAmosWang/ProSelfLC-AT}. 
		\end{abstract}

	}

	\maketitle

	\IEEEdisplaynontitleabstractindextext

	%
	\IEEEpeerreviewmaketitle

	\ifCLASSOPTIONcompsoc
	\IEEEraisesectionheading{\section{Introduction}\label{introduction}}
	\else
	\section{Introduction}
	\label{introduction}
	\fi

	
	%
	%

\begin{figure*}[!t]
		\centering
		%
		%
		\begin{subfigure}[h!]{\textwidth}
			\centering
			\includegraphics[clip, trim=0cm 2.8cm 0.2cm 1.42cm, width=0.85\textwidth]{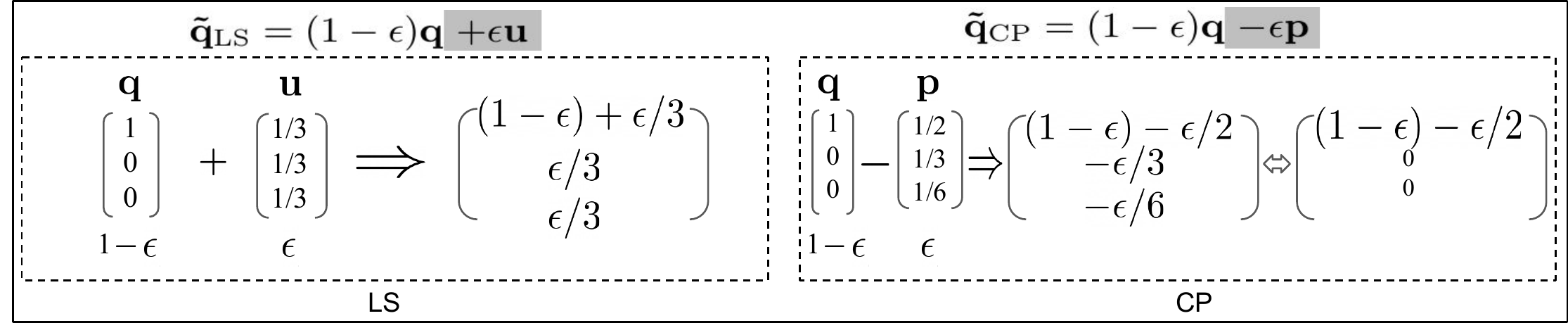}
			\vspace{-0.06cm}
			\caption{OR includes LS \cite{szegedy2016rethinking}
				and CP  
				\cite{pereyra2017regularizing}. 
				LS softens a target by adding a uniform label distribution. 
				CP changes the probability 1 to a smaller value $1-\epsilon$ in the one-hot target. 
				The double-ended arrow means factual equivalence, because an output is definitely non-negative after a softmax layer. 
			}
			\label{fig:LS_CP}
		\end{subfigure}
		
		\vspace{-0.1cm}
		\begin{subfigure}[h!]{\textwidth}
			\centering
			\includegraphics[clip, trim=-0.2cm 3.2cm 0.3cm 0.8cm, width=0.85\textwidth]{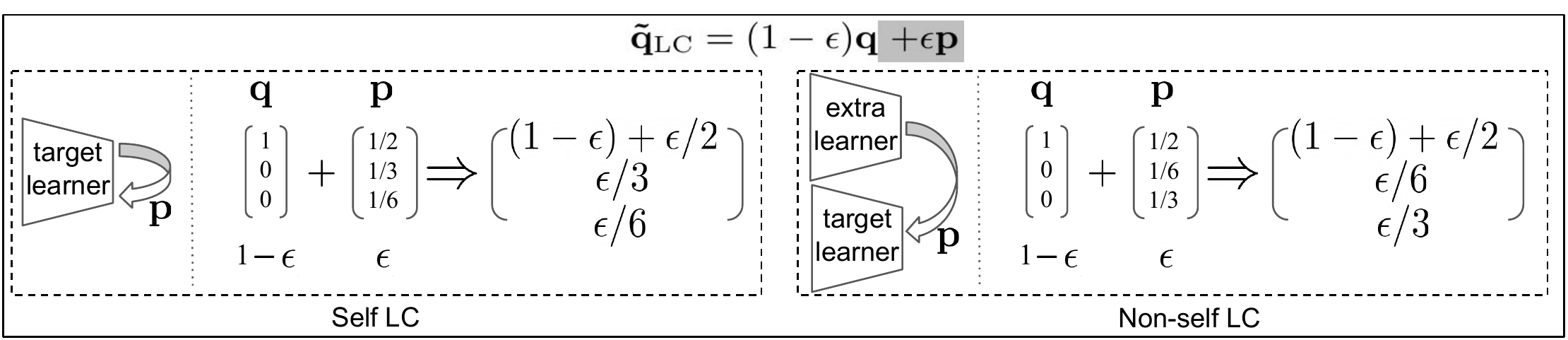}
			\vspace{-0.06cm}
			\caption{LC contains Self LC \cite{lee2013pseudo,reed2015training,tanaka2018joint,yuan2020revisiting}  and Non-self LC \cite{hinton2015distilling}.
				%
				%
				The parameter $\epsilon$ defines how much a predicted label distribution is trusted. 
			}
			\label{fig:LC}
		\end{subfigure}
		%
		\caption{Target modification includes OR (LS and CP), and LC (Self LC and Non-self LC).  
			Assume there are three training classes. 
			$\mathbf{q}$ is the one-hot target. 
			$\mathbf{u}$ is a uniform probability distribution and 
			$\mathbf{p}$ denotes a predicted one. 
			$\epsilon \in [0,1]$ is the  coefficient. 
		}
		\label{fig:illustration_LS_CP_LC}
		\vspace{-0.38cm}
	\end{figure*}

	\IEEEPARstart{T}{he} label
	modification is a supervision improvement strategy for model optimisation.
	It redefines the target probability distribution of a data point by combining a one-hot distribution, which is the target if no label modification, and another one, which could be either predicted or predefined. 
	The existing target (label) modification algorithms can be roughly categorized into two types: (1) 
	Output regularisation (OR), including label smoothing (LS) \cite{szegedy2016rethinking,muller2019does} and confidence penalty (CP) \cite{pereyra2017regularizing}. 
	OR penalises over-confident predictions to regularise deep neural networks;
	(2) Label correction (LC). 
	LC can not only \textit{correct the semantic classes} of noisy probability distributions, but also regularise the trained models by adding the similarity structure information over training classes to one-hot probability distributions so that the learning targets are aware of the similarity hierarchy over training data.
	%
	%
	%
	%
	%
	%
	%
	%
	%
	LC can be finely categorized into two subclasses: Non-self LC and Self LC. The former requires extra learners, hence the name-``Non-self''. 
	Accordingly, Self LC represents that a model bootstraps itself during training. 
	A widely-adopted representative of Non-self LC is knowledge distillation (KD). KD supervises a model using the predictions of other model(s), usually named teacher(s) \cite{hinton2015distilling}. 
	Self LC methods contain Pseudo-Label \cite{lee2013pseudo}, bootstrapping (Boot-hard and Boot-soft) \cite{reed2015training}, Joint Optimisation (Joint-hard and Joint-soft) \cite{tanaka2018joint}, and Tf-KD$_{self}$ \cite{yuan2020revisiting}, etc. 
	%
	%
	We display an overview in Fig.~\ref{fig:illustration_LS_CP_LC} with detailed mathematical analysis in the Section~\ref{section:preliminary} and Table~\ref{table:summary_CE_LS_CP_LC}. 
	
	%
	%
	%
	%

	Firstly, we are interested in adopting Self LC in practice for three reasons: 
	(1) OR methods naively penalise confident outputs without leveraging easily accessible knowledge from other learner(s) or itself (Fig.~\ref{fig:LS_CP}); 
	(2) Non-self LC requires auxiliary models to generate accurate predictions (Fig.~\ref{fig:LC}). 
	(3) Self LC leverages a model's self knowledge and does not need extra learner(s). 
	But we note a core question which is not well answered:
	\begin{center}
		\textit{How much should we trust a learner to leverage its knowledge to revise labels as training proceeds?}
	\end{center}
	\noindent
	As illustrated in Fig.~\ref{fig:LC}, in Self LC, we have two labels for any data point--a predefined one-hot $\mathbf{q}$ and a predicted $\mathbf{p}$ (a.k.a., self knowledge). The learning target is redefined to be $(1-\epsilon)\mathbf{q}+\epsilon \mathbf{p}$, where $\epsilon$ defines the trust score of $\mathbf{p}$. 
	%
	%
	In existing methods, $\epsilon$ is fixed without considering the fact that a model's knowledge could improve as the training proceeds.
	Taking the bootstrapping \cite{reed2015training} as an example, 
	$\epsilon$
	is fixed throughout the training.  
	%
	While Joint Optimisation 
	stage-wisely trains a model. Concretely, it fully trusts predicted probability distributions when a stage ends and uses them as the targets of next stage, mathematically, $\epsilon=1$. 
	Tf-KD$_{self}$ trains a model by two stages: $\epsilon=0$ in the first one while $\epsilon$ is tuned for the second one. 
	We remark that the stage-wise training requires significant human intervention about the duration of each stage and tuning $\epsilon$ for the next stage, etc, thus being time-consuming in practice. 
	
	
	
	

	

	To improve Self LC, 
	we propose a novel method named Progressive Self Label Correction (ProSelfLC), which is end-to-end trainable and needs negligible extra cost. 
	Most importantly, ProSelfLC modifies the target progressively and adaptively as training goes. 
	{Two design inspirations of ProSelfLC are}:  (1) If a model learns from scratch, its predictions are unreliable in the early phase, so that human annotations have to be relied on for supervision even though they could be noisy; 
	(2) As time progresses, the model learns semantically meaningful patterns before fitting noise, even when severe label noise exists \cite{arpit2017closer}. 
	Therefore, we can leverage a model's accurate and confident knowledge to revise pre-annotated labels, then the model will not fit noise.  
	
	%
	

	Secondly, note that OR methods penalise low entropy while LC rewards it, intuitively leading to the second vital question:  
	\begin{center}
		\textit{Should we penalise a low-entropy status or reward it?}
	\end{center}
	\noindent
	Entropy minimisation is the most widely used principle in unsupervised and semi-supervised machine learning scenarios \cite{hartigan1979algorithm,rumelhart1986learning,grandvalet2005semi,grandvalet2006entropy,lecun2015deep}. In standard supervised classification, minimising categorical cross entropy (CCE) also optimises a model towards a minimum-entropy state defined by one-hot labels. 
	However, when it comes to large-scale machine learning where noisy data generally exists, confidence penalty becomes popular recently for reducing noisy fitting. 
	In contrast, we prove that it is more effective to reward a semantically meaningful low-entropy state redefined by ProSelfLC. 
	By showing the effectiveness of ProSelfLC, we defend entropy minimisation against the recent confidence penalty practices \cite{szegedy2016rethinking,muller2019does,pereyra2017regularizing,dubey2018maximum}.

	
	Thirdly, we disclose a common phenomenon which hinders a model from confident learning towards a correct low-entropy target state. 
	By reporting the confidence metrics in Fig.~\ref{fig:cce_accuracy_confidence_metrics} and  Fig.~\ref{fig:cce_dynamics_confidence_accuracy_resnet18}, we reveal this phenomenon: using the standard CCE loss, \textit{when the training data contains severe noise, before fitting noise, a deep model has much lower confidence than its accuracy}. 
	For LC, if the predictions are high-entropy, the modified targets will be high-entropy too. 
	Therefore, we develop an Annealed Temperature (AT) as a plug-in module to reduce the entropy of self knowledge.
	Empirically (see Table~\ref{table:without_or_with_AT_confidence_calibration}), for the Self LC methods including Boot-soft and ProSelfLC, with an AT plugged in, we are able to exploit the low-temperature (i.e., low-entropy) self knowledge to redefine a corrected low-entropy target state. Consequently, a model can learn confidently and generalise well.   
	

	We summarise our main contributions as follows: 
	\begin{itemize}
		\item We provide a theoretical study on common {target modification methods} through entropy and KL divergence \cite{kullback1951information}.  
		Accordingly, we reveal their drawbacks and propose ProSelfLC as a solution. ProSelfLC can: 
		(1) enhance the similarity structure information over training classes; 
		(2) correct the semantic classes of noisy label distributions. 
		ProSelfLC is the first method to progressively and adaptively trust a low-temperature self knowledge. 
		
		\item  
		We uncover a finding which complements the recent findings \cite{zhang2017understanding,arpit2017closer,guo2017calibration,minderer2021revisiting}: \textit{when a higher label noise exists, deep models are significantly less confident of learning semantically meaningful patterns before fitting noise}. Correspondingly, we propose to decrease the entropy of self knowledge using an AT and learn towards a revised low-temperature entropy state. 
		
		
		\item Our extensive experiments: (1) defend the entropy minimisation principle; (2) demonstrate ProSelfLC's effectiveness in clean and noisy settings of
		two very diverse data domains, i.e., image and protein datasets.
		This demonstrates the general applicability of our method. 
	\end{itemize}

	\section{Mathematical analysis and theory}
	\label{section:preliminary}
	

	\textbf{Notations}. 
	Let $\mathbf{X}=\{(\mathbf{x}^i, y^i)\}_{i=1}^n$ represent $n$ training examples, where $(\mathbf{x}^i, y^i)$ denotes $i-$th sample with input $\mathbf{x}^i \in \mathbb{R}^d$ and label $y^i \in \{1,2, ..., c\}$. $c$ is the number of classes. A deep neural network $\mathrm{z}$ consists of an embedding network $\mathrm{f}(\cdot): \mathbb{R}^d \rightarrow \mathbb{R}^k$ and a linear classifier $\mathrm{g}(\cdot): \mathbb{R}^k \rightarrow \mathbb{R}^c$, i.e., $\mathbf{z}^i=\mathrm{z}(\mathbf{x}^i)=\mathrm{g}(\mathrm{f}(\mathbf{x}^i)): \mathbb{R}^d \rightarrow \mathbb{R}^c$.
	For the brevity of analysis, whenever there is no confusion, we take a data point $(\mathbf{x}^i, y^i)$ and omit its superscript so that it is denoted by $(\mathbf{x}, y)$.  
	The linear classifier is usually the last fully-connected layer. Its output is named logit vector $\mathbf{z} \in \mathbb{R}^c$. 
	We produce its classification probabilities $\mathbf{p}$ by normalising the logits using a softmax function: 
	\begin{equation}
		\begin{aligned}
			\mathbf{p}(j|\mathbf{x})= 
			{\exp (\mathbf{z}_{j})}
			/{ \sum\nolimits_{v=1}^{c} \exp (\mathbf{z}_{v}) },
		\end{aligned}
		\label{chapter:DM_eq:softmax_normalisation}
	\end{equation} 
	where $\mathbf{p}(j|\mathbf{x})$ is the probability of $\mathbf{x}$ belonging to class $j$. 
	Its corresponding ground-truth is usually denoted by a one-hot representation $\mathbf{q}$: $\mathbf{q}(j|\mathbf{x})=1$, if $j=y$; $\mathbf{q}(j|\mathbf{x})=0$, otherwise.
	Our definition of knowledge confidence is: 
	
	\noindent
	\textbf{Definition 1} (\textit{Knowledge Confidence}). A model's knowledge with respect to $\mathbf{x}$ is defined by $\mathbf{p}$. The knowledge confidence measures how certain $\mathbf{p}$ is, and mathematically defined by how distant $\mathbf{p}$ is from a uniform distribution $\mathbf{u} \in \mathbb{R}^c, \text{ and } \forall j,  \mathbf{u}_j = \frac{1}{c}$.
	We can calculate the knowledge confidence using two formulations: 
	\begin{equation}
		\mathrm{conf}(\mathbf{p}) =  
		\begin{cases} 
			 \mathrm{conf}_{\mathrm{top}}(\mathbf{p}) = \max_j \mathbf{p}(j|\mathbf{x});\\
			
			\mathrm{conf}_{\mathrm{all}}(\mathbf{p}) = 1 - 
			{\mathrm{H}(\mathbf{p})}/{\mathrm{H}(\mathbf{u})},
		\end{cases}
	\end{equation}
	where $\mathrm{conf}_{\mathrm{top}}(\mathbf{p})$ \textit{is widely adopted in} \cite{guo2017calibration,kumar2019verified,minderer2021revisiting} to measure the miscalibration degree between confidence and accuracy,  
	while $\mathrm{conf}_{\mathrm{all}}(\mathbf{p})$ \textit{is our proposed confidence metric}. 
	"top" indicates only the top probability is used while "all" denotes all probabilities are considered. 
	Both metrics are agnostic to the semantic class and accuracy.   
	Generally, they have a strong positive correlation, thus being interchangeable in practice.

	\vspace{-0.10cm}
	\subsection{Semantic class and similarity structure in $\mathbf{p}$}
	\vspace{-0.10cm}
	
	$\mathbf{q} \in \mathbb{R}^c$ provides 
	semantic information about the probabilities of $\mathbf{x}$ being different training classes.
	We could also interpret $\mathbf{q}(j|\mathbf{x})$ as the similarity between $\mathbf{x}$ and $j$-th class. 
	Consequently, $\mathbf{q}$ should not be exactly one-hot, and is proposed to be corrected at training, so that it can define a more structured target probability distribution. 
	%
	For better clarity, we present two definitions: 
	
	\noindent
	\textbf{Definition 2} (\textit{Semantic Class}). Given a target label distribution $\mathbf{\tilde{q}}(\mathbf{x}) \in \mathbb{R}^c$,  the semantic class is defined by $\argmax\nolimits_j {\mathbf{{\tilde{q}}}(j|\mathbf{x})}$, i.e., the class whose probability is the largest.  

	\noindent
	\textbf{Definition 3} (\textit{Similarity Structure}). 
	In $\mathbf{\tilde{q}}(\mathbf{x})$, $\mathbf{x}$ has $c$ probabilities of being predicted to $c$ classes. The similarity structure of $\mathbf{x}$ versus $c$ classes is defined by these probabilities and their differences. 

	\vspace{-0.10cm}
	\subsection{Revisit CCE, LS, CP and LC}
	\vspace{-0.10cm}

	\noindent
	\textbf{Standard CCE}.
	For any input $(\mathbf{x}, y)$, the minimisation objective of standard CCE is: 
	\vspace{-0.12cm}
	\begin{equation}
		\label{eq:cross_entropy}
		\vspace{-0.12cm}
		\begin{aligned}
			\mathrm{L}_{\mathrm{CCE}} ( \mathbf{q}, \mathbf{p} )  =
			\mathrm{H}(\mathbf{q}, \mathbf{p}) 
			&= \mathrm{E}_\mathbf{q} ( -\log~\mathbf{p} ), 
		\end{aligned}
	\end{equation}
	where $\mathrm{H(\cdot, \cdot)}$ represents the cross entropy. $\mathrm{E}_\mathbf{q} ( \mathbf{-\log~\mathbf{p}} ) $ denotes the expectation of negative log-likelihood, and $\mathbf{q}$ serves as the probability mass function.

	
	\noindent
	\textbf{Label smoothing}.
	In LS \cite{szegedy2016rethinking,hinton2015distilling}, 
	we soften one-hot targets by adding $\mathbf{u}$:
	$\mathbf{\tilde{q}_{\mathrm{LS}}} = (1-\epsilon)\mathbf{q}+\epsilon \mathbf{u}$.
	As a result, 
	\vspace{-0.12cm}
	\begin{equation}
		\label{eq:label_smoothing}
		\vspace{-0.12cm}
		\begin{aligned}
			\mathrm{L}_\mathrm{{CCE+LS}}(\mathbf{{q}}, \mathbf{p}; \epsilon) 
			&=
			\mathrm{H}(\mathbf{\tilde{q}_{\mathrm{LS}}}, \mathbf{p}) 
			= \mathrm{E}_\mathbf{\tilde{q}_{\mathrm{LS}}} ( -\log~\mathbf{p} ) 
			\\&
			= (1-\epsilon) \mathrm{H}(\mathbf{q}, \mathbf{p})
			{+
				\epsilon \mathrm{H}(\mathbf{u}, \mathbf{p}).}
		\end{aligned}
	\end{equation}

	\noindent
	\textbf{Confidence penalty}.
	CP \cite{pereyra2017regularizing} penalises highly confident predictions:  
	\vspace{-0.12cm}
	\begin{equation}
		\label{eq:confidence_penalty}
		\vspace{-0.12cm}
		\begin{aligned}
			\mathrm{L}_\mathrm{{CCE+CP}}(\mathbf{{q}}, \mathbf{p}; \epsilon) 
			&
			= (1-\epsilon) \mathrm{H}(\mathbf{q}, \mathbf{p})
			{-
				\epsilon \mathrm{H}(\mathbf{p}, \mathbf{p})}. 
		\end{aligned}
	\end{equation}
	%
	%
	%
	%
	%
	%
	%
	%
	
	\noindent
	\textbf{Label correction}.
	As illustrated in Fig.~\ref{fig:illustration_LS_CP_LC}, LC is a family of algorithms, where the one-hot $\mathbf{q}$ is modified to a convex combination of itself and a predicted distribution: 
	\vspace{-0.12cm}
	\begin{equation}
		\label{eq:label_correction}
		\vspace{-0.12cm}
		\begin{aligned}
			\mathbf{\tilde{q}_{\mathrm{LC}}} = (1-\epsilon)\mathbf{q}+\epsilon \mathbf{p}
			~\Rightarrow~
			&\mathrm{L}_\mathrm{{CCE+LC}}(\mathbf{{q}}, \mathbf{p}; \epsilon) 
			= \mathrm{H}(\mathbf{\tilde{q}_{\mathrm{LC}}}, \mathbf{p})  
			\\&= (1-\epsilon) \mathrm{H}(\mathbf{q}, \mathbf{p})
			{+
				\epsilon \mathrm{H}(\mathbf{p}, \mathbf{p})}.
		\end{aligned}
	\end{equation}
	We remark that if $\epsilon$ is large, and $\mathbf{p}$ is confident in predicting a different class, i.e.,  $\argmax\nolimits_j \mathbf{p}(j|\mathbf{x}) \neq \argmax\nolimits_j \mathbf{q}(j|\mathbf{x})$, $\mathbf{\tilde{q}_{\mathrm{LC}}}$ defines a different semantic class from $\mathbf{q}$.

	\begin{table*}[!t]
		\caption{
			Summary of CCE, LS, CP and LC. 
			In the upper block, we display their learning targets, loss calculations using equal and interchangeable cross entropy and KL divergence. 
			In the bottom block, we present their properties from the viewpoints of entropy minimisation, semantic class and structure.  
		}
		\centering
		\setlength{\tabcolsep}{14.0pt} 
		
		\vspace{-0.22cm}
		\begin{tabular}{l c c c c}
			\toprule
			&  \multicolumn{1}{c}{CCE}
			&  \multicolumn{1}{c}{LS}
			
			& \multicolumn{1}{c}{CP}
			
			& \multicolumn{1}{c}{LC}
			\\
			\midrule
			Target probability distribution
			&  $\mathbf{q}$ & $ \mathbf{\tilde{q}_{\mathrm{LS}}}=(1-\epsilon)\mathbf{q} \highlightbox{+\epsilon \mathbf{u}}$ & $ \mathbf{\tilde{q}_{\mathrm{CP}}}=(1-\epsilon)\mathbf{q}\highlightbox{-\epsilon \mathbf{p}}$& $ \mathbf{\tilde{q}_{\mathrm{LC}}}=(1-\epsilon)\mathbf{q}\highlightbox{+\epsilon \mathbf{p}}$\\
			\\
			Cross entropy & $\mathrm{E}_\mathbf{q} ( -\log~\mathbf{p} )$
			& $\mathrm{E}_\mathbf{\tilde{q}_{\mathrm{LS}}} ( -\log~\mathbf{p} )$ 
			& $\mathrm{E}_\mathbf{\tilde{q}_{\mathrm{CP}}} ( -\log~\mathbf{p} )$
			& $\mathrm{E}_\mathbf{\tilde{q}_{\mathrm{LC}}} ( -\log~\mathbf{p} )$
			\\
			\\
			KL divergence &  $\mathrm{D}_\mathrm{KL}(\mathbf{q}||\mathbf{p})$ & \specialcell{$(1-\epsilon) \mathrm{D}_\mathrm{KL}(\mathbf{q}||\mathbf{p})
				$\\
				$\highlightbox{+\epsilon \mathrm{D}_\mathrm{KL}(\mathbf{u}||\mathbf{p})}$} & 
			\specialcell{$(1-\epsilon) \mathrm{D}_\mathrm{KL}(\mathbf{q}||\mathbf{p})
				$\\
				$\highlightbox{+\epsilon \mathrm{D}_\mathrm{KL}(\mathbf{p}||\mathbf{u})}$} &
			\specialcell{$(1-\epsilon) \mathrm{D}_\mathrm{KL}(\mathbf{q}||\mathbf{p})
				$\\
				$\highlightbox{-\epsilon \mathrm{D}_\mathrm{KL}(\mathbf{p}||\mathbf{u})}$} 
			\\
			
			\vspace{-0.28cm}
			&&&&\\
			\cdashline{1-5}
			\vspace{-0.28cm}
			&&&&\\
			
			Entropy minimisation & -- & \highlightbox{Penalise} over CCE & \highlightbox{Penalise} over CCE & \highlightbox{Reward} over CCE\\

			Semantic class & Annotated & Annotated & Annotated & Annotated and Learned \\
			
			Similarity structure & No & No & No & Yes\\
			
			\bottomrule
		\end{tabular}
		\label{table:summary_CE_LS_CP_LC}
		\vspace{-0.20cm}
	\end{table*}
	
	\vspace{-0.1cm}
	\subsection{Theory on CCE, LS, CP and LC}
	\label{subsec:theory on CCE, LS, CP and LC}
	\vspace{-0.10cm}
	
	\textbf{Proposition 1.} \textit{Compared with the standard CCE, the learning targets are modified in LS, CP and LC.}\\
	\textit{Proof.}  
	$\mathrm{L}_\mathrm{{CCE+CP}}(\mathbf{{q}}, \mathbf{p}; \epsilon) 
	= (1-\epsilon) \mathrm{H}(\mathbf{q}, \mathbf{p})
	{-
		\epsilon \mathrm{H}(\mathbf{p}, \mathbf{p})}=\mathrm{E}_\mathbf{(1-\epsilon)\mathbf{q}{-\epsilon \mathbf{p}}} ( \mathbf{-\log~\mathbf{p}} )$. Therefore, 
	$ \mathbf{\tilde{q}_{\mathrm{CP}}}=(1-\epsilon)\mathbf{q}{-\epsilon \mathbf{p}}$. 
	Additionally,   
	$ \mathbf{\tilde{q}_{\mathrm{LS}}}=(1-\epsilon)\mathbf{q} {+\epsilon \mathbf{u}}$,  
	$\mathbf{\tilde{q}_{\mathrm{LC}}}=(1-\epsilon)\mathbf{q}{+\epsilon \mathbf{p}}$. \hfill\(\Box\)

	\noindent
	\textbf{Proposition 2.} \textit{When a target probability distribution $\mathbf{\tilde{q}}$ is fixed, minimising the cross entropy $\mathrm{H}(\mathbf{\tilde{q}}, \mathbf{p})$ is equivalent to minimising the KL divergence \cite{kullback1951information} of $\mathbf{\tilde{q}}$ from $\mathbf{p}$, i.e., the relative entropy of $\mathbf{\tilde{q}}$ with respect to $\mathbf{p}$. }
	\\
	\textit{Proof.}  
	Let $\mathrm{D}_\mathrm{KL}(\cdot||\cdot)$ denote the KL divergence, we have $\mathrm{H}(\mathbf{\tilde{q}}, \mathbf{p}) = \mathrm{D}_\mathrm{KL}(\mathbf{\tilde{q}} || \mathbf{p}) + \mathrm{H}(\mathbf{\tilde{q}}, \mathbf{\tilde{q}}).$ 
	As $\mathbf{\tilde{q}}$ is fixed, $\mathrm{H}(\mathbf{\tilde{q}}, \mathbf{\tilde{q}})$ is a constant so that we can leave it out of loss minimisation.  
	\hfill\(\Box\)

	\noindent
	\textbf{Proposition 3.} \textit{Some KD methods, which aim to minimise the KL divergence between predictions of a teacher and a student, belong to the family of label correction. }\\
	\textit{Proof.} In general, a loss function of such methods can be defined to be $\mathrm{L}_\mathrm{KD}(\mathbf{q},\mathbf{p}_t, \mathbf{p}) = (1-\epsilon)\mathrm{H}(\mathbf{q}, \mathbf{p}) + \epsilon \mathrm{D}_\mathrm{KL}(\mathbf{p}_t||\mathbf{p})$ \cite{yuan2020revisiting}.  As $\mathrm{D}_\mathrm{KL}(\mathbf{p}_t||\mathbf{p}) = \mathrm{H}(\mathbf{p}_t, \mathbf{p})-\mathrm{H}(\mathbf{p}_t, \mathbf{p}_t)$, $\mathbf{p}_t$ is from a teacher and fixed when training a student. We can omit $\mathrm{H}(\mathbf{p}_t, \mathbf{p}_t)$:
	\vspace{-0.12cm}
	\begin{equation}
		\label{eq:KD_KLptp}
		\vspace{-0.12cm}
		\begin{aligned}
			\mathrm{L}_\mathrm{KD}(\mathbf{q},\mathbf{p}_t, \mathbf{p}) 
			&= (1-\epsilon)\mathrm{H}(\mathbf{q}, \mathbf{p}) + \epsilon \mathrm{H}(\mathbf{p}_t,\mathbf{p})
			\\&= \mathrm{E}_{(1-\epsilon)\mathbf{q}{+\epsilon \mathbf{p}_t}} ( \mathbf{-\log~\mathbf{p}} ) \\&\Rightarrow \mathbf{\tilde{q}_{\mathrm{KD}}} = (1-\epsilon)\mathbf{q}+\epsilon \mathbf{p}_t .
		\end{aligned}
	\end{equation}
	Consistent with LC in Eq~(\ref{eq:label_correction}), $\mathrm{L}_\mathrm{KD}(\mathbf{q},\mathbf{p}_t, \mathbf{p})$ revises a label using the knowledge $\mathbf{p}_t$. \hfill\(\Box\)

	\noindent
	\textbf{Proposition 4.} \textit{Compared with CCE, LS and CP penalise entropy minimisation while LC reward it.}

	\noindent
	\textbf{Proposition 5.} \textit{In CCE, LS and CP, a data point $\mathbf{x}$ has the same semantic class. In addition, $\mathbf{x}$ has an identical probability of belonging to other classes except for its semantic class.
	}
	%
	%

	The proofs of propositions 4 and 5 are presented in the Appendix \ref{appendix_sec:proof_of_propositions}. 
	Only LC exploits informative information and has the ability to correct labels, while LS and CP only relax the hard targets.
	We summarise CCE, LS, CP and LC in Table~\ref{table:summary_CE_LS_CP_LC}. Constant terms are ignored for concision.



	\begin{table}[!t]
		\caption{
			The values of $\epsilon_{\mathrm{ProSelfLC}} =\mathrm{g}(t) \times\mathrm{l}(\mathbf{p})$ under different cases. 
			We use concrete values for concise interpretation.   
			We bold the special case when the semantic class is changed.
			Consistency is defined by whether $\mathbf{p}$ and $\mathbf{q}$ share the semantic class or not. 
			%
		}
		\centering
		\setlength{\tabcolsep}{4.2pt} 
		
		\vspace{-0.22cm}
		\begin{tabular}{l c c c c}
			\toprule
			
			
			~~~~~~~~~~~~ &  & \makecell[c]{$\mathrm{l}( \mathbf{p})=0.1$\\(non-confident)} & \makecell[c]{$\mathrm{l}( \mathbf{p})=0.9$\\(confidently \\consistent)} & \makecell[c]{$\mathrm{l}( \mathbf{p})=0.9$\\(confidently\\ inconsistent)} \\
			
			\midrule
			{Earlier phase}  & $\mathrm{g}( t)=0.1$ & 0.01 & ~~~~0.09 & {0.09}\\
			{Later phase} & $\mathrm{g}( t)=0.9$ & 0.09 & ~~~~0.81 & \textbf{0.81}
			\\
			\bottomrule
		\end{tabular}
		\label{table:case_analysis_ProSelfLC}
	\end{table}

	\vspace{-0.1cm}
	\section{ProSelfLC: Progressive and adaptive label correction 
	}
	\label{section:proselflc}
	\vspace{-0.10cm}

	In the standard CCE, the semantic class is considered while the similarity hierarchy over all classes is ignored. This is mainly due to the difficulty of annotating the similarity structure for every $\mathbf{x}$, especially when $c$ is large \cite{xu2020variational}. 
	Recent progress demonstrates that there are some effective approaches to define the similarity structure over samples without annotation: 
	(1) In KD, an auxiliary teacher model provides a student model the similarity hierarchy information \cite{hinton2015distilling,muller2019does}; 
	(2) In Self LC, e.g., Boot-soft, a model can bootstrap itself by exploiting the knowledge it has learned so far. 
	We focus on the end-to-end Self LC and improve it in this work. 
	
	
	
	In Self LC, $\epsilon$ indicates how much a predicted label distribution is trusted.  
	In ProSelfLC, we propose to set it adaptively according to the learning time $t$ and $\mathrm{conf}(\mathbf{p})$. 
	For any $\mathbf{x}$, we summarise its equations of loss $\mathrm{L}$, label and trust below.  
	\vspace{-0.12cm}
	%
	%
	\begin{mathframed}
		\begin{fleqn}[-0.05in]
			\begin{equation}
				\begin{aligned}
					\mathrm{L}( \mathbf{\tilde{q}_{\mathrm{ProSelfLC}}}, \mathbf{p})  
					&=\mathrm{H}(\mathbf{\tilde{q}_{\mathrm{ProSelfLC}}}, \mathbf{p});
				\end{aligned}
			\end{equation}
			\begin{equation}
				\text{Label: } \mathbf{\tilde{q}_{\mathrm{ProSelfLC}}}
				=(1- \mathrm{trust}(t, \mathbf{p}))  \mathbf{q}+\mathrm{trust}(t, \mathbf{p}) \mathbf{p}; 
			\end{equation}
			\begin{equation}
				\text{Self trust } \epsilon_{\mathrm{ProSelfLC}}\text{: }
				\mathrm{trust}(t, \mathbf{p}) = 
				\mathrm{g}(t)
				\times 
				\mathrm{l}(\mathbf{p});
			\end{equation}
			\begin{equation}
				\text{~~~~- Global trust: } \mathrm{g}(t) = \mathrm{h}({t}/{\Gamma}-\Theta, \emB) \in (0,1); 
			\end{equation}
			\begin{equation}
				\text{~~~~- Local trust: } \mathrm{l}(\mathbf{p}) =  
				\begin{cases} 
					1; \\
					\mathrm{conf}_{\mathrm{top}}(\mathbf{p}); \\
					\mathrm{conf}_{\mathrm{all}}(\mathbf{p}).
				\end{cases}
			\end{equation}
		\end{fleqn}
		\vspace{0.08cm}
	\end{mathframed}
	$t$ and $\Gamma$ are the iteration counter and the number of total iterations, respectively. 
	$\mathrm{h}(\eta, \emB)= {1}/({1+\exp(-\eta \times \emB)})$ defines a sigmoid curve. Here, $\eta = {t}/{\Gamma}-\Theta, \text{where } \Theta \in [0, 1]$. $\Theta$ decides the inflection point.  
	$\emB, \Gamma$ are task-dependent and can be chosen according to a validation set in practice.  
	We show three options to compute the local trust $\mathrm{l}(\mathbf{p})$, being either a constant or knowledge confidence-dependent.  
	$\Gamma$ and $\Theta$ are highly correlated, therefore, we fix $\Theta=0.5$ and only tune $\Gamma$ following the standard practice. 
	For brevity, we refer to $\mathrm{trust}(t, \mathbf{p})$ as $\epsilon_{\mathrm{ProSelfLC}}$ and make them interchangeable when there is no confusion.

	\vspace{-0.1cm}
	\subsection{The design inspirations of self trust: $\epsilon_{\mathrm{ProSelfLC}}$}
	\vspace{-0.10cm}
	
	{\textbf{Global trust}}. $\mathrm{g}(t)$ denotes overall how much we trust a learner. $\mathrm{g}( t)$ grows as $t$ rises and is independent of data points, thus being global. 
	$\emB$ adjusts the exponentiation's base and growth speed of $\mathrm{g}( t)$.
	Theoretically and practically, $\mathrm{g}( t)$ could be many other formats. For example, for the sigmoid function $\mathrm{h}$, with no loss of generality, we use a logistic function. In practice, there are many other alternatives, e.g., generalised logistic functions, hyperbolic tangent functions, and smoothstep ones. We leave the exploration of these alternatives to future work. 

	The design of $\mathrm{g}( t)$ is inspired by the human learning process.  
	In the earlier learning phase, i.e., $t < \Gamma * \Theta$, $\mathrm{g}( t) < 0.5 \Rightarrow \epsilon_{\mathrm{ProSelfLC}} < 0.5, \forall \mathbf{p}$, so that the predefined supervision dominates and ProSelfLC only modifies the similarity structure a bit. 
	%
	%
	Because when a learner has not seen the training data for enough times, its knowledge $\mathbf{p}$ with respect to $\mathbf{x}$ is less reliable. 
	%
	%
	When it comes to the later training phase, i.e., $t > \Gamma * \Theta$, we have $\mathrm{g}( t) > 0.5$. 
	$\Gamma * \Theta$ represents the global trust inflection time.

	\noindent
	\textbf{Local trust}. $\mathrm{l}( \mathbf{p})$ represents how much we trust  $\mathbf{p}$. If $\mathrm{l}( \mathbf{p})=1$, all predictions are treated equally. When $\mathrm{l}( \mathbf{p})=\mathrm{conf}_{\mathrm{top}}(\mathbf{p})$ or $\mathrm{conf}_{\mathrm{all}}(\mathbf{p})$, a more confident prediction has a higher local trust. 
	$\mathrm{l}( \mathbf{p})$ is designed to regularise the later learning phase. 
	If $\mathbf{p}$ is of higher entropy, $\mathrm{l}( \mathbf{p})$ is lower, hence $\epsilon_{\mathrm{ProSelfLC}}$ is smaller. 
	If $\mathbf{p}$ is highly confident, we trust it and $\epsilon_{\mathrm{ProSelfLC}}$ is large. 
	%
	%
	%
	
	We will empirically discuss $\mathrm{g}( t)$ using different $B$ and three $\mathrm{l}( \mathbf{p})$ options in the Section~\ref{sec:ablation_study}, 
	where $\mathrm{conf}_{\mathrm{top}}(\mathbf{p})$ and $\mathrm{conf}_{\mathrm{all}}(\mathbf{p})$ are found to work better than their constant counterpart.   
	

	
	%
	%

	\vspace{-0.10cm}
	\subsection{Cases analysis}
	\vspace{-0.1cm}
	
	Due to the potential memorisation in the earlier phase (though less likely to be severe), we may get undesired  confidently wrong predictions for noisy labels, but their trust scores are small as $\mathrm{g}( t)$ is small.  
	We conduct the cases analysis of ProSelfLC in Table~\ref{table:case_analysis_ProSelfLC} and summarise its core tactics as follows:

	(1) {{Correct the similarity structure for every data point in all cases}}, thanks to exploiting the growing self knowledge of a learner as its training proceeds. 
	
	(2)
	{{Revise the semantic class when $t$ is large enough and $\mathbf{p}$ is confidently inconsistent.}}
	When both conditions are met, as highlighted in Table~\ref{table:case_analysis_ProSelfLC}, we have $\epsilon_{\mathrm{ProSelfLC}}  > 0.5$ and 
	$\argmax\nolimits_j \mathbf{p}(j|\mathbf{x}) \neq \argmax\nolimits_j \mathbf{q}(j|\mathbf{x})$. Therefore, $\mathbf{p}$ redefines the semantic class. 
	For example, if $\mathbf{p} = [0.95, 0.01, 0.04], \mathbf{q} = [0, 0, 1], \epsilon_{\mathrm{ProSelfLC}}=0.8 \Rightarrow \mathbf{\tilde{q}_{\mathrm{ProSelfLC}}}=(1- \epsilon_{\mathrm{ProSelfLC}})  \mathbf{q}+\epsilon_{\mathrm{ProSelfLC}} \mathbf{p}=[0.76, 0.008, 0.232]$.  
	%
	We emphasise that ProSelfLC also becomes robust against 
	lengthy exposure
	to the noisy data, which is empirically demonstrated in Fig.~\ref{fig:cce_dynamics_confidence_accuracy_resnet18}, Fig.~\ref{fig:proselflc_dynamics_confidence_accuracy_resnet18}, Fig.~\ref{fig:comprehensive_dynamics},  and Table~\ref{table:cifar100_SOTA_Symmetric}.  
	\section{Learn towards a low-temperature entropy state}
	
	
	
	%
	

	\subsection{Generic coarse signed calibration error}
	
	We first revisit the Expected Calibration Error (ECE) using multiple bins and our definition of Generic coarse Signed Calibration Error (GSCE). 
	Formally, according to \cite{brocker2009reliability}, with respect to any data point $(\mathbf{x}, y)$,  the network $\mathrm{z}$ is perfectly calibrated if 
	\begin{equation}
		\forall j, 
		\mathrm{P}(y=j~|~\mathbf{p}(j|\mathbf{x})) = \mathbf{p}(j|\mathbf{x}). 
	\end{equation}
	Recently, a weaker but more practical condition \cite{guo2017calibration,kumar2019verified,minderer2021revisiting}, where only the most likely class is considered, is named argmax or top-label calibration and adopted: 
	\begin{equation}
		\mathrm{P}(y \in \argmax _j \mathbf{p}(j|\mathbf{x})~|~\mathrm{conf}_{\mathrm{top}}(\mathbf{p})) = \mathrm{conf}_{\mathrm{top}}(\mathbf{p}). 
		\label{eq:top-label-calibration}
	\end{equation}
	To approximately compute Eq.~(\ref{eq:top-label-calibration}), an ECE estimator with multiple bins is proposed \cite{guo2017calibration,kumar2019verified,minderer2021revisiting}. It has three steps: (1) the predictions $\{\mathbf{p}^i\}_{i=1}^n$ of  $\mathbf{X}=\{(\mathbf{x}^i, y^i)\}_{i=1}^n$ are bucketed into $m$ bins (i.e., groups) $G_1, G_2,..., G_m$ based on $\mathrm{conf}_{\mathrm{top}}(\mathbf{p})$; (2) for each group, the absolute error between confidence mean $\mathrm{conf}(G_i)$ and accuracy $\mathrm{accu}(G_i)$ is computed; (3) we calculate the ECE by the expected error over bins. Let $|G_i|$ denote the number of samples in $G_i$, $\llbracket \cdot \rrbracket $ is the
	Iverson bracket, we summarize:  
	%
			\begin{equation}
				\mathrm{conf}_{\mathrm{top}}(G_i) = \sum\nolimits_{j\in G_i} \mathrm{conf}_{\mathrm{top}}(\mathbf{p}^j) / |G_i|,
			\end{equation}   
			\begin{equation}
				\mathrm{accu}(G_i) = \sum\nolimits_{j\in G_i} 
				\llbracket y^j \in \argmax _v \mathbf{p}(v|\mathbf{x}^j) \rrbracket
				/ |G_i|,
			\end{equation}
			\begin{equation}
				\mathrm{ECE}(\mathrm{z},\mathbf{X}) = \sum\nolimits_{i=1}^{m}|\mathrm{conf}_{\mathrm{top}}(G_i)-\mathrm{accu}(G_i)| \frac{|G_i|}{n}.  
				\label{eq: ece}
			\end{equation}  
	%
	It has two main inconveniences making ECE with multiple bins a tool to measure miscalibration: (1) it depends on the number of bins and the distribution of confidences; (2) if an ECE is large, we are unclear whether a model is over-confident or under-confident.  
	Therefore, we \textit{propose a signed, simpler and faster-to-compute alternative to capture the degree of miscalibration}, named GSCE:  
	\begin{equation}
		\mathrm{GSCE}(\mathrm{z},\mathbf{X}) =  
		\begin{cases} 
			\mathrm{GSCE}_{\mathrm{top}}(\mathrm{z},\mathbf{X}) = \mathrm{conf}_{\mathrm{top}}(\mathbf{X}) 
			-\mathrm{accu}(\mathbf{X});\\
			
			\mathrm{GSCE}_{\mathrm{all}}(\mathrm{z},\mathbf{X}) = \mathrm{conf}_{\mathrm{all}}(\mathbf{X}) 
			-\mathrm{accu}(\mathbf{X}).
		\end{cases}
	\end{equation}
	GSCE uses single bin, thus being a coarser metric than Eq.~(\ref{eq: ece}).
	However, it is more generic because $\mathrm{conf}(\cdot)$ could be $\mathrm{conf}_{\mathrm{all}}(\cdot)$ and other variants in addition to $\mathrm{conf}_{\mathrm{top}}(\cdot)$ in Eq.~(\ref{eq: ece}).  
	A positive GSCE represents an over-confident miscalibration while a negative GSCE denotes an under-confident one.  
	

	\subsection{Miscalibration analysis: a model has much lower confidence than accuracy before fitting noise}
	\label{subsec:Miscalibration analysis}
	
\begin{figure}[!t]
		\centering
		\begin{subfigure}[!h]{0.90\linewidth}
			\centering
			\captionsetup{width=\linewidth}
			\includegraphics[clip, trim=0.2cm 11.11cm 0.5cm 0.4cm, width=0.99\linewidth]{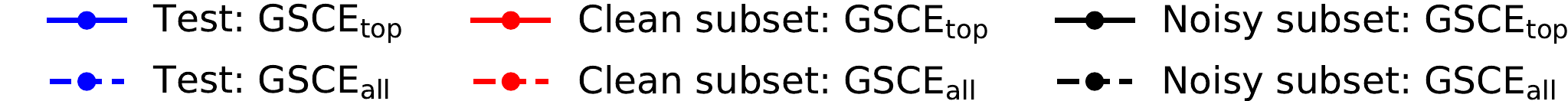}
		\end{subfigure}
		\vspace{0.10cm}
		
		\begin{subfigure}[!h]{0.492\linewidth}
			\centering
			\captionsetup{width=\linewidth}
			\includegraphics[clip, trim=0.45cm 0.30cm 7.3cm 1.78cm, width=0.99\linewidth]{gsce_topandall/shufflenetv2_crossentropy_withAT.pdf}
			\caption{ShuffleNetV2.}
		\end{subfigure}
		\begin{subfigure}[!h]{0.492\linewidth}
			\centering
			\captionsetup{width=\linewidth}
			\includegraphics[clip, trim=0.45cm 0.30cm 7.3cm 1.78cm, width=0.99\linewidth]{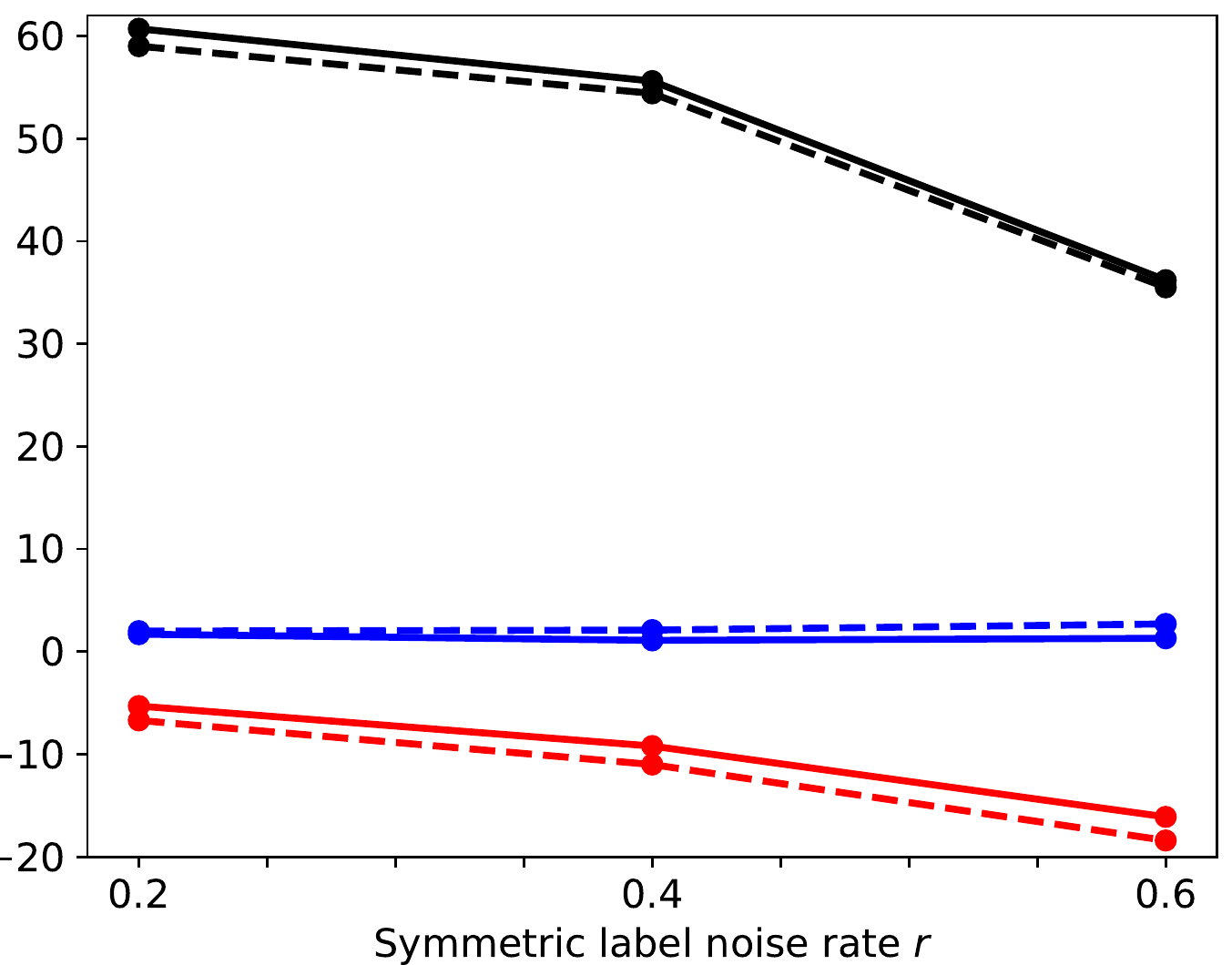}
			\caption{ResNet18.}
		\end{subfigure}
		\vspace{-0.19cm}
		\caption{ 
			Model miscalibration when training on CIFAR-100 using the standard CCE. 
			For a comprehensive illustration, we train two networks, use three symmetric label noise rates, and report 
			$\mathrm{GSCE}_{\mathrm{top}}$ and $\mathrm{GSCE}_{\mathrm{all}}$.  
			The final model is used when the training stops. 
			All results are multiplied by 100 before plotting, for clear illustrations. 
		}
		\label{fig:cce_accuracy_confidence_metrics}
	\end{figure}

\begin{figure}[!t]
		\centering
		\begin{subfigure}[!h]{0.90\linewidth}
			\centering
			\captionsetup{width=\linewidth}
			\includegraphics[clip, trim=0.2cm 11.018cm 0.5cm 0.4cm, width=0.99\linewidth]{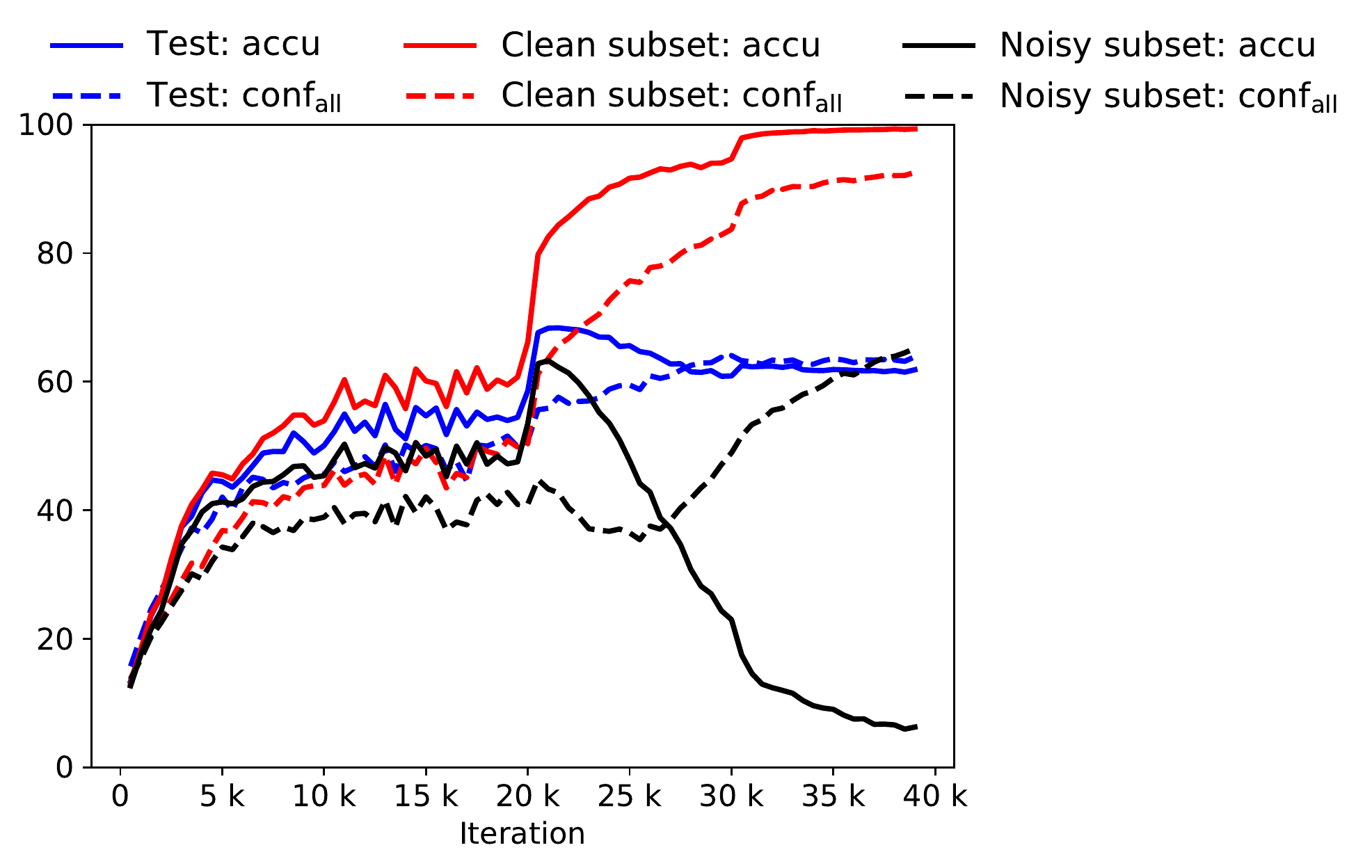}
		\end{subfigure}
		\vspace{0.15cm}
		
		\begin{subfigure}[!h]{0.513\linewidth}
			\centering
			\captionsetup{width=\linewidth}
			\includegraphics[clip, trim=0.32cm 0.32cm 5.8cm 1.66cm, width=0.99\linewidth]{multiloss_accuracy_confidence/0.2_resnet18/005_labelsmoothing_warm0_20220403-075152/all_subsets.pdf}
			\caption{
				$r=20\%$.
			}
		\end{subfigure}
		\begin{subfigure}[!h]{0.476\linewidth}
			\centering
			\captionsetup{width=\linewidth}
			\includegraphics[clip, trim=1.31cm 0.32cm 5.8cm 1.66cm, width=0.99\linewidth]{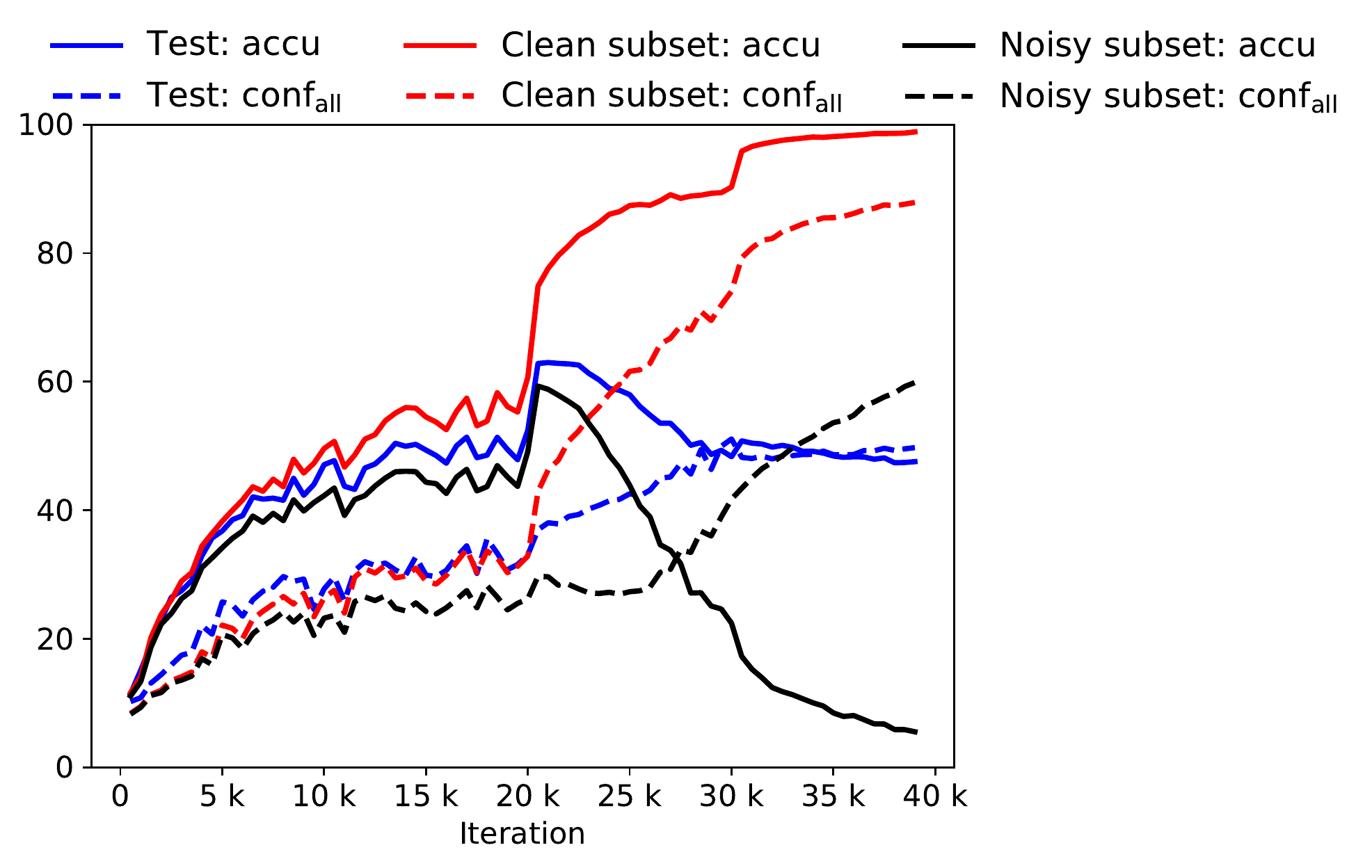}
			\caption{
				$r=40\%$.
			}	
		\end{subfigure}
		\vspace{-0.19cm}
		\caption{ 
			$\mathrm{accu}$ and $\mathrm{conf}_{\mathrm{all}}$ 
			along with the iteration
			when training ResNet18 on CIFAR-100 using the standard CCE. The symmetric noise rate is $r$. For a more transparent and stratified analysis, we store models of different iterations and report the results of three sets, i.e., test set, clean and noisy subsets of the training data.    
			All results are multiplied by 100 before plotting.   
		}
		\label{fig:cce_dynamics_confidence_accuracy_resnet18}
	\end{figure}

	There exist three vital findings about the learning behaviours of deep networks: 
	(1) deep models easily fit random noise \cite{zhang2017understanding}; 
	(2) deep networks learn simple semantic patterns before fitting noise \cite{arpit2017closer}; 
	(3) modern deep neural works tend to be over-confident \cite{guo2017calibration,minderer2021revisiting}. 
	In this section, we disclose a new notable one: 
	
	\begin{center}
		\textit{When a higher label noise exists, deep models are significantly less confident of learning semantically meaningful patterns before fitting noise. }
	\end{center} 
	This discovery is illustrated in  Fig.~\ref{fig:cce_accuracy_confidence_metrics} and Fig.~\ref{fig:cce_dynamics_confidence_accuracy_resnet18} with details below.
	
	In Fig.~\ref{fig:cce_accuracy_confidence_metrics}, we have some counter-intuitive observations: for both small (ShuffleNetV2) and large (ResNet18) networks, regardless of using $\mathrm{GSCE}_{\mathrm{top}}$ or $\mathrm{GSCE}_{\mathrm{all}}$ as the miscalibration metric, {the model is over-confident in noisy training data while under-confident in clean training data, and has a small miscalibration on the test data.} 
	However, it is well known that a model trained using the CCE generalises poorly when noise exists \cite{zhang2017understanding,arpit2017closer}. 
	
	In Fig.~\ref{fig:cce_dynamics_confidence_accuracy_resnet18}, the learning track along with the iteration helps us comprehend the observations in Fig.~\ref{fig:cce_accuracy_confidence_metrics}. 
	We observe:   
	(1) the model is highly miscalibrated and has much higher accuracy than its confidence across three sets before fitting noise (i.e., before the accuracy of the noisy subset starts to drop); 
	(2) the miscalibration becomes more dramatic when $r$ changes from $20\%$ to $40\%$; 
	(3) when the model starts to fit noise at around 21k iterations, the miscalibration hits a negative peak.

	\begin{figure}[!t]
		\centering
		\begin{subfigure}[!h]{0.90\linewidth}
			\centering
			\captionsetup{width=\linewidth}
			\includegraphics[clip, trim=0.2cm 11.018cm 0.5cm 0.4cm, width=0.99\linewidth]{multiloss_accuracy_confidence/0.2_resnet18/005_labelsmoothing_warm0_20220403-075152/all_subsets.pdf}
		\end{subfigure}
		\vspace{0.15cm}
		
		\begin{subfigure}[!h]{0.513\linewidth}
			\centering
			\captionsetup{width=\linewidth}
			\includegraphics[clip, trim=0.32cm 0.32cm 5.8cm 1.66cm, width=0.99\linewidth]{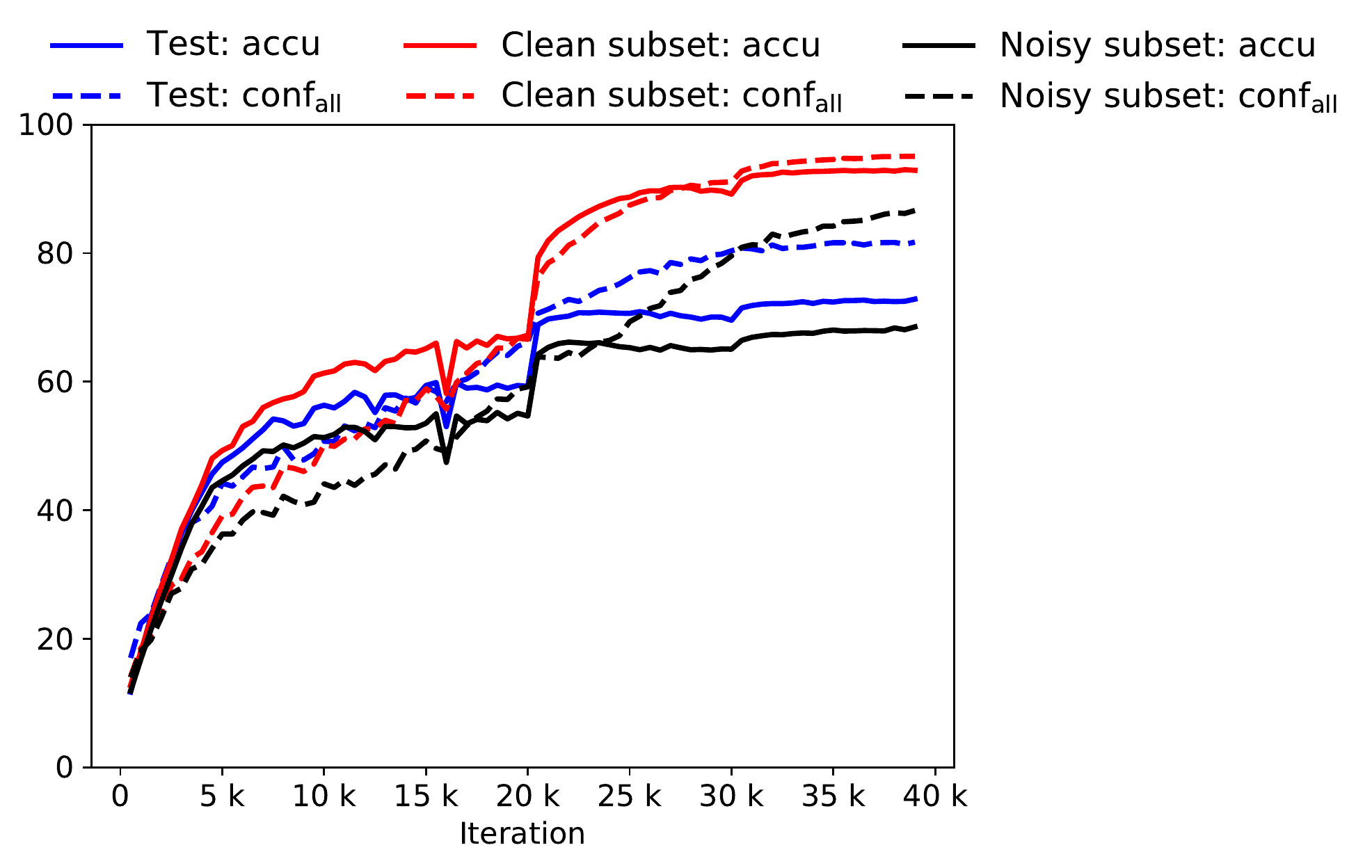}
			\caption{
				$r=20\%$.
			}
		\end{subfigure}
		\begin{subfigure}[!h]{0.476\linewidth}
			\centering
			\captionsetup{width=\linewidth}
			\includegraphics[clip, trim=1.31cm 0.32cm 5.8cm 1.66cm, width=0.99\linewidth]{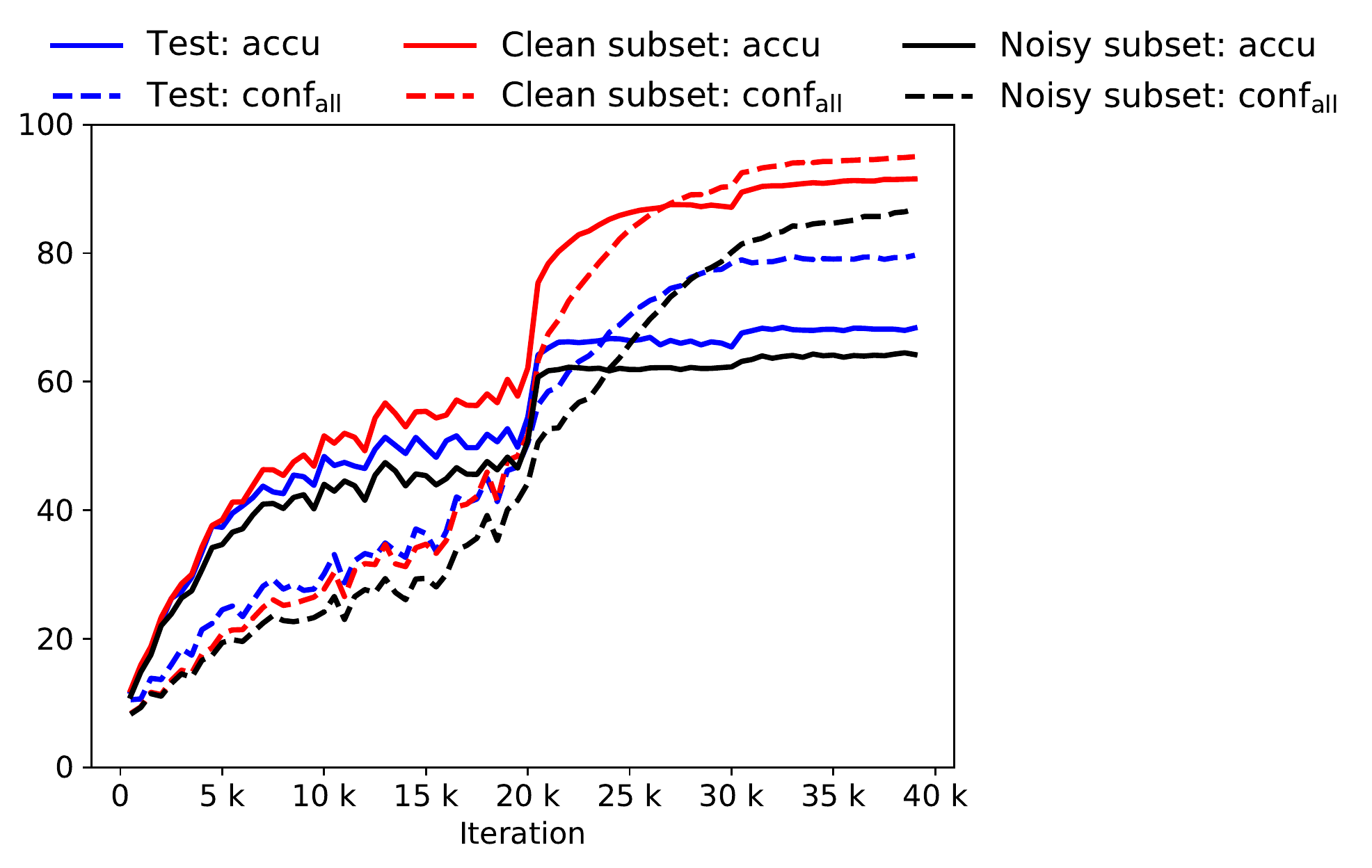}
			\caption{
				$r=40\%$.
			}	
		\end{subfigure}
		\vspace{-0.19cm}
		\caption{ 
			$\mathrm{accu}$ and $\mathrm{conf}_{\mathrm{all}}$ along with the iteration when using our proposed ProSelfLC with an AT. The experiment and plot settings are exactly the same as Fig.~\ref{fig:cce_dynamics_confidence_accuracy_resnet18}. 
			ProSelfLC increases the knowledge and confidence on three sets along with the iteration.  
		}
		\label{fig:proselflc_dynamics_confidence_accuracy_resnet18}
	\end{figure}

	\subsection{Integrate an AT into ProSelfLC}
	
	Inspired by the analysis in subsection~\ref{subsec:Miscalibration analysis}, 
	we apply a low temperature $T$ to decrease the entropy of self knowledge without affecting its accuracy. 
	By exploiting it, we are able to define a revised low-entropy target state.  
	Mathematically, other than using $\mathbf{p}$ to correct labels, we use   
	\begin{equation}
		\begin{aligned}
			\mathbf{p}_{T}(j|\mathbf{x})= 
			{\exp (\mathbf{z}_{j} /T )}
			/{ \sum\nolimits_{m=1}^{C} \exp (\mathbf{z}_{m}  /T ) }.
		\end{aligned}
		\label{chapter:DM_eq:softmax_normalisation}
	\end{equation}  
	An annealed temperature (denoted by AT, $0 < T < 1$) works better consistently, which is demonstrated by our extensive experiments, e.g., Fig~\ref{fig:proselflc_dynamics_confidence_accuracy_resnet18}.  
	%
	For a comprehensive analysis, we also study AT integrated with other target modification approaches in the Section~\ref{sec:ablation_study}. 
	Interestingly, AT boosts the low-entropy rewarding algorithms (i.e., Boot-soft and ProSelfLC) but does not help the low-entropy penalising method (i.e., CP).

	\section{Experiments}

	In deep learning, 
	due to the stochastic batch-wise training scheme, 
	small implementation differences (e.g., random accelerators like cudnn and different frameworks like Caffe \cite{jia2014caffe}, Tensorflow \cite{abadi2016tensorflow} and PyTorch \cite{paszke2019pytorch}) may lead to a large gap of final performance. 
	Therefore, to compare more properly, we re-implement CCE, LS and CP using PyTorch. 
	To allow for a deterministic behaviour in all experiments, we follow the PyTorch reproducibility guidelines\footnote{\url{https://pytorch.org/docs/stable/notes/randomness.html\#reproducibility}} to fix all randomness sources, e.g., using identical seeds in all experiments, avoiding nondeterministic algorithms, and setting the CUDA environment variable $\text{CUBLAS\_WORKSPACE\_CONFIG=:4096:8}$\footnote{\url{https://docs.nvidia.com/cuda/cublas/index.html\#cublasApi_reproducibility}}, etc. 
	Regarding Self LC methods, we re-implement Boot-soft \cite{reed2015training}, where $\epsilon$ is fixed throughout training.  
	We do not re-implement stage-wise Self LC and KD methods, e.g., Joint Optimisation and Tf-KD$_{self}$ respectively, because time-consuming tuning is required. 
	In addition, our ProSelfLC can also be treated as an iteration-wise Self LC method. 
	%
	%
	%
	By default, 
	in clean and synthetic noisy cases, 
	we train on 80\% training data (corrupted in synthetic noisy cases) and use 20\% trusted training data as a validation set to search all hyperparameters, e.g., $\Gamma, \epsilon, \emB, T$ and settings of an optimiser.  
	{Note that $\Gamma$ and an optimiser's settings are searched first and then fixed for all methods.}  
	Finally, we retrain a model on the entire training data (corrupted in synthetic noisy cases) and report its accuracy on the test data to fairly compare with prior results. 
	In real-world label noise, the used datasets have a separate clean validation set.
	%
	Here, we use the clean dataset only for validating hyperparameters, instead of training a network's learnable parameters in \cite{vahdat2017toward,veit2017learning,jiang2018mentornet,ren2018learning,li2017learning,hendrycks2018using,yeung2017learning,zhang2020distilling}. 
	
	%

	\begin{table}[!t]
		\caption{
			Test accuracy (\%) on CIFAR-100 clean test set in the clean setting. 
			We put the intermediately obtained best accuracy in the bracket.  
			A lower drop from an intermediate best accuracy to the final one can be interpreted as a learner's higher robustness against a long time being exposed to the training data. 
			We bold the best results. The criteria of ``best'' is  a highest intermediate accuracy followed with a highest final accuracy.    
		}
		\centering
		\setlength{\tabcolsep}{8.99pt} 
		\begin{tabular}{lcc}
			\toprule
			Method & ShuffleNetV2 & ResNet-18 \\
			\midrule
			CCE & 70.1 (70.4)  & 74.5 (74.8) \\
			LS & $70.3\pm0.3$ ($70.7\pm0.4$) &$75.6\pm0.2$ ($75.9\pm0.3$)\\
			
			CP & $ 70.3\pm0.2$ ($70.5\pm0.1$) &$75.1\pm0.1$ ($75.5 \pm 0.3$)\\
			
			Boot-soft & $70.3\pm0.3$ ($70.6\pm0.4$) &$75.1\pm0.2$ ($75.3\pm0.1$)\\
			
			ProSelfLC & $\textbf{71.2}\pm0.2$ ($\textbf{71.4}\pm0.2$) &$\textbf{76.2}\pm0.2$ ($\textbf{76.3}\pm0.2$)\\
			\bottomrule
		\end{tabular}
		\label{table:clean_setting}
	\end{table}

	\begin{figure*}[!t]
		\centering
		\begin{subfigure}[!h]{0.33\textwidth}
			\centering
			\captionsetup{width=\textwidth}
			\includegraphics[clip, trim=3.05cm 8.9cm 4.2cm 7.9cm, width=0.99\textwidth]{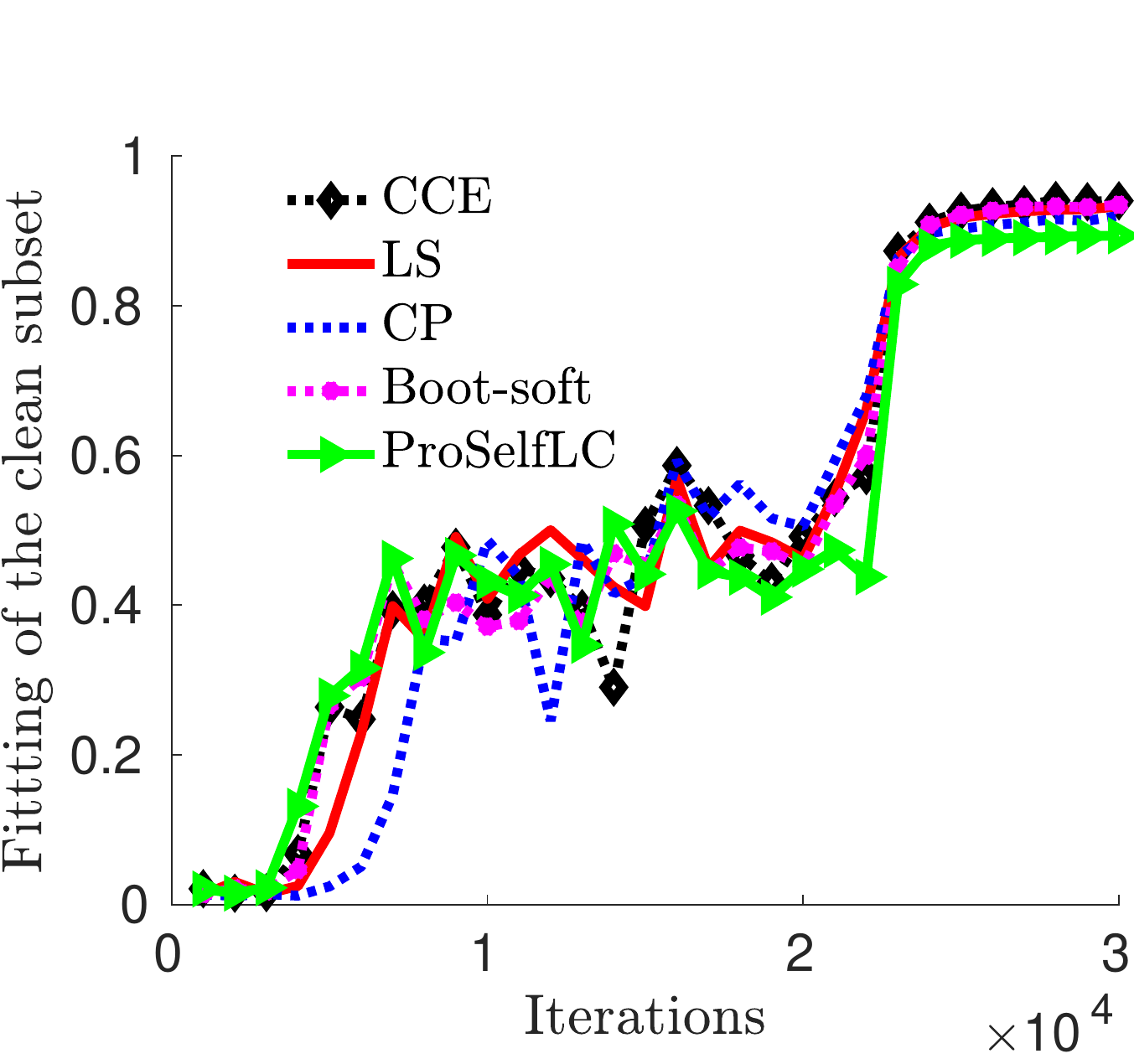}
			\caption{Correct fitting.}
			\label{fig:CIFAR100Asy0_4_cleanFittting}
		\end{subfigure}
		\begin{subfigure}[!h]{0.329\textwidth}
			\centering
			\captionsetup{width=\textwidth}
			\includegraphics[clip, trim=3.05cm 8.9cm 4.2cm 7.9cm, width=0.99\textwidth]{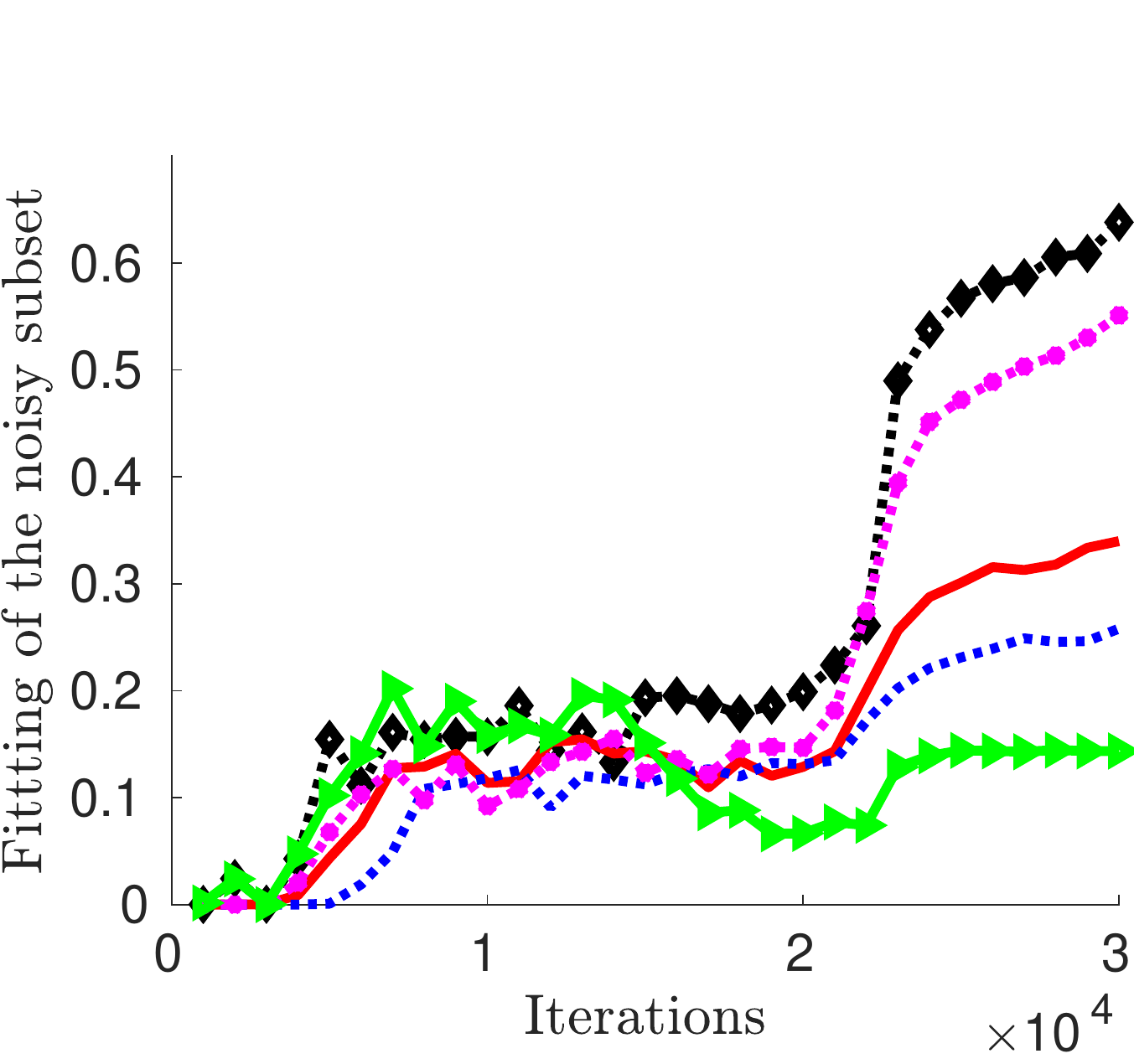}
			\caption{Wrong fitting.}
			\label{fig:CIFAR100Asy0_4_corruptedFittting}
		\end{subfigure}
		\begin{subfigure}[!h]{0.329\textwidth}
			\centering
			\captionsetup{width=\textwidth}
			\includegraphics[clip, trim=3.05cm 8.9cm 4.2cm 7.9cm, width=0.99\textwidth]{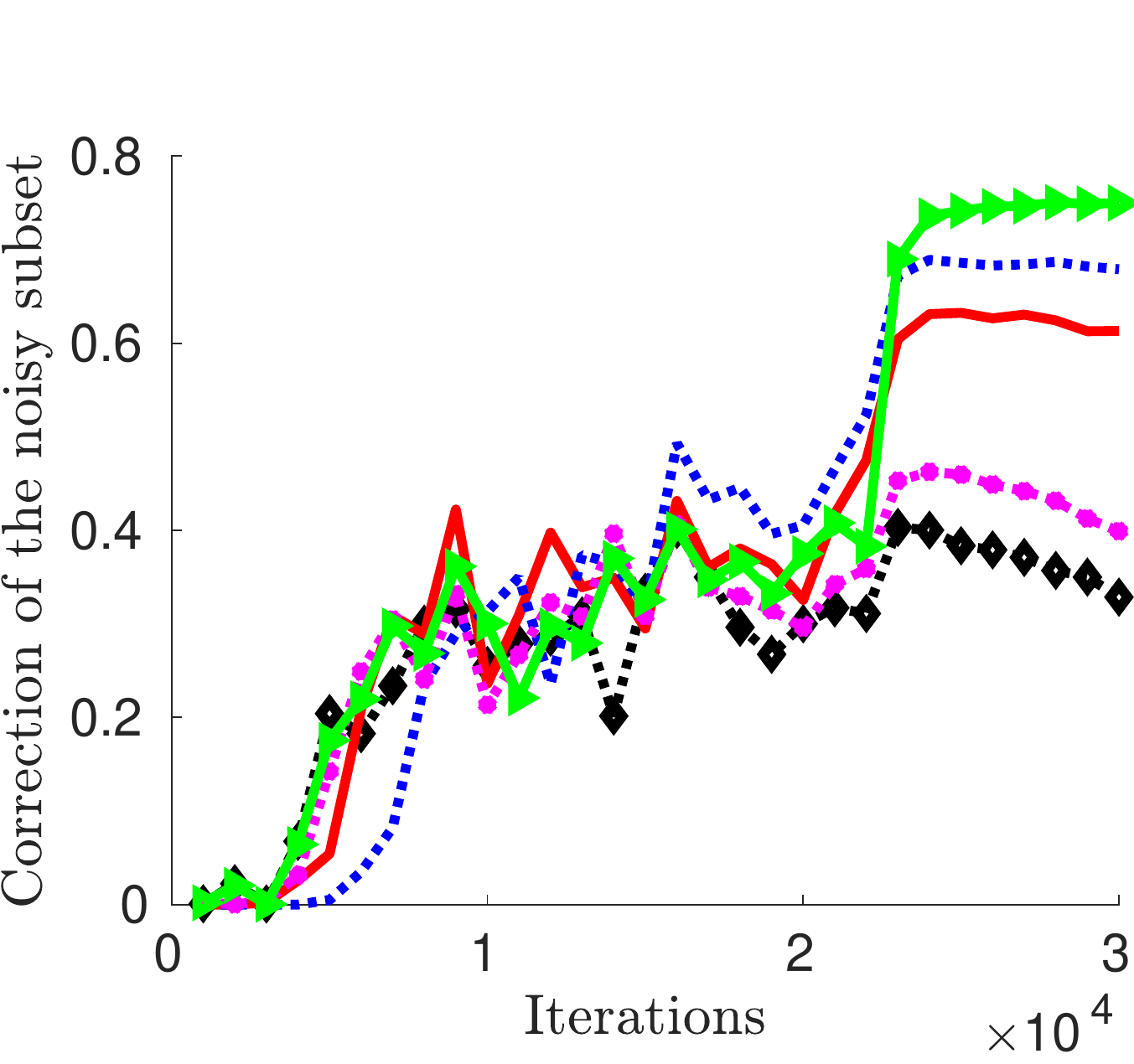}
			\caption{Semantic class correction}
			\label{fig:CIFAR100Asy0_4_semanticCorrection}
		\end{subfigure}
		
		\vspace{-0.39cm}
		\begin{subfigure}[!h]{0.329\textwidth}
			\centering
			\captionsetup{width=\textwidth}
			\includegraphics[clip, trim=3.05cm 8.9cm 4.2cm 7.9cm, width=0.99\textwidth]{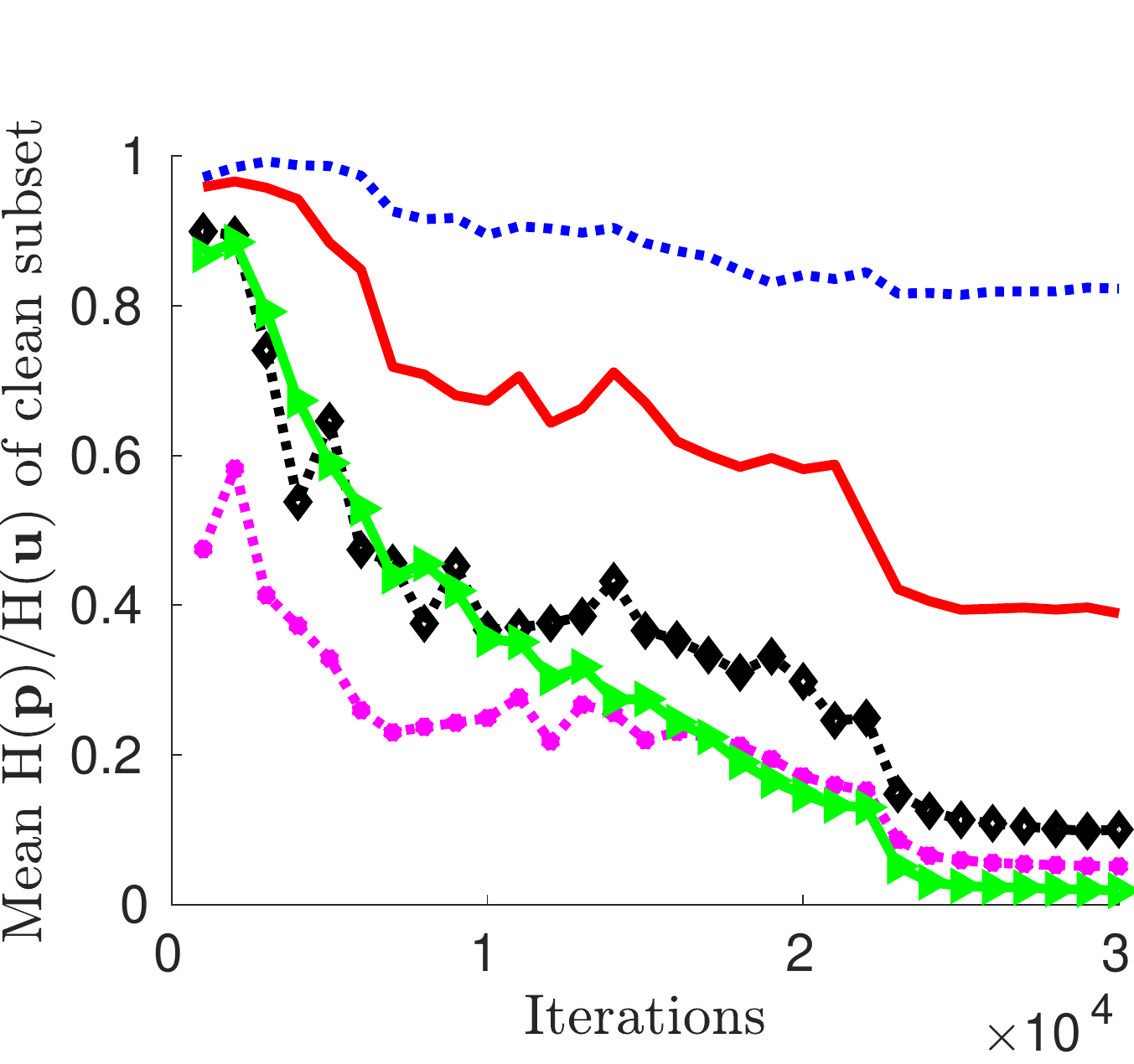}
			\caption{Entropy of clean subset.}
			\label{fig:CIFAR100Asy0_4_mean_entropy_CleanSubset}
		\end{subfigure}
		\begin{subfigure}[!h]{0.329\textwidth}
			\centering
			\captionsetup{width=\textwidth}
			\includegraphics[clip, trim=3.05cm 8.9cm 4.2cm 7.9cm, width=0.99\textwidth]{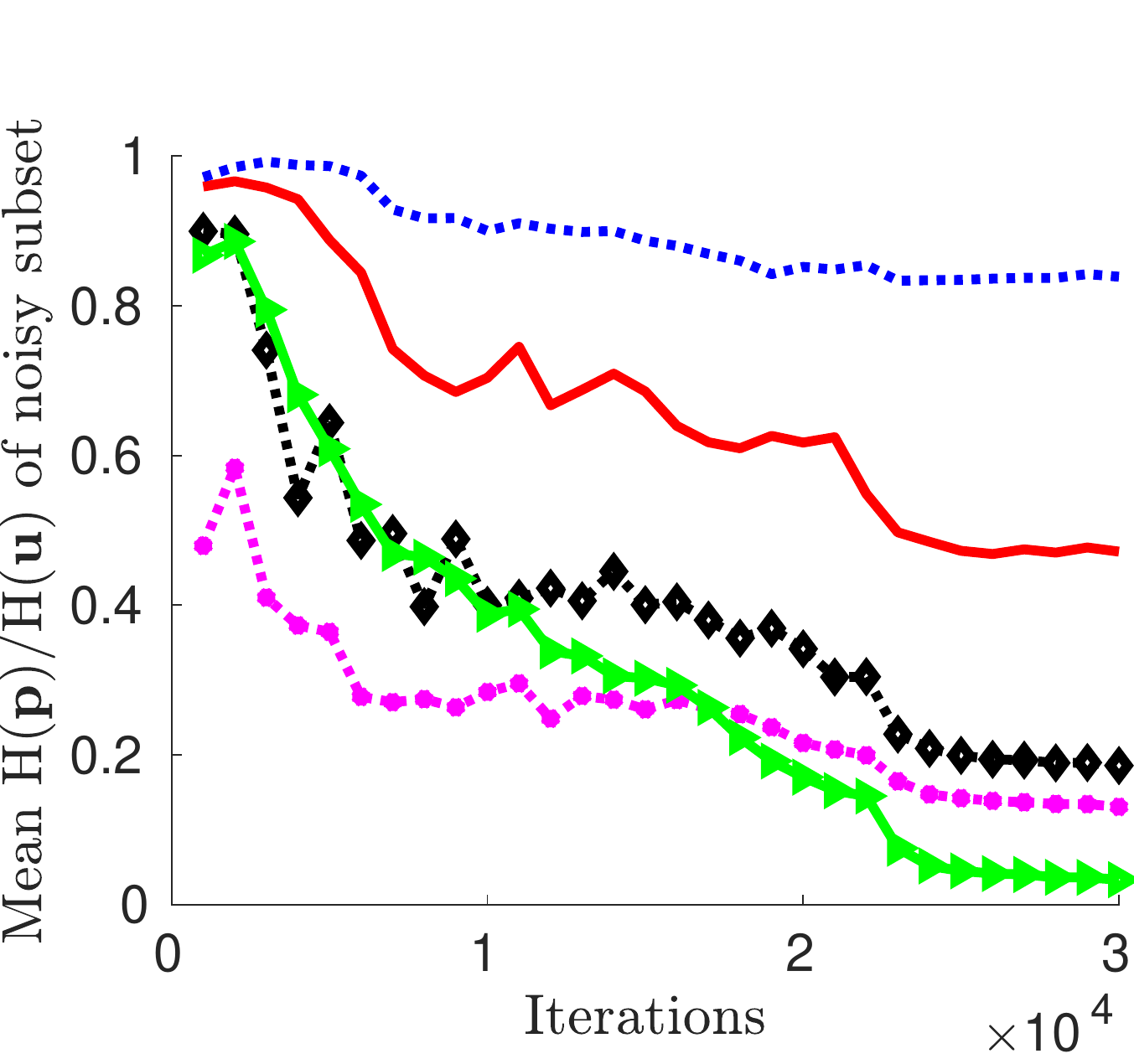}
			\caption{Entropy of noisy subset.}
			\label{fig:CIFAR100Asy0_4_mean_entropy_NoisySubset}
		\end{subfigure}
		\begin{subfigure}[!h]{0.33\textwidth}
			\centering
			\captionsetup{width=\textwidth}
			\includegraphics[clip, trim=3.05cm 8.9cm 4.2cm 7.9cm, width=0.99\textwidth]{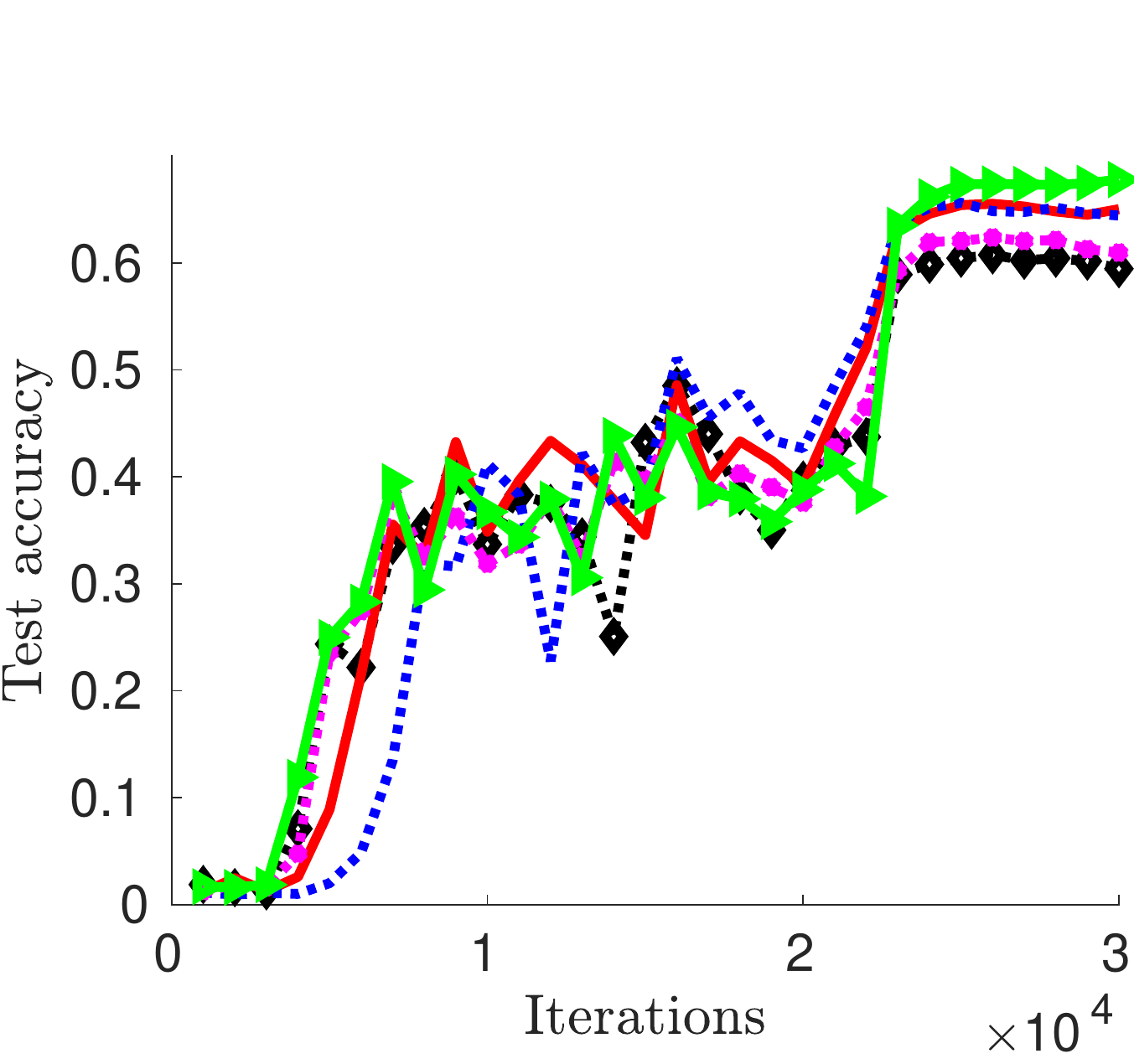}
			\caption{Generalisation.}
			\label{fig:CIFAR100Asy0_4_accuracy30000}
		\end{subfigure}
		%
		
		\caption{ 
			Dynamic learning statistics on CIFAR-100 with $40\%$ asymmetric label noise. 
			{{At training, a learner is \textit{not given} whether a label is trusted or not.}} Therefore, the target modification methods can be treated as unsupervised. 
			We store all intermediate models and plot their results of six  metrics. 
			Vertical axes and subcaptions describe evaluation metrics.    
		}
		\label{fig:comprehensive_dynamics}
	\end{figure*}

	\subsection{Compare with baselines on clean CIFAR-100}
	\label{sec:clean_setting}

	
	%
	\textbf{Dataset and training details.}
	{CIFAR-100} \cite{krizhevsky2009learning} has 20 coarse categories and 5 fine classes in a coarse class. 
	There are 500 and 100 images per class in the training and testing sets, respectively. 
	The image size is 32$\times$32. 
	We apply simple data augmentation \cite{he2016deep}, i.e., we pad 4 pixels on every side of the image, and then randomly crop it with a size of 32$\times$32. 
	Finally, this crop is horizontally flipped with a probability of 0.5.
	We train the widely used ShuffleNetV2 \cite{ma2018shufflenet} and ResNet-18 \cite{he2016deep}.
	SGD is used with its settings as: (a) a learning rate of 0.2; (b) a momentum of 0.9; (c) the batch size is 128 and the number of training iterations is 39k, i.e., 100 epochs. 
	We divide the learning rate by 10 at 20k and 30k iterations.  
	The weight decay is $2e$-$3$ for ResNet-18 while $1e$-$3$ for ShuffleNetV2 because ResNet-18 has a larger fitting capacity.  
	%
	%
	%

	\noindent
	\textbf{Result analysis.} 
	We check the methods' sensitivity to hyper-parameters, which generally makes more sense than that to random seeds.
	We report the mean and standard deviation of multiple hyper-parameters other than random seeds.  
	Therefore, CCE has one run. While for LS, CP and Boot-soft, we run several different $\epsilon$ and $T$.  Analogously, for ProSelfLC, we run several $B$ and $T$. 
	In detail,  $\epsilon \in [0.125, 0.25, 0.375, 0.50]$, $T \in [1.0, 0.8, 0.6, 0.4]$, $B \in [20, 16, 12, 8].$
	The hyper-parameter space size is the same for each method except for CCE and LS. 
	We report the mean and standard deviation of the top three results in Table~\ref{table:clean_setting}. 
	Compared with the baselines, ProSelfLC performs the best for both networks. 
	First, we do not observe a big enough difference of those methods in the clean setting. Therefore, we focus on noisy scenarios hereafter. 
	Second, the sensitivity to hyper-parameters is also small. 
	Therefore, we do not report the standard deviation in the Table~\ref{table:cifar100_SOTA_Symmetric}. 
	Instead, we further discuss hyper-parameters in the Section~\ref{sec:ablation_study}.

	\begin{table}
		\captionsetup{width=\linewidth}
		\caption{
			Test accuracy (\%) on CIFAR-100 clean test set.  
			The train set has a symmetric label noise rate of $r$. 
			All compared methods use ResNet-18. 
			We show the final accuracy and put the intermediate best 
			one in the corresponding bracket.  
			A lower drop from the intermediate best to the final can be interpreted as a model's higher robustness against a long time being exposed to the noise. 
			We bold the best results. 
		}
		\label{table:cifar100_SOTA_Symmetric}
		\centering
		\setlength{\tabcolsep}{5.6pt} 
		\begin{tabular}{ccccc}
			\toprule
			\multirow{2}{*}{} & 
			& \multicolumn{3}{c}{Symmetric Noisy Labels}  \\
			\cmidrule(lr){3-5}
			& Method &  ~~~$r$=0.2 & ~~~$r$=0.4 & $r$=0.6\\
			\midrule
			
			\multirow{15}{*}{\specialcell{Results From \\ Topo \cite{wu2020topological}}}
			& CCE & 56.5 & 50.4 & 38.7 \\
			&Forgetting & 56.5 & 50.6 & 38.7 \\
			&Forward & 56.4 & 49.7 & 38.0 \\
			&Decoupling & 57.8 & 49.9 & 37.8 \\
			&MentorNet & 62.9 & 52.8 & 36.0 \\
			&Co-teaching & 64.8 & 60.3 & 46.8 \\
			&Co-teaching+ & 64.2 & 53.1 & 25.3 \\
			&IterNLD & 57.9 & 51.2 & 38.1 \\
			&RoG & 63.1 & 58.2 & 47.4 \\
			&PENCIL & 64.9 & 61.3 & 46.6 \\
			&GCE & 63.6 & 59.8 & 46.5 \\
			&SL & 62.1 & 55.6 & 42.7 \\
			&TopoFilter & 65.6 & 62.0 & 47.7 \\
			
			\midrule
			\multirow{5}{*}{\specialcell{Our Trained \\Results}}

			&CCE & 61.9 (68.4) & 47.6 (63.0) & 32.2 (55.9) \\
			& LS & 68.4 (70.8) & 57.6 (64.8) & 41.9 (54.0) \\
			& CP & 66.1 (69.5) & 54.1 (63.9) & 38.8 (55.0) \\
			& Boot-soft  & 68.5 (69.6) & 61.1 (63.7) & 45.4 (54.5) \\
			& ProSelfLC  & \textbf{72.8} (\textbf{72.8}) & \textbf{68.4} (\textbf{68.4}) & \textbf{58.8} (\textbf{59.0}) \\
			
			\bottomrule
		\end{tabular}
	\end{table}

	\begin{table}[!t]
		\caption{
			Test accuracy (\%) on CIFAR-100 clean test set when the train set has 40\% symmetric label noise. 
			The Self LC (i.e., Self KD) baselines are implemented in \cite{li2021mutual} to test their robustness under label noise. 
			The best results are bolded.
		}
		\centering
		\setlength{\tabcolsep}{3.999pt} 
		\begin{tabular}{lccccc}
			\toprule
			& Tf-KD$_{reg}$ \cite{yuan2020revisiting} & SSKD \cite{xu2020knowledge} & Li's LC \cite{li2021mutual} & ProSelfLC \\
			\midrule
			ShuffleNetV2 & 44.7 &  57.2 & 61.2 & \textbf{61.6} \\
			ResNet-18 & 51.1 &  52.8 & 55.9 & \textbf{68.4} \\
			\bottomrule
		\end{tabular}
		\label{table:selflc_selfKD_comparison}
	\end{table}

	\subsection{Compare with the state-of-the-art methods on noisy CIFAR-100}
	\label{sec:cifar100} 

	\textbf{Generating noisy train labels}. 
	(1) {Symmetric label noise}: the original label of an image is uniformly changed to one of the other classes with a probability of $r$.
	(2) {Asymmetric label noise}: we follow \cite{wang2019symmetric} to generate asymmetric label noise.  
	Within each coarse class, we randomly select two fine classes $A$ and $B$. Then we flip $r\times100\%$ labels of $A$ to $B$, and $r\times100\%$ labels of $B$ to $A$. We remark that the overall label noise rate is smaller than $r$.

	\noindent
	\textbf{Competitors.}\footnote{We do not consider DisturbLabel \cite{xie2016disturblabel}, which flips labels randomly and is counter-intuitive. It weakens the generalisation because generally the accuracy drops as the uniform label noise increases.}  
	We compare with the results reported recently in SL \cite{wang2019symmetric} and Topo \cite{wu2020topological}.
	Forward is a loss correction approach that uses a noise-transition matrix \cite{patrini2017making}.  
	GCE denotes generalised cross entropy \cite{zhang2018generalized} and SL is symmetric cross entropy \cite{wang2019symmetric}. They are robust losses designed for solving label noise. 
	Regarding the other robust loss functions including 
	focal loss (FL) \cite{lin2017focal}, 
	NLNL \cite{kim2019nlnl}, 
	and normalised losses \cite{ma2020normalized,wang2019imae,wang2019derivative}, according to the experimental report in  active passive loss (APL) \cite{ma2020normalized} where a deeper ResNet-34 is used though, their results are much worse than ours. Therefore, we do not compare with them in the table. 
	Similarly, although TVD \cite{zhang2021learningicml} uses ResNet-18, its reported results are much lower and not compared in the table. 
	The other recent approaches are Forgetting \cite{arpit2017closer},  
	Decoupling \cite{malach2017decoupling},  
	MentorNet \cite{jiang2018mentornet}, 
	Co-teaching \cite{han2018co},  
	Co-teaching+ \cite{yu2019does}, 
	IterNLD \cite{wang2018iterative},  
	RoG \cite{lee2019robust}, PENCIL \cite{yi2019probabilistic} and 
	TopoFilter \cite{wu2020topological}.
	%
	Tf-KD$_{reg}$ \cite{yuan2020revisiting}, SSKD \cite{xu2020knowledge} and Li's LC \cite{li2021mutual} are three Self LC methods. 
	%
	%

	\noindent
	\textbf{Results analysis.} 
	Training details are the same as Section ~\ref{sec:clean_setting}.
	For all methods, we report their final results when training terminates. Therefore, {we test the robustness of a model against not only label noise, but also a long time being exposed to the noise.}
	In Table~\ref{table:cifar100_SOTA_Symmetric}, we observe that: 
	(1) 
	ProSelfLC outperforms all the baselines, which is significant in most cases; 
	(2) 
	By default, we use AT and better baseline results are obtained. 
	Despite that, our ProSelfLC further improves the standard Self LC (i.e., Boot-soft) and is the best of all.
	%
	In addition, we visualize and comprehend the dynamic learning statistics of ProSelfLC versus baselines in Fig.~\ref{fig:comprehensive_dynamics}, which clarifies why ProSelfLC works better. 
	According to Table~\ref{table:selflc_selfKD_comparison}, ProSelfLC is superior to other recent Self LC methods.  
	%
	%
	%
	%

	
	%
	%
	%
	%
	%
	%
	\noindent\textbf{Revising the semantic class and similarity structure.} 
	In Fig. \ref{fig:CIFAR100Asy0_4_corruptedFittting} and Fig.~\ref{fig:CIFAR100Asy0_4_semanticCorrection}, we show dynamic statistics of different approaches on fitting wrong labels and correcting them, respectively. ProSelfLC is much better than its counterparts. High semantic class correction means that the learned similarity structure revises the semantic class and similarity hierarchy corrupted by the noise.  
	%
	%
	%
	%
	%
	%

	%
	%
	%
	%
	%

	\begin{table}[!t]
		\caption{
			Test accuracy (\%) on the real-world noisy dataset Clothing1M, which contains asymmetric noise \cite{yi2019probabilistic} and instance-dependent noise \cite{berthon2021confidence,qu2021dat}. 
			We note that some approaches use the full training set while others sample a label-balanced subset. As two practices affect the performance a lot, we group the results into two columns for a clear and fair comparison. 
			For the sampled noisy training data, each label has 18976 images, leading to about 260k images in total.
			* indicates online label-balanced sampling for each mini-batch. 
			%
			The first two blocks present the results from ICLR 2022 papers \cite{xia2022sample} and  \cite{jiang2022information}, respectively.  The fourth block contains the results from \cite{chen2021noise}. 
			Results of the third block are from multiple recent papers, as noted in the second column.  
			When one method is reported in different papers and has different results, we keep the highest one only.   
		}
		\centering
		\setlength{\tabcolsep}{0.18pt} 
		\begin{tabular}{l|ccc}
			\toprule
			Method & \makecell[c]{Where result\\was compared} & \makecell{Full\\ training data:\\ label-imbalanced~~} & \makecell[c]{Sampled  \\training data:\\~label-balanced} \\
			
			\midrule
			
			S2E & \cite{xia2022sample} & 68.03 & --\\
			MentorNet & \cite{xia2022sample} & 67.25 & --\\
			SIGUA & \cite{xia2022sample} & 65.37 & --\\
			JoCoR & \cite{xia2022sample} & 69.06 & --\\

			CNLCU-S & \cite{xia2022sample} & 71.57 & --\\ 
			
			\midrule

			Reweight & \cite{jiang2022information} &  70.82 & --\\ 
			
			T-Revision & \cite{jiang2022information} &  71.27 & --\\ 
			
			PTD-F-V & \cite{jiang2022information} &  70.62 & --\\ 
			ELR & \cite{jiang2022information} &  71.86 & --\\
			IF-F-V & \cite{jiang2022information} &  72.29 & --\\

			\midrule

			MD+DYR+SH  &\cite{arazo2019unsupervised,li2020dividemix,wu2020topological}&  71.00 & --\\
			
			TVD  & \cite{zhang2021learningicml} & 71.65 &  -- \\
			
			Meta-Cleaner &\cite{zhang2019metacleaner,li2020dividemix} &  72.50* & --\\
			
			\makecell[l]{Meta-Learner} &\cite{li2019learning,li2020dividemix,zhang2021learning,wu2020topological}&  73.47 & --\\
			
			Joint-Optim &
			\makecell[c]{
				[\citenum{liu2020early,tanaka2018joint,li2019learning,li2020dividemix}, \\
				\citenum{wu2020topological,yi2019probabilistic,zhang2019metacleaner,zhang2021learning}]
			}
			&  72.23 & --\\

			kMEIDTM & \cite{cheng2022instance} & 73.34& -- \\ 

			P-correction  &\cite{yi2019probabilistic,li2020dividemix,zhang2021learning,wu2020topological}&  -- & 73.49 \\

			PLC  & \cite{zhang2021learning} & -- &  74.02 \\

			TopoFilter & \cite{wu2020topological} & -- & 74.10 \\


			VolMinNet & \cite{li2021provably} & -- & 72.42* \\
			
			HOC & \cite{zhu2021clusterability} & -- & 73.39* \\

			\makecell[l]{$\text{CORES}^2$+Mixup} & \cite{cheng2022instance,cheng2021learning,zhu2021second,zhu2021clusterability} & -- & 73.24* \\
			
			\makecell[l]{
				$\text{CORES}^2$+Mixup+CAL} & \cite{zhu2021second,cheng2022instance} & -- & 74.17 \\

			\midrule

			CE &  
			\makecell[c]{
				[\citenum{jiang2022information, chen2021noise,cheng2021learning}, \\
				\citenum{patrini2017making,wu2020topological,yi2019probabilistic},
				\\
				\citenum{zhang2019metacleaner,zhang2021learning,zhu2021second, zhu2021clusterability}]
			} 
			& 68.94 & 71.12 $\pm$ 0.32\\
			
			Forward &
			\makecell[c]{
				[\citenum{li2019learning,chen2021noise,jiang2022information}, \\
				\citenum{patrini2017making, li2021provably,wu2020topological}, \\
				\citenum{yi2019probabilistic,zhang2019metacleaner,zhang2021learning,zhu2021clusterability}]
			}
			& 69.84 & 71.28 $\pm$ 0.27\\
			
			Backward & \cite{patrini2017making,chen2021noise} & 69.13 & 71.03 $\pm$ 0.33\\
			
			Co-teaching & \cite{chen2021noise,zhu2021clusterability} & 70.19$\pm$0.28 & 72.14 $\pm$ 0.28\\
			
			DivideMix & \cite{chen2021noise}& -- & 73.81$\pm$0.41\\

			SLN &\cite{chen2021noise}& 70.42 $\pm$ 0.34 & 72.95 $\pm$ 0.31 \\
			
			SLN+MO &\cite{chen2021noise}& 71.15 $\pm$ 0.21 & 72.98 $\pm$ 0.15 \\
			
			SLN+MO+LC &\cite{chen2021noise}& 72.61 $\pm$ 0.23 & 74.08 $\pm$ 0.18 \\
			
			\midrule

			ProSelfLC & Ours & \textbf{73.93 $\pm$ 0.02} & \textbf{74.48 $\pm$ 0.01} \\
			
			\bottomrule
		\end{tabular}
		
		\label{table:Clothing1M_competitors}
	\end{table}

	\noindent
	\textbf{To redefine and reward a low-temperature entropy state.} 
	On the one hand, LS and CP work well, being consistent with prior claims. 
	In Fig.~\ref{fig:CIFAR100Asy0_4_mean_entropy_CleanSubset} and Fig.~\ref{fig:CIFAR100Asy0_4_mean_entropy_NoisySubset}, the entropies of both clean and noisy subsets are much higher in LS and CP, correspondingly their generalisation is the best except for ProSelfLC in Fig.~\ref{fig:CIFAR100Asy0_4_accuracy30000}. 
	On the other hand, ProSelfLC has the lowest entropy while performs the best, which proves that a learner's confidence does not necessarily weaken its generalisation performance. 
	%
	%
	Instead, {a model needs to be cautious about what to be confident in}. 
	According to Fig.~\ref{fig:CIFAR100Asy0_4_corruptedFittting} and Fig.~\ref{fig:CIFAR100Asy0_4_semanticCorrection}, ProSelfLC has the lowest wrong fitting and highest semantic class correction, which indicates that the learned model reaches a low-entropy target state redefined by corrected labels.

	\subsection{Outperform the state-of-the-art methods on real-world noisy Clothing1M and Food-101N}

	\textbf{Datasets.} 
	(1) \textit{Clothing1M} \cite{xiao2015learning} has about 1 million images of 14 fine-grained classes from shopping websites and around 38.46\% label noise in the training data. 
	Noisy labels are generated from description texts. 
	The exact noise structure is unknown. 
	In addition, Clothing1M is highly imbalanced, with images per label ranging from 18,976 to 88,588. 
	The 14 classes are fine-grained and challenging to classify. A recent work Class2Simi \cite{wu2021class2simi} merges two similar classes into one class. 
	To compare with the majority of algorithms under the same setting, we do not merge classes.   
	We train models only on the 1M noisy data and validate on clean test data. 
	(2) \textit{Food-101N} \cite{lee2018cleannet} has 101 fine-grained categories and contains 310K train images with noisy labels. The 25k curated test images are from the original Food-101 dataset \cite{bossard2014food}.
	In the train set, Food-101N \cite{lee2018cleannet} has 55k extra ``verification labels" (55k VLs) which annotate whether noisy labels are correct or not.
	Following the recent papers \cite{zhang2019metacleaner,zhang2021learning,cheng2022instance,han2019deep}, we train on the more noisy dataset Food-101N \cite{lee2018cleannet} and validate on the original clean test data \cite{bossard2014food}.  
	As a reference, although the recent robust early-learning method CDR \cite{xia2021robust} uses the less noisy Food-101 train data, its test accuracy is 86.36\% and lower than ours.

	\noindent
	\textbf{Competitors.} 
	%
	We compare with (1) sample selection methods including 
	Search to Exploit (S2E) \cite{yao2020searching}, CNLCU-S \cite{xia2022sample}, TopoFilter \cite{wu2020topological}, COnfidence REgularized Sample Sieve ($\text{CORES}^2$) \cite{cheng2021learning};  
	%
	(2) Noise transition matrix estimation algorithms including Reweight \cite{liu2015classification}, T-Revision \cite{xia2019anchor}, PTD-F-V \cite{xia2020part}, IF-F-V \cite{jiang2022information}, kMEIDTM \cite{cheng2022instance}, TVD \cite{zhang2021learningicml}, VolMinNet \cite{li2021provably}, High-Order-Consensus (HOC) \cite{zhu2021clusterability};
	(3) Co-training methods, i.e., JoCoR \cite{wei2020combating} and Co-teaching \cite{han2018co}; 
	(4) Meta-learning approaches, i.e.,  Meta-Cleaner \cite{zhang2019metacleaner} and Meta-Learner (MLNT) \cite{li2019learning}; 
	(5) Iterative label correction methods including 
	Joint-Optim \cite{tanaka2018joint}, P-correction \cite{yi2019probabilistic} and PLC \cite{zhang2021learning}; 
	(6) Example weighting algorithms, i.e., MentorNet \cite{jiang2018mentornet} and  
	SIGUA \cite{han2020sigua};  
	%
	%
	%
	%
	%
	(7) Compounded methods (linked by +):
	MD+DYR+SH \cite{arazo2019unsupervised}  comprises three techniques, i.e., dynamic mixup (MD), dynamic bootstrapping together with label regularisation (DYR) and soft to hard (SH). 
	MD+DYR+SH
	models sample loss with BMM \cite{ma2011bayesian} and applies Mixup \cite{zhang2018mixup}.
	%
	%
	%
	%
	%
	%
	%
	%
	%
	%
	Covariance-Assisted Learning (CAL) \cite{zhu2021second} exploits $\text{CORES}^2$ \cite{cheng2021learning} and Mixup.
	In the recent work \cite{chen2021noise}, stochastic label noise (SLN) is proposed to regularize the training. SLN can be combined with Momentum (MO), which updates a model's parameters in a moving average way.  
	DivideMix \cite{li2020dividemix} combines a set of strategies, including semi-supervised learning following dataset division and networks co-training.
	DivideMix uses a different training schedule.
	Therefore, for a fair comparison, we compare with the DivideMix recently reproduced by SLN \cite{chen2021noise}. 
	ELR \cite{liu2020early} denotes early-learning regularization. 
	%
	%
	%
	%
	%
	The recent ILFC \cite{berthon2021confidence} is not compared because it uses an extra clean dataset and trains ResNet18. Nested co-teaching \cite{chen2021boosting} trains ResNet18 and uses a very large batch size of 448. 
	DAT \cite{qu2021dat} trains ResNet50 but uses a batch size of 256. 
	UniCon \cite{karim2022unicon} integrates Auto-augment \cite{cubuk2019autoaugment}, contrastive learning and semi-supervised training, thus having slightly better results than us.   
	The other methods have been introduced heretofore. 

	\noindent
	\textbf{Experimental details.} 
	On both datasets, the ResNet-50 is pretrained on ImageNet and publicly available in PyTorch \cite{paszke2019pytorch}. 
	For Clothing1M, we follow the recent settings in \cite{chen2021noise,zhang2021learning,zhu2021second} and use a small batch size of 32. 
	%
	The other training details are similar to Section~\ref{sec:clean_setting} with small changes: we start with a learning rate of 0.01 and use a weight decay of 0.02.    
	They are chosen according to the separate clean validation set.
	For Food-101N, we follow the same settings as the recent work \cite{zhang2021learning}. 
	The batch size is 128 and we train 72k iterations. 
	We report the mean and standard deviation results of three random trials as in \cite{chen2021noise,zhang2021learning}.
	%
	%

	\noindent
	\textbf{Results analysis.} In Table~\ref{table:Clothing1M_competitors} of Clothing1M results, for both label-imbalanced data and label-balanced data, 
	ProSelfLC has the highest accuracy, which demonstrates its effectiveness against real-world asymmetric and instance-dependent label noise. 
	In Table~\ref{table:food101n_competitors}, the results of Food-101N confirm again that ProSelfLC is superior to existing algorithms. 
	We remark that Clothing1M and Food-101N contain fine-grained categories, thus being challenging. ProSelfLC obtains the state-of-the-art performance on both.

	\begin{table}[!t]
		\caption{
			Test accuracy (\%) on the real-world noisy dataset Food-101N, composed of 101 fine-grained food classes.
		}
		\vspace{-0.2cm}
		\centering
		\setlength{\tabcolsep}{1.8pt} 
		\begin{tabular}{lcccc}
			\toprule
			Method & \makecell[c]{Where result\\was compared} & Train Data & Accuracy \\
			
			\midrule

			
			CCE & \cite{lee2018cleannet,zhang2019metacleaner,cheng2022instance} & Food-101N & 81.44 \\
			
			CleanNet ($w_{hard}$) &
			\makecell[c]{
				[\citenum{cheng2022instance,han2019deep}, \\
				\citenum{lee2018cleannet,zhang2019metacleaner}]
			} & Food-101N+55k VLs & 83.47\\
			CleanNet ($w_{soft}$) &
			\makecell[c]{
				[\citenum{cheng2022instance,han2019deep}, \\
				\citenum{lee2018cleannet,zhang2019metacleaner,zhang2021learning}]
			} & Food-101N+55k VLs & 83.95 \\
			
			Meta-Cleaner &\cite{zhang2019metacleaner} & Food-101N & 85.05 &\\
			
			\makecell[l]{Multi-Prototypes} & \cite{han2019deep,cheng2022instance} & Food-101N & 85.11\\
			
			PLC &\cite{zhang2021learning} & Food-101N & 85.28±0.04\\
			
			
			DivideMix & \cite{cheng2022instance} & Food-101N & 84.39 \\
			kMEIDTM+DivideMix & \cite{cheng2022instance} & Food-101N & 85.61 \\ 
			
			\midrule
			
			ProSelfLC & Ours &Food-101N& \textbf{86.62}±0.01\\
			\bottomrule
		\end{tabular}
		\label{table:food101n_competitors}
	\end{table}

	\begin{table}[!t]
		\caption{
			{
				Results of robust protein classification against cropping noise and symmetric label noise. 
				For comprehensiveness, we report accuracy and confidence metrics on both train and test sets. We show two confidence metrics: $\mathrm{conf}_{\mathrm{all}}$ and $\mathrm{conf}_{\mathrm{top}}$. 
				We use the final model when training ends without selecting the intermediate best ones.      
				The highest accuracy or confidence in each row is bolded. 
			}
		}
		\centering
		\setlength{\tabcolsep}{3.50pt} 
		\begin{tabular}{lllccccccc}
			\toprule
			\multirow{1}{*}{Noise type} & 
			\multirow{1}{*}{Metric}
			&
			\multirow{1}{*}{Data}
			&\multirow{1}{*}{CCE} 
			& 
			\multirow{1}{*}{LS} 
			&   \multicolumn{1}{c}{CP } &  
			\multicolumn{1}{c}{Boot-soft } &  
			\multicolumn{1}{c}{ProSelfLC }
			\\
			\midrule

			\multirow{9}{*}{\makecell[l]{Cropping \\noise}} & \multirow{2}{*}{Accuracy} &  Test & 88.7&
			90.4&
			90.9&
			90.4&
			\textbf{92.4}
			\\
			
			&& Train & 
			\textbf{99.3}&
			\textbf{99.3}&
			98.8&
			95.9&
			91.8
			\\

			\vspace{-0.08cm}
			&&&&\\
			\cdashline{2-8}
			\vspace{-0.08cm}
			&&&&\\
			
			&\multirow{2}{*}{
				$\mathrm{conf}_{\mathrm{all}}$
			} & Test & 
			86.9&
			59.5&
			83.8&
			89.0&
			\textbf{96.4} 
			\\
			
			&& Train & 
			89.0&
			63.3&
			82.0&
			92.6&
			\textbf{97.6}
			\\

			\vspace{-0.08cm}
			&&&&\\
			\cdashline{2-8}
			\vspace{-0.08cm}
			&&&&\\
			
			&\multirow{2}{*}{$\mathrm{conf}_{\mathrm{top}}$} & Test & 
			96.4&
			90.1&
			95.5&
			96.8&
			\textbf{99.2}
			\\
			
			&& Train &
			97.6&
			92.2&
			95.7&
			98.5&
			\textbf{99.7}
			\\

			\\
			\midrule
			\\		
			
			\multirow{9}{*}{
				\makecell[l]{Cropping\\ noise\\+\\Label \\noise}} & \multirow{2}{*}{Accuracy} &  Test & 
			84.2&
			89.4&
			89.4&
			89.6&
			\textbf{92.2}
			\\
			
			&& Train &
			\textbf{84.8}&
			75.2&
			72.8&
			77.2&
			{75.7}
			\\

			\vspace{-0.08cm}
			&&&&\\
			\cdashline{2-8}
			\vspace{-0.08cm}
			&&&&\\
			
			&\multirow{2}{*}{
				$\mathrm{conf}_{\mathrm{all}}$
			} & Test &
			65.3&
			38.7&
			37.4&
			57.6&
			\textbf{96.1}
			\\
			
			&& Train &
			42.0&
			20.0&
			18.0&
			31.3&
			\textbf{95.2}
			\\

			\vspace{-0.08cm}
			&&&&\\
			\cdashline{2-8}
			\vspace{-0.08cm}
			&&&&\\
			
			&\multirow{2}{*}{
				$\mathrm{conf}_{\mathrm{top}}$
			} & Test &
			88.9&
			80.4&
			78.4&
			86.0&
			\textbf{99.2}
			\\
			
			&& Train &
			80.7&
			70.0&
			67.5&
			74.9&
			\textbf{99.1}
			\\		
			
			%
			
			%

			
			%
			%
			%
			%
			%
			%
			%
			%
			%
			%
			%
			%
			%
			%
			%
			%
			%
			
			%
			%
			%
			%
			%
			%
			%
			
			\bottomrule
		\end{tabular}
		\label{table:prottrans_DeepLocMS}
	\end{table}

	\begin{table*}[!t]
		\caption{
			{
				Results of target modification methods without/with annealed temperature (AT) for curating the target-state entropy. We train models on CIFAR-100 whose training labels contain symmetric noise. 
				We do not select the intermediate best models and report the generalisation accuracy and confidence metrics on the clean test set  when training terminates. 
				There are two confidence measurements: $\mathrm{conf}_{\mathrm{all}}$ and $\mathrm{conf}_{\mathrm{top}}$.
				The highest accuracy or confidence of each row is bolded.  
				%
			}
		}
		\centering
		\fontsize{7.9pt}{7.9pt}\selectfont
		
		\begin{tabular}{lllcccccccccc}
			\toprule
			\multirow{2}{*}{Noise rate} & 
			\multirow{2}{*}{Network} & 
			&\multirow{2}{*}{CCE} & 
			\multicolumn{3}{c}{LS } &   \multicolumn{2}{c}{CP } &  
			\multicolumn{2}{c}{Boot-soft} &  
			\multicolumn{2}{c}{ProSelfLC}
			\\
			\cmidrule(lr){5-7}
			\cmidrule(lr){8-9}
			\cmidrule(lr){10-11}
			\cmidrule(lr){12-13}
			&
			&
			& & 0.125 & 0.25 & 0.50 
			& without AT & with AT  
			& without AT & with AT  
			& without AT & with AT  
			\\
			\midrule

			\multirow{12.5}{*}{20\%} & 
			\multirow{3}{*}{ShuffleNetV2} & \makecell[l]{
				Accuracy} & 
			60.5&
			61.1&
			64.0&
			65.3&
			62.4&
			61.7&
			61.9&
			64.4&
			65.6&
			\textbf{66.8}
			\\
			
			& & \makecell[l]{
				 $\mathrm{conf}_{\mathrm{all}}$} & 
			64.7 &
			49.7 & 
			39.4 & 
			22.3&
			58.6&
			51.7&
			59.0&
			73.2&
			44.6&
			\textbf{86.4}
			\\
			
			& & \makecell[l]{
				 $\mathrm{conf}_{\mathrm{top}}$} & 
			62.8&
			52.3&
			45.5&
			31.5&
			59.6&
			53.9&
			59.9&
			73.8&
			47.7&
			\textbf{84.8}
			\\

			\vspace{-0.08cm}
			&&&&\\
			\cdashline{2-13}
			\vspace{-0.08cm}
			&&&&\\

			& 
			\multirow{3}{*}{ResNet18} & \makecell[l]{Accuracy} & 
			61.9 & 63.0 & 65.3 & 68.4 & 66.1 & 65.9 & 65.1 & 68.5 & 71.4 & \textbf{72.8}\\
			
			& & \makecell[l]{
				 $\mathrm{conf}_{\mathrm{all}}$} & 
			63.9 & 
			44.2 & 
			33.7 &
			20.0 &
			56.3 &
			49.7&
			56.7&
			71.3&
			21.0&
			\textbf{81.7}
			\\
			
			& & \makecell[l]{
				 $\mathrm{conf}_{\mathrm{top}}$} & 
			63.6&
			48.8&
			40.5&
			29.1&
			59.4&
			54.0&
			59.7&
			73.2&
			27.9&
			\textbf{82.4}
			\\
			

			%
			%
			
			\midrule

			\multirow{13}{*}{40\%} & 
			\multirow{3}{*}{ShuffleNetV2} & \makecell[l]{Accuracy} & 
			49.6
			&52.2
			&54.1
			&56.1
			&54.0
			&53.7
			&53.4
			&58.9
			&59.7
			&\textbf{61.6}
			\\
			
			& & \makecell[l]{
				 $\mathrm{conf}_{\mathrm{all}}$} & 
			49.1
			&37.5
			&29.4
			&15.1
			&41.2
			&34.8
			&41.5
			&59.9
			&21.6
			&\textbf{84.1}
			\\
			
			& & \makecell[l]{
				 $\mathrm{conf}_{\mathrm{top}}$} & 
			47.6&
			39.7&
			34.4&
			22.3&
			43.8&
			38.1&
			43.8&
			62.9&
			26.9&
			\textbf{83.3}
			\\
			
			\vspace{-0.08cm}
			&&&&\\
			\cdashline{2-13}
			\vspace{-0.08cm}
			&&&&\\

			& 
			\multirow{3}{*}{ResNet18} & \makecell[l]{Accuracy} & 
			47.6
			&49.0
			&51.7
			&57.6
			&54.1
			&54.0
			&53.9
			&61.1
			&65.7
			&\textbf{68.4}
			\\
			
			& & \makecell[l]{
				 $\mathrm{conf}_{\mathrm{all}}$} & 
			49.7
			&34.2
			&25.0
			&13.1
			&38.4
			&19.1
			&39.0
			&59.6
			&9.6
			&\textbf{79.7}
			\\
			
			& & \makecell[l]{
				 $\mathrm{conf}_{\mathrm{top}}$} & 
			48.7&
			36.8&
			30.1&
			19.9&
			42.3&
			24.5&
			42.8&
			63.1&
			15.6&
			\textbf{80.7}
			\\

			\bottomrule
		\end{tabular}
		\label{table:without_or_with_AT_confidence_calibration}
	\end{table*}

	\subsection{Train robust transformers on noisy protein classification datasets}
	
	We follow the recent ProtTrans to do experiments on protein classification \cite{elnaggar2021prottrans}.
	Training deep models to predict a protein's properties is challenging as the length of amino acid sequences varies from several tens to multiple thousands \cite{almagro2017deeploc,elnaggar2021prottrans,rives2021biological}. 
	Some recent approaches crop amino acid sequences to decrease training time and GPU memory consumption \cite{almagro2017deeploc,elnaggar2021prottrans,rives2021biological}.   
	In this work, we find that cropping input sequences adds noise to model training as some proteins have important functional regions interspersed across the protein length \cite{almagro2017deeploc}. We empirically demonstrate this by training models on the cropped proteins. 
	In addition to cropping noise, we further design high-noise experiments by including unlabelled proteins. Conceptually, it is semi-supervised learning. We bridge semi-supervised learning and label-noise learning by assigning random labels to those unlabelled proteins, so that we establish the synthetic dataset to validate ProSelfLC for training robust protein transformers against label noise.         
	
	\noindent
	\textbf{Datasets.} The DeepLoc train set used in \cite{elnaggar2021prottrans} contains 6,622 proteins that are annotated to be membrane (i.e., they are found on the membrane), water-soluble (i.e., they are from the lumen of the organelle), or unknown (missing information about where they are found) \cite{almagro2017deeploc}. 
	Specifically, there are 1,518 transmembrane proteins, 2,227 water-soluble proteins and 2,877 proteins with unknown labels.  
	There are 1,842 proteins in total and 1,087 proteins with known labels in the test set. 
	We present the cropping noise and synthetic symmetric label noise as follows: 
	\begin{itemize}
		\item \textit{Cropping noise}. We train on proteins with known labels. 
		Their length ranges from 40 to 13,100, with a median of 434.
		Therefore, we truncate proteins longer than 434 at the end so that all amino acid sequences have a length no longer than 434. 
		The goal of this setting is to validate whether ProSelfLC could be robust to cropping noise if there is a practical need to crop sequences for speeding up training and reducing GPU memory requirement.   
		\item \textit{Cropping noise+}\textit{Label noise}: 
		First, we keep the cropping noise as we crop proteins to decrease training time and reduce GPU memory requirement. 
		Second, we add symmetric label noise by assigning uniform random labels to 2,877 unlabelled proteins. 
		Cropping noise together with label noise makes the noise level high.
		The objective is to evaluate whether ProSelfLC could be robust to severe noise when it is expensive to remove it in practice.
	\end{itemize}
	

	\begin{figure*}[!t]
		\centering
		\begin{subfigure}[!h]{0.239\textwidth}
			\centering
			\captionsetup{width=\textwidth}
			\includegraphics[clip, trim=3.30cm 1.2cm 3.6cm 2.3cm, width=0.99\textwidth]{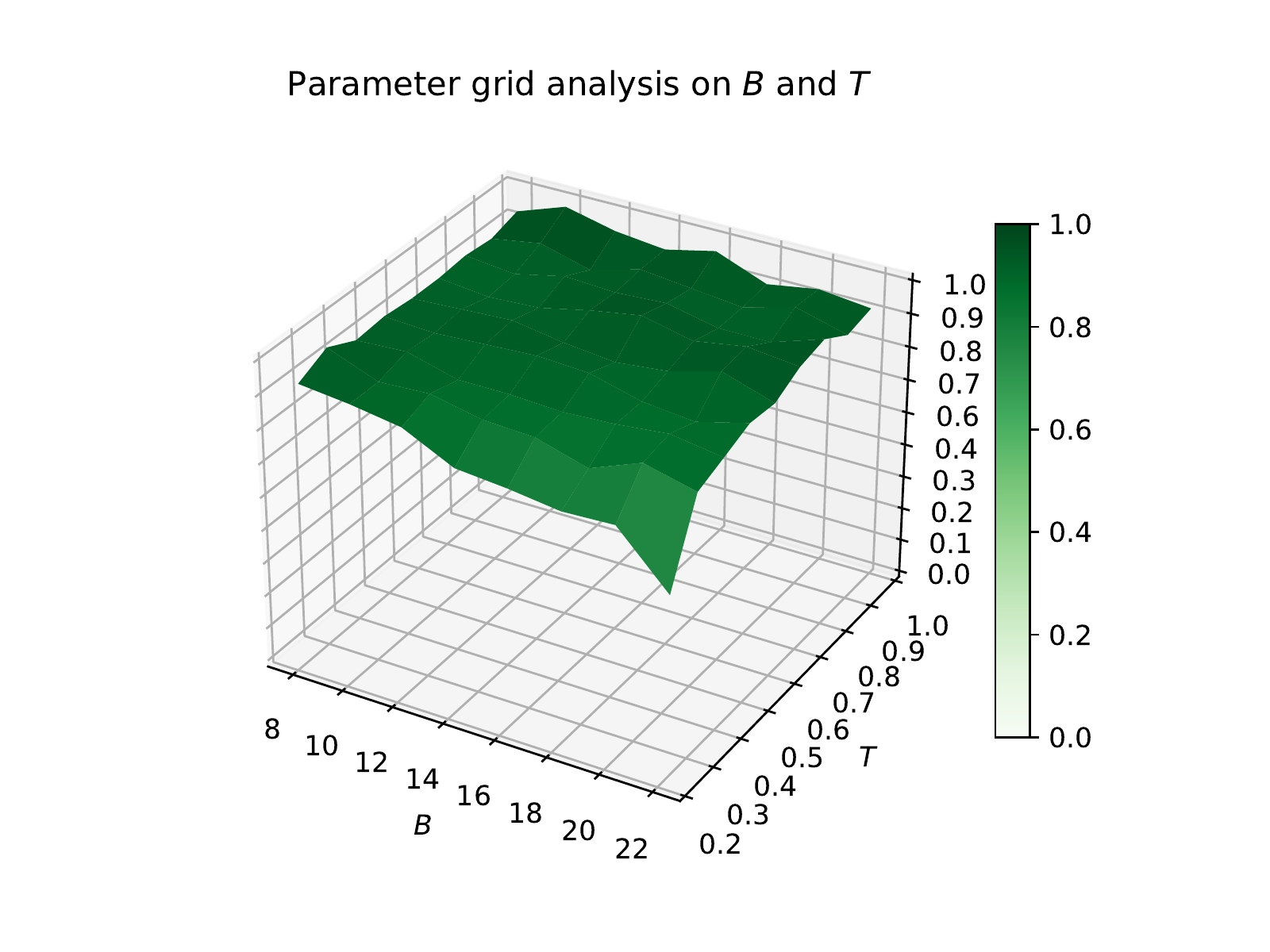}
			\caption{ShuffleNetV2: $r=0$.}
			\label{fig:0.0_shufflenetv2}
		\end{subfigure}
		\begin{subfigure}[!h]{0.239\textwidth}
			\centering
			\captionsetup{width=\textwidth}
			\includegraphics[clip, trim=3.30cm 1.2cm 3.6cm 2.3cm, width=0.99\textwidth]{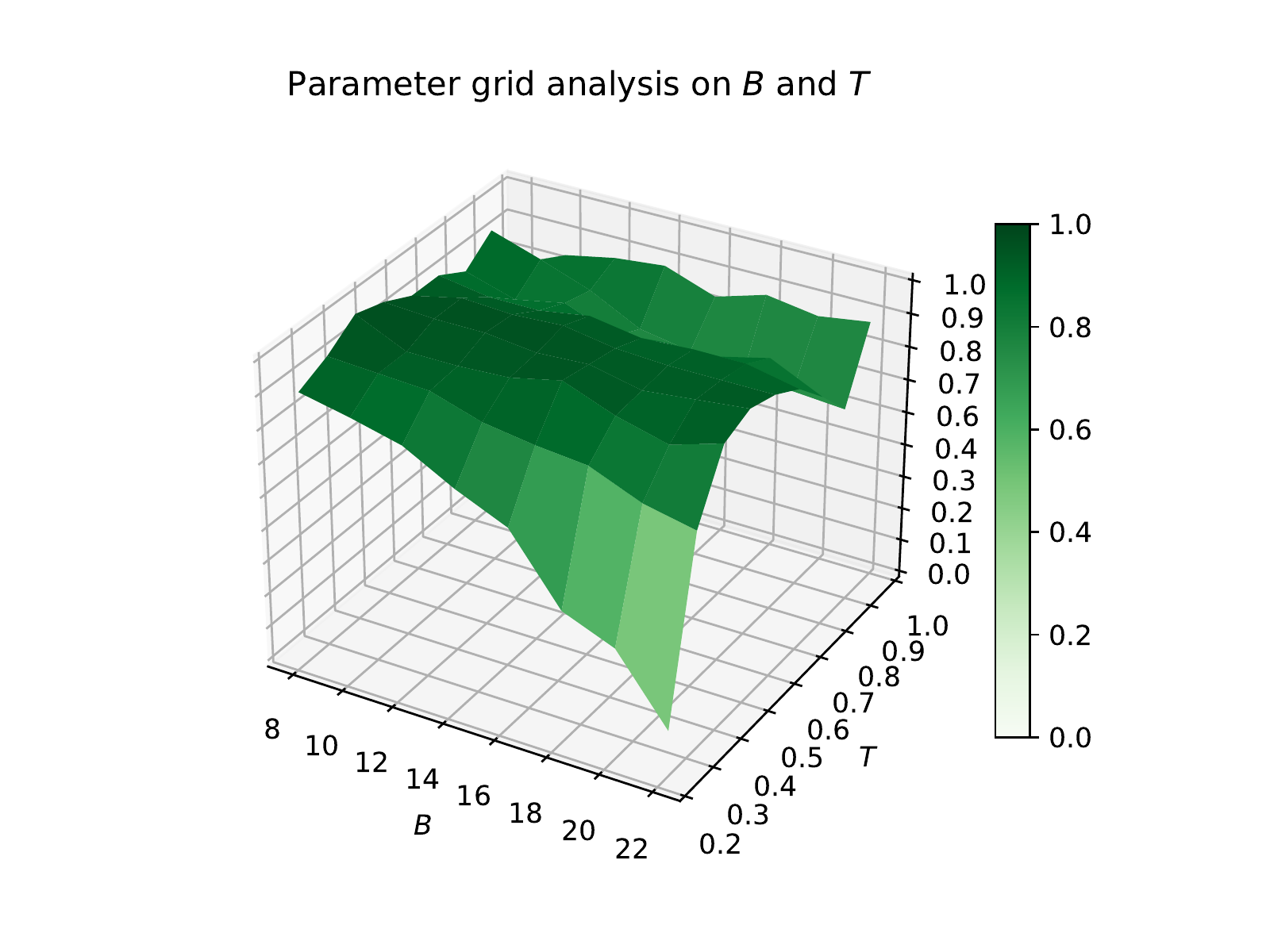}
			\caption{ShuffleNetV2: $r=20\%$.}
			\label{fig:0.2_shufflenetv2}
		\end{subfigure}
		\begin{subfigure}[!h]{0.239\textwidth}
			\centering
			\captionsetup{width=\textwidth}
			\includegraphics[clip, trim=3.30cm 1.2cm 3.6cm 2.3cm, width=0.99\textwidth]{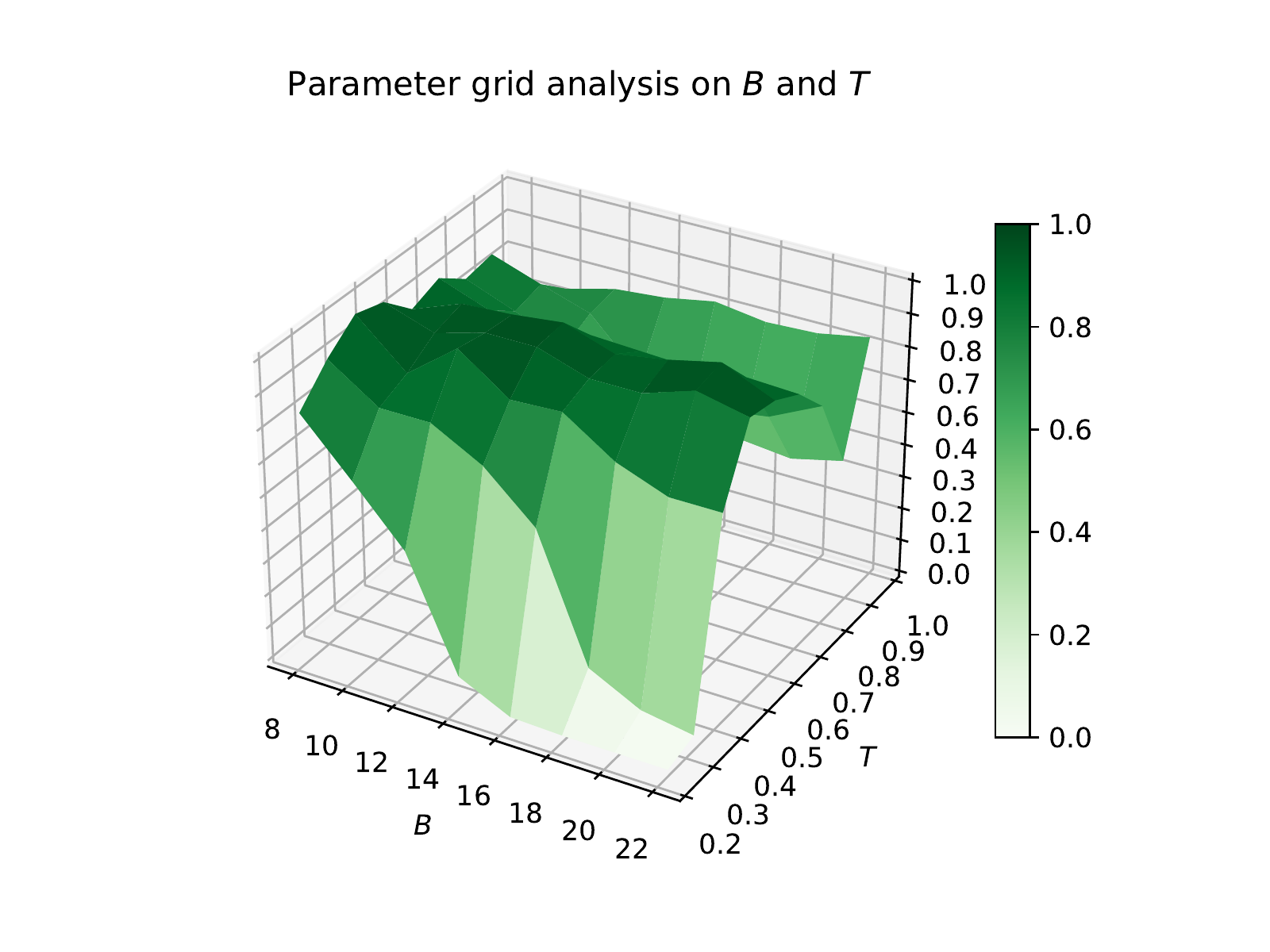}
			\caption{ShuffleNetV2: $r=40\%$.}
			\label{fig:0.0_resnet34}
		\end{subfigure}
		\begin{subfigure}[!h]{0.268\textwidth}
			\centering
			\captionsetup{width=\textwidth}
			\includegraphics[clip, trim=3.30cm 1.2cm 2.1cm 2.3cm, width=0.99\textwidth]{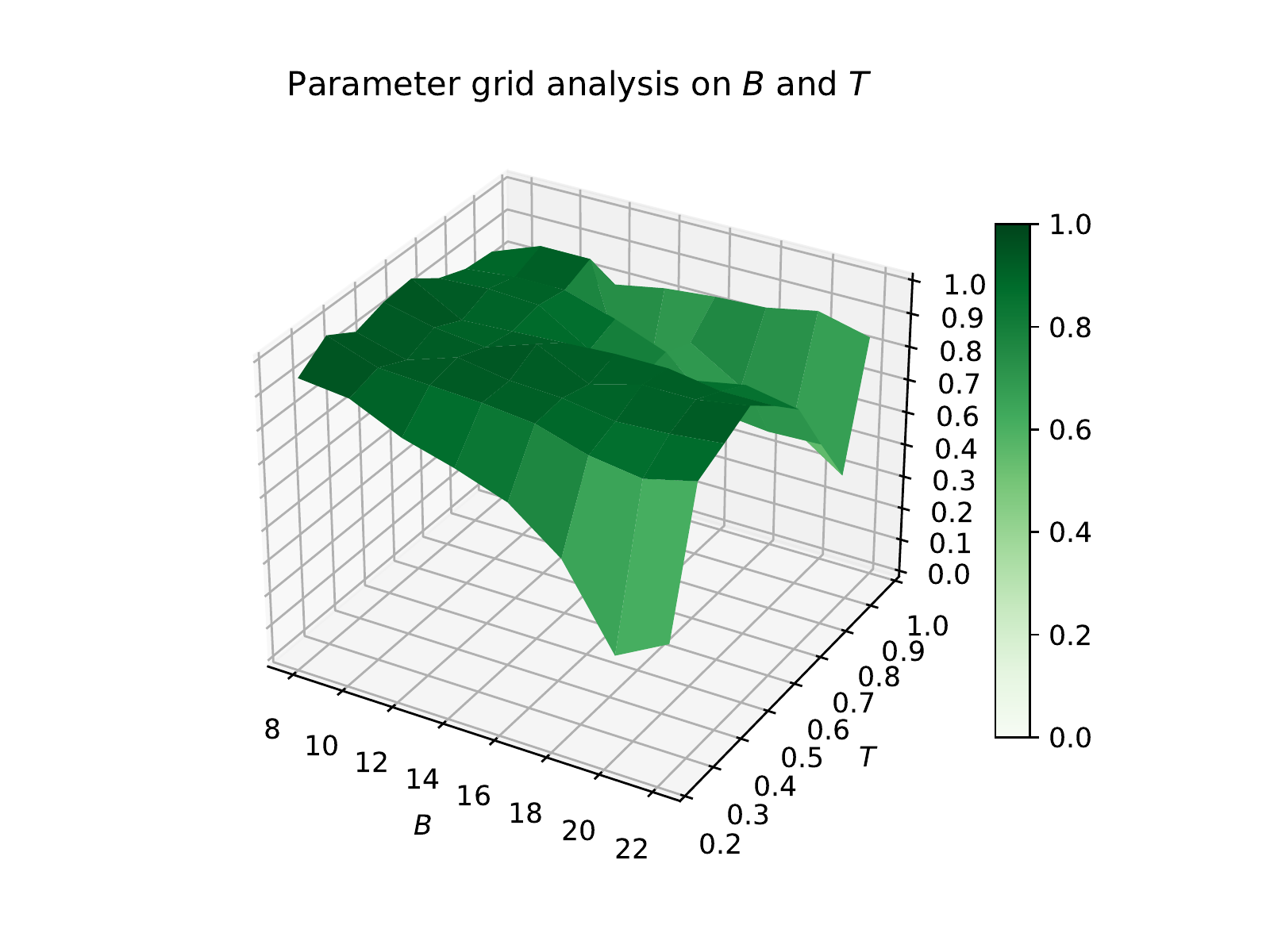}
			\caption{ResNet18: $r=20\%$.}
			\label{fig:0.2_resnet34}
		\end{subfigure}
		%
		\caption{ 
			Grid analysis on the hyper-parameters space of $B$ and $T$. We use CIFAR-100 with a symmetric label noise rate of $r$. For a clearer visualization, in each subfigure, all results are plotted in an exponential ratio to the best one. 
			%
		}
		\label{fig:grid_analysis_B_T}
	\end{figure*}

	\begin{figure*}[!t]
		\centering
		\begin{subfigure}[!h]{0.50\textwidth}
			\centering
			\captionsetup{width=\textwidth}
			\includegraphics[clip, trim=1.71cm 9.1cm 1.30cm 0.06cm, width=0.99\textwidth]{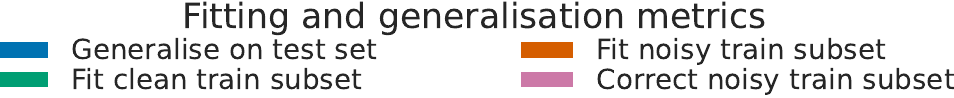}
		\end{subfigure}\\
		~\\
		\vspace*{-0.32cm}
		~\\
		\begin{subfigure}[!h]{0.33\textwidth}
			\centering
			\captionsetup{width=\textwidth}
			\includegraphics[clip, trim=0.68cm 0.0cm 0.58cm 1.09cm, width=0.99\textwidth]{trust_schemes/cifar100_symmetric_noise_rate_0.2_shufflenetv2_merged_metrics_with_paramssort_lossname_accu_withoutAT-False_None_None}
			\caption{ShuffleNetV2: $r$=20\%.}
			\label{fig:}
		\end{subfigure}
		\begin{subfigure}[!h]{0.329\textwidth}
			\centering
			\captionsetup{width=\textwidth}
			\includegraphics[clip, trim=0.68cm 0.0cm 0.58cm 1.09cm, width=0.99\textwidth]{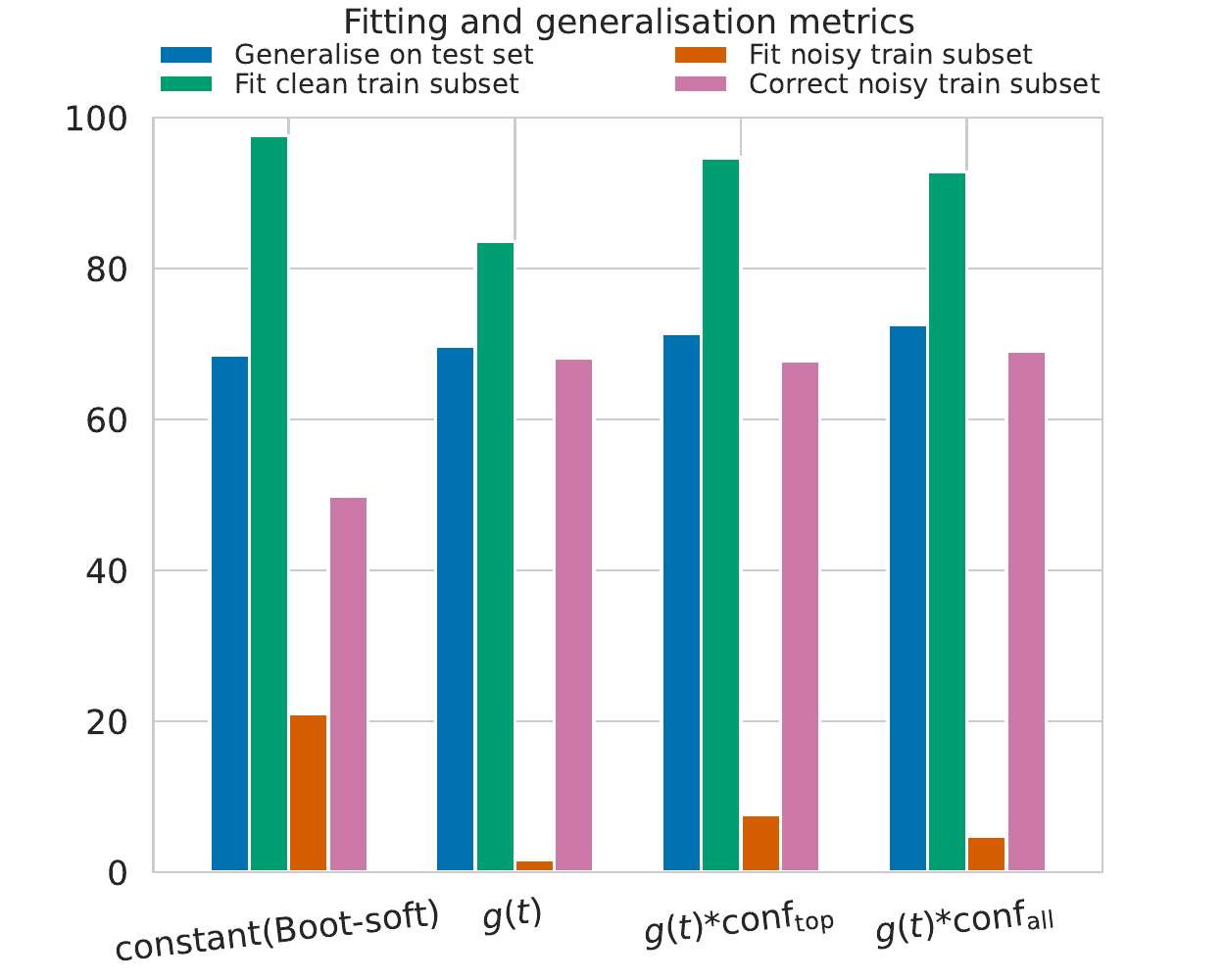}
			\caption{ResNet18: $r$=20\%.}
			\label{fig:}
		\end{subfigure}
		\begin{subfigure}[!h]{0.329\textwidth}
			\centering
			\captionsetup{width=\textwidth}
			\includegraphics[clip, trim=0.68cm 0.0cm 0.58cm 1.09cm, width=0.99\textwidth]{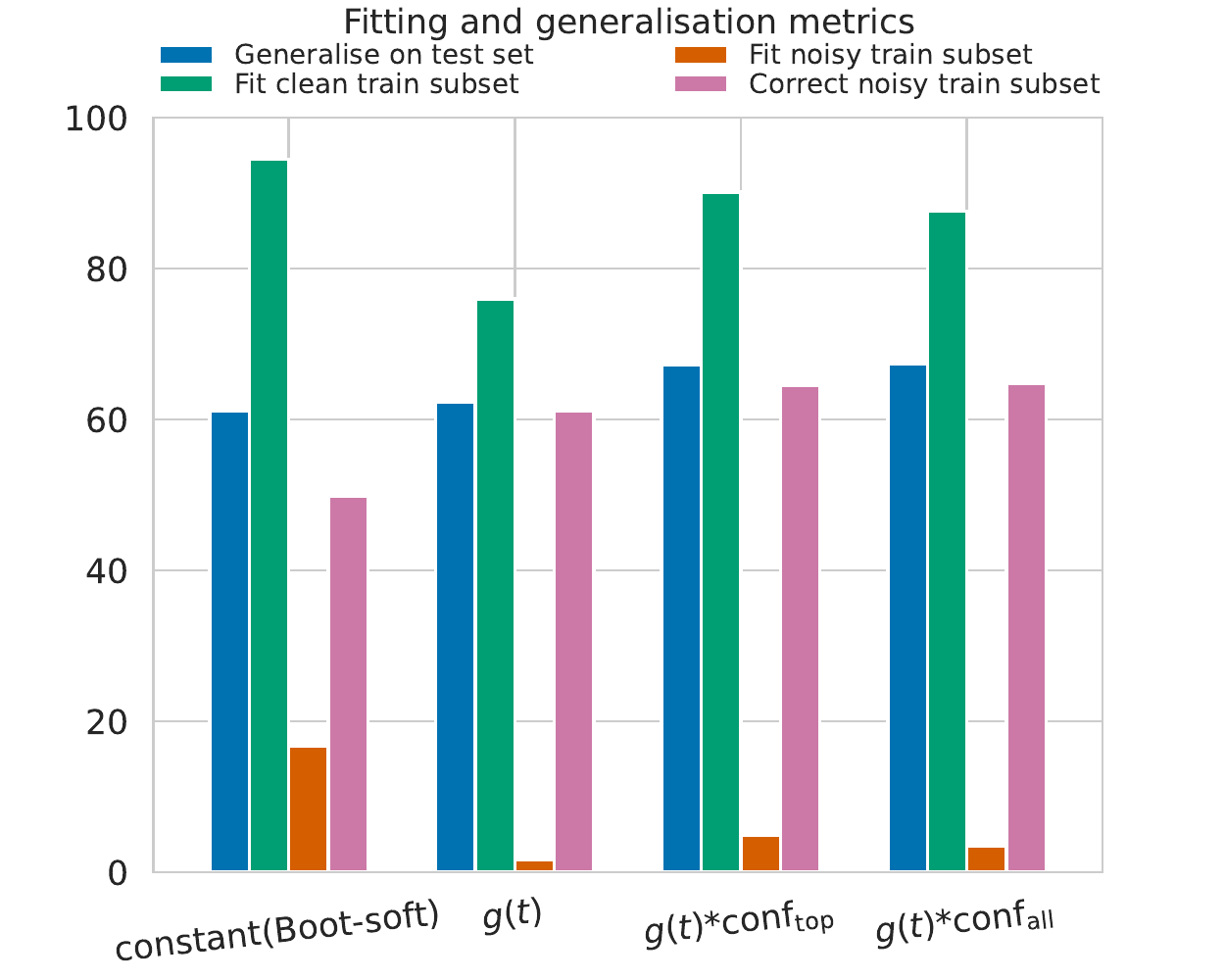}
			\caption{ResNet18: $r$=40\%.}
			\label{fig:AblationNoEntropyCIFAR100Asy0_4_accuracy30000}
		\end{subfigure}

		\vspace{-0.28cm}
		\caption{
			Results of computing $\epsilon_{\mathrm{ProSelfLC}}$ by four self trust schemes, denoted in the horizontal axis.   
			Experiments are done on CIFAR-100 with symmetric label noise. 
			A subfigure's caption describes the used network and noise rate $r$.   
			We report four metrics (\%) and one coloured bar per metric along the vertical axis.   
			\textit{For fitting the noisy train subset, a lower value is better. 
				For the other three metrics, a higher value is better}.    
			According to the finding of AT's effectiveness for boosting Boot-soft and ProSelfLC in Table~\ref{table:without_or_with_AT_confidence_calibration},  
			\textit{we add AT on top of every self trust scheme}. Therefore, we observe all schemes have competitive results and their performance gaps become smaller. 
			{{During training, a learner is not given whether a label is noisy or not.}} We use the final model when training ends. 
		}
		\label{fig:four_self_trust_schemes}
		\vspace{-0.25cm}
	\end{figure*}

	\noindent
	\textbf{Network and training details.} 
	We train sequence transformers to 
	classify a single amino acid sequence (without using homology information at all) to be either transmembrane or water-soluble. 
	The transformer network is a subnet of ProtBert-BFD \cite{elnaggar2021prottrans}, a protein language model pretrained on BFD-100 dataset \cite{steinegger2019protein, steinegger2018clustering}. 
	According to \cite{elnaggar2021prottrans}, to use a larger batch size, ProtBert-BFD is first trained on sequences with a maximum length of 512, then tuned on sequences with a maximum length of 2k.   
	We name this subnet ProtBert-H16-D6, where the D6 denotes its depth is 6 (D6), i.e., a stack of 6 hidden transformer layers. H16 means that in each transformer layer, the number of transformer blocks (a.k.a., heads) is 16.   
	The released model has a depth of 30, so that our subnet is 5 times shallower.
	We choose to train this subnet mainly because it requires a small-memory GPU and is faster to train. In addition, it benefits little from pretraining, so it indicates our algorithm can be applied to train sequence transformers from scratch. 
	Finally, we will release this subnet and make it convenient to reproduce our results with a 16GB GPU machine. 
	We use a batch size of 32 and the SGD optimiser. The weight decay is 0.0001. For cropping noise, a starting learning rate of 0.02 is used. When noise rate is high, i.e., cropping noise+label noise, we use a smaller starting learning rate of 0.01. We train 40 epochs in total.   
	We stress that the generic hyperparameters are coarsely searched by visualizing the statistical training curves, i.e., without brute-force and confidently fitting all training data as noise exists.   
	Besides, we report the metrics of the final model when training terminates other than select the best intermediate model, leaving our reported metrics less biased. 
	

	\noindent
	\textbf{Results analysis and discussion.} We present the accuracy and confidence metrics of ProSelfLC and baseline algorithms in Table~\ref{table:prottrans_DeepLocMS}. ProSelfLC's fitting of the noisy train set is much lower than that of CCE, which indicates that ProSelf does not overfit the training set. 
	This observation is obvious for both noise types. 
	Furthermore, ProSelfLC learns and generalises better and more confidently compared with other widely used baselines. 
	Our experiments demonstrate that the misleading effect by cropping noise can be alleviated by ProSelfLC, as ProSelfLC's performance (92.4\%) is even slightly better than the DeepLoc ensemble model (92.3\%) \cite{almagro2017deeploc} and the large transformer model (91.0\%) \cite{elnaggar2021prottrans}.  
	With the high label noise added, though 2,877 out of 6,622 proteins have random labels, the generalisation performance of ProSelfLC decreases little from 92.4\% to 92.2\%, which confirms that ProSelfLC can be a robust solution when it is expensive to remove severe noise in practice.
	
	
	

	\subsection{Ablation studies}
	\label{sec:ablation_study}
	
	
	\textbf{Normal-temperature entropy state versus low-temperature entropy state}. 
	For CP, Boot-soft and our ProSelfLC, the target state's entropy can be adjusted by the temperature. 
	We denote annealed temperature by AT. 
	We study the normal-temperature state (i.e., without AT) versus the low-temperature state (i.e., with AT) and display their results in Table~\ref{table:without_or_with_AT_confidence_calibration}. 
	Generic training parameters are the same for all methods. 
	We observe: 
	(1) Compared with the baseline CCE, the confidence-penalty approaches (LS and CP) indeed learn better and lower-confidence models, which is consistent with the
	motivations of proposing them; 
	(2) However, confidence-reward algorithms (Boot-soft and ProSelfLC) can perform better. ProSelfLC with AT generalises the best with the highest confidence, i.e., the lowest entropy; 
	(3) Across networks and noise rates, CP is less sensitive to target-state entropy while Boot-soft is the most sensitive. This demonstrates that target-state entropy is also very important for standard label correction (i.e., Boot-soft). For ProSelfLC, AT improves the performance consistently and reaches a curated low-temperature entropy state. Therefore, for both Boot-soft and ProSelfLC, we use AT by default in other experiments.    
	
	\noindent
	\textbf{Hyper-parameters space of $B$ and $T$}. 
	To be more comprehensive, 
	we do experiments on CIFAR-100 without/with symmetric label noise. Additionally, we train both ShuffleNetV2 and ResNet18. 
	In those experiments, we only change $B$ and $T$. All other training parameters are the same. We use the model when training ends without selecting the intermediate ones.
	According to Fig.~\ref{fig:grid_analysis_B_T}, 
	on both networks and both noise rates, generally, the performance is more sensitive to $T$ when $B$ is large. This confirms a human's intuitive concept that if we trust a learner itself at a faster speed, the confidence adjustment of this learner's predictions becomes more crucial.  
	When the noise rate increases, better results can be obtained by using a relatively smaller $T$ to optimise the model towards a low-temperature entropy state.

	\noindent
	\textbf{Self trust schemes}. 
	We study the differences of four self trust schemes described in Section~\ref{section:proselflc}. 
	When $\epsilon_{\mathrm{ProSelfLC}}$ is constant at training, it degrades to Boot-soft.  
	According to Fig.~\ref{fig:four_self_trust_schemes}, we observe that (1) Compared with ``constant'', $\mathrm{g}( t)$ outperforms Boot-soft in three metrics except for sacrificing fitting clean train subset a lot. 
	(2) Compared with ``constant'' and $\mathrm{g}( t)$,  $\mathrm{g}( t)*\mathrm{conf}_{\mathrm{all}}$ and $\mathrm{g}( t)*\mathrm{conf}_{\mathrm{top}}$ are better in balancing fitting and generalisation. 
	By default, in all other experiments, we use $\mathrm{g}( t)*\mathrm{conf}_{\mathrm{all}}$ due to its slightly better results.

	We further discuss post-training model calibration \cite{guo2017calibration} in Appendix \ref{appendix:error_and_model_calibration}, and the changes of entropy and $\epsilon_{\mathrm{ProSelfLC}}$ during training in Appendix \ref{appendix:entropy_and_coef_dynamics}.

	\vspace{-0.22cm}
	\section{Related work and discussion}
	\label{section:related_work}
	
	
	\subsection{Label noise and semi-supervised learning}
	%
	The target modification algorithms are great strategies for model optimisation in the scenarios of label noise and semi-supervised learning, which are closely related.
	In the setting of semi-supervised learning, we are given partially annotated training data.  
	Therefore, its key is to reliably ``fill missing labels'' and continue to learn based on them. 
	Interestingly, when the missing labels are not perfectly filled, which is usually the case, the challenge of semi-supervised training changes to label noise.  
	For a further comparison, in the semi-supervised learning, the given annotations are generally clean and reliable, so that the label noise only exists in the unannotated set. 
	While in the setting of label noise, we are not given any lead about whether an example is trusted or not, thus being even more challenging.      

	\subsection{LC and knowledge distillation (KD)}
	%
	%
	In the section~\ref{subsec:theory on CCE, LS, CP and LC}, we have mathematically derived that some KD methods \cite{bucila2006model,hinton2015distilling,li2021mutual,xu2020knowledge,yuan2020revisiting} also modify labels. 
	Therefore, LC and KD are interchangeable in those cases.  
	We use the term LC other than KD mainly for two reasons: (1) LC is more descriptive; (2) the scope of KD becomes much larger than label modification.  
	For example, when two models are trained, the consistency between their predictions of a data point is rewarded in \cite{ba2014deep,zhang2018deep}, and a large distance between their feature maps is penalised in \cite{romero2015fitnets}. 
	Recently, multiple networks are trained for KD \cite{furlanello2018born}. 
	Regarding self KD, the intraclass samples are constrained to have consistent probability distributions \cite{xu2019data,yun2020regularizing}.  
	In another self KD \cite{zhang2019your}, the deepest classifier provides knowledge to supervise the shallower classifiers. 
	In a recent self KD method \cite{yuan2020revisiting}, Tf-KD$_{self}$ applies two-stage training. 
	%
	%
	In this work, we focus on improving the end-to-end self LC.  
	Therefore, some self KD methods \cite{xu2019data,yun2020regularizing,zhang2019your}, maximising the consistency of
	different classifiers or intraclass samples’ predictions, do not modify labels and are less relevant for comparison. 
	When it comes to the two-stage self LC method \cite{yuan2020revisiting}, in our view, it can be an add-on, i.e., an enhancement plugin. 
	%
	%
	Therefore, exploiting ProSelfLC to improve non-self and stage-wise LC approaches is an interesting area for future work.

	%
	%

	\begin{figure}[!t]
		\centering
		\begin{subfigure}[!h]{0.90\linewidth}
			\centering
			\captionsetup{width=\linewidth}
			\includegraphics[clip, trim=0.0cm 11.06cm 0.5cm 0.4cm, width=0.99\linewidth]{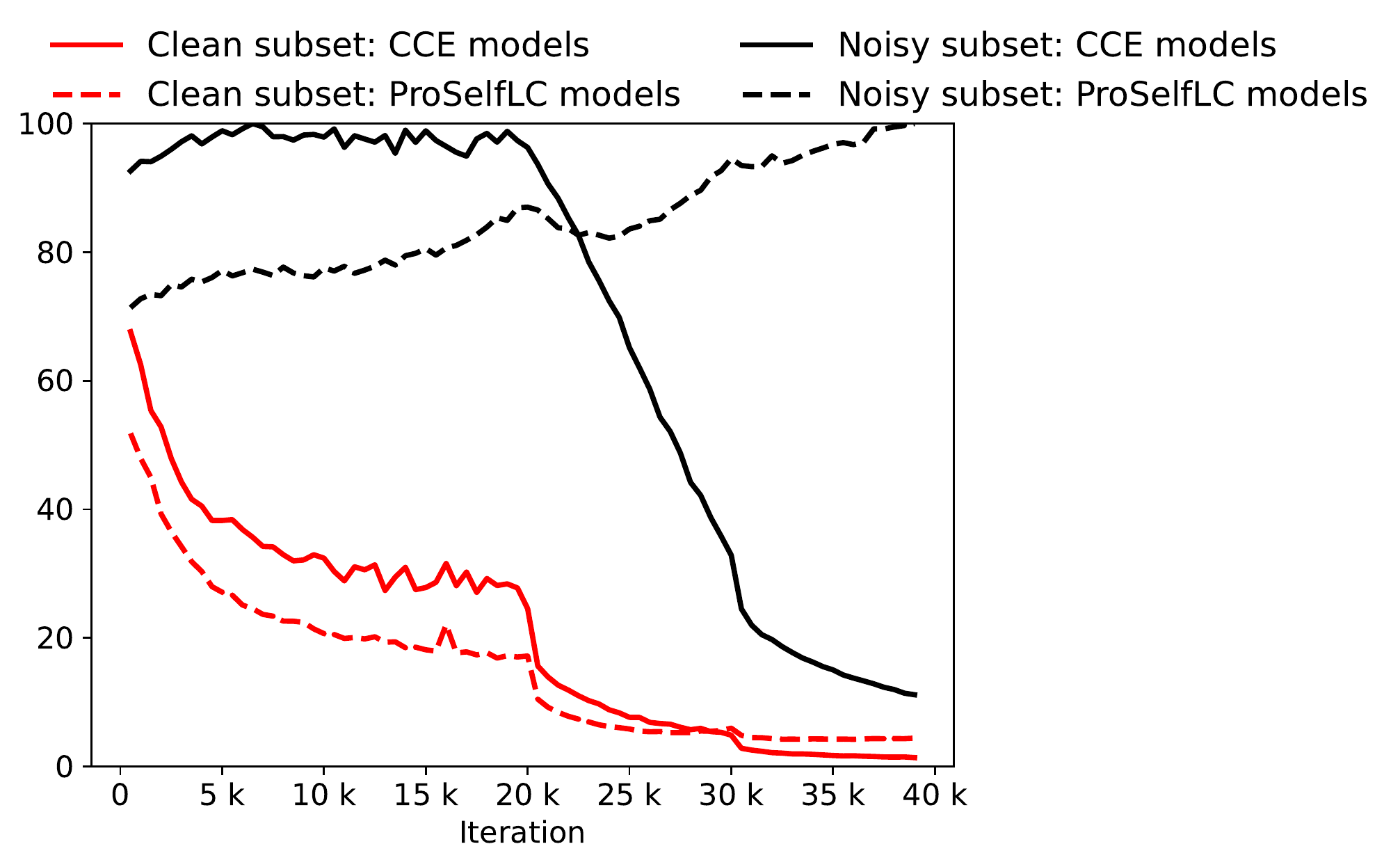}
		\end{subfigure}
		\vspace{0.15cm}
		
		\begin{subfigure}[!h]{0.513\linewidth}
			\centering
			\captionsetup{width=\linewidth}
			\includegraphics[clip, trim=0.34cm 0.32cm 6.3cm 1.64cm, width=0.99\linewidth]{multiloss_accuracy_loss/0.2_resnet18/all_subsets_all_losses.pdf}
			\caption{$20\%$ of training data is noisy.}
		\end{subfigure}
		\begin{subfigure}[!h]{0.476\linewidth}
			\centering
			\captionsetup{width=\linewidth}
			\includegraphics[clip, trim=1.326cm 0.32cm 6.3cm 1.64cm, width=0.99\linewidth]{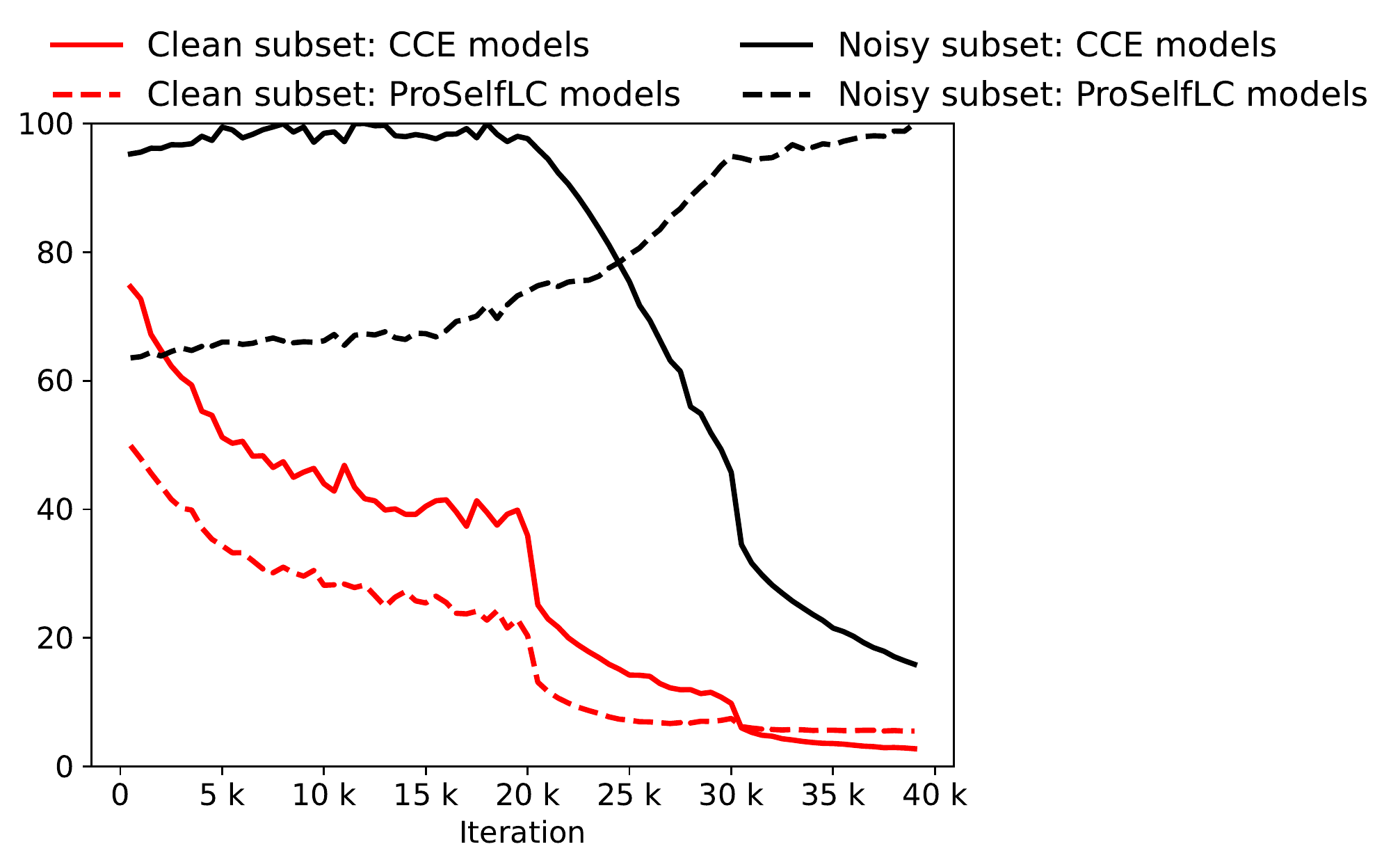}
			\caption{$40\%$ of training data is noisy.}	
		\end{subfigure}
		\vspace{-0.19cm}
		\caption{
			The change of cross entropy losses during training ResNet18 on CIFAR-100 with symmetric label noise. For a stratified analysis, we snapshot the model every 500 iterations and report the average losses of clean and noisy parts of the training data. 
			For plotting, we first divide a loss by the maximum loss at training then multiply it by 100.      
			For two noise rates, though the loss of the clean data decreases steadily for both CCE and ProSelfLC, \textit{the loss of the noisy data decreases only at a later phase when using CCE, while it increases throughout training when using ProSelfLC}. 			
		}
		\label{fig:dynamics_loss_resnet18}
	\end{figure}

	\subsection{Sample selection using the small-loss criterion}
	
	Recently, there is a popular family of algorithms which propose to learn from small-loss samples when severe label noise exists \cite{jiang2018mentornet, han2018co, yu2019does, han2020sigua,yao2020searching}. 
	Their underlying assumptions are that small-loss examples are clean and learning from clean data only mitigates fitting noise.   
	Generally, there are two key issues which significantly affect their performance in practice: (1) the selection schedule;  (2) the proportion of selected small-loss data.    	
	For example, to address the first issue, MentorNet \cite{jiang2018mentornet} learns a data selection curriculum while Co-teaching \cite{han2018co} gradually selects fewer clean samples as training proceeds. The given design reason of Co-teaching \cite{han2018co} is that a deep network starts to memorize the noisy data in the later training phase. 	
	To address the second issue, S2E \cite{yao2020searching} proposes an automated machine learning method to control the selection process so that a higher proportion of clean instances is selected and better performance is obtained. 
	
	To clearly understand why ProSelfLC outperforms the recent small-loss sample selection methods, as compared in Tables~\ref{table:cifar100_SOTA_Symmetric} and ~\ref{table:Clothing1M_competitors}, we display the change of cross entropy loss as training proceeds in Fig.~\ref{fig:dynamics_loss_resnet18}. We have the following insightful observations:  
	\begin{itemize}
		\item When using CCE, the loss of the noisy data decreases significantly in the later phase. This confirms the importance of the selection schedule of dropping very few samples at the early stage while leaving out more at the later stage.  If using CCE with small-loss data selection, the corrupted-label examples will not be selected to train the model. 
		\item When using ProSelfLC, the loss of the noisy data even increases steadily. 
		The model does not fit noisy labels and keeps improving knowledge at both clean and noisy training subsets, as also demonstrated in Fig.~\ref{fig:proselflc_dynamics_confidence_accuracy_resnet18}.  
	\end{itemize}
	
	In summary, first, ProSelfLC learns from all data while sample selection methods \cite{jiang2018mentornet,han2018co,yu2019does,yao2020searching,han2020sigua} only learn from small-loss data. Second, the proportion of selected data matters \cite{yao2020searching} and small-loss instances are more likely to be correct but not certain \cite{han2020sigua}. They are the reasons why ProSelfLC is superior.

	\section{Conclusion}
	
	Theoretically, we comprehensively study multiple label modification techniques from the viewpoints of entropy and KL divergence.  
	Methodologically, we propose ProSelfLC as an advanced self LC approach. 
	ProSelfLC is the first approach to trust low-temperature self knowledge progressively and adaptively.
	Extensive experiments prove its superiority over existing methods in clean and noisy scenarios of two diverse domains, i.e., image and protein datasets. 
	
	In terms of new insightful findings, we disclose and illustrate that deep neural networks become less confident of learning semantic patterns before fitting noise when the label noise rises, which complements the findings in \cite{zhang2017understanding,arpit2017closer,guo2017calibration,minderer2021revisiting}.  
	In addition, ProSelfLC promotes entropy minimisation, which is in marked contrast to the recent practices of confidence penalty \cite{szegedy2016rethinking,pereyra2017regularizing,dubey2018maximum}. The effectiveness of ProSelfLC defends the entropy minimisation principle. 
	
	%

	%
	
	
	

	
	%


	\ifCLASSOPTIONcompsoc
	\else
	\fi
	
	
	
	

	\ifCLASSOPTIONcaptionsoff
	\newpage
	\fi

	
	
	%
	%
	%
	
	{\footnotesize
		\bibliographystyle{ieee}
		\bibliography{ICLR2021_ProSelfLC}
	}
	
	%

	\vspace{-0.1cm}
	
	\begin{IEEEbiographynophoto}{Xinshao Wang} is a senior researcher of Zenith Ai and a visit scholar of University of Oxford. 
		He was a postdoctoral researcher at the Department of Engineering Science, University of Oxford after finishing his PhD at the Queens University of Belfast, UK.
		%
		%
		Xinshao Wang is working on core deep learning techniques with applications to visual recognition, disease prediction based on electronic health records, and protein engineering. Concretely,  
		he has been working on the following topics: (1) Deep metric learning: to learn discriminative and robust representations for downstream tasks, e.g., object retrieval and clustering;
		(2) Robust deep learning: robust learning and inference under adverse conditions, e.g., noisy labels, missing labels (semi-supervised learning), out-of-distribution training examples, sample imbalance, etc;
		(3) Computer vision: video/set-based person re-identification; image/video classification/retrieval/clustering; 
		(4) AI healthcare: electrocardiogram classification; 
		(5) ML-assisted gene and protein engineering.     
	\end{IEEEbiographynophoto}

	\vspace{-20pt}
	\begin{IEEEbiographynophoto}{Yang Hua}
		is presently a lecturer at the Queen’s University of Belfast, UK. He received his Ph.D. degree from Universit\'e Grenoble Alpes / Inria Grenoble Rhne-Alpes, France, funded by Microsoft Research Inria Joint Center. He won PASCAL Visual Object Classes (VOC) Challenge Classification Competition in 2010, 2011 and 2012, respectively and the Thermal Imagery Visual Object Tracking (VOTTIR) Competition in 2015. His research interests include machine learning methods for image and video understanding. He holds three US patents and one China patent.
	\end{IEEEbiographynophoto}
	
	
	\vspace{-20pt}
	\begin{IEEEbiographynophoto}{Elyor Kodirov}
		is a Lead Researcher at Zenith Ai and leads the PROEML team.
		He received his Ph.D. degree from Queen Mary University of London, UK.
		He was supervised by Tao Xiang. His research interests include machine
		learning, computer vision and NLP. Previously, he worked on computer
		vision with application to person re-identification/detection and face recognition.
		At present, his focuses on AI for bioinformatics and cheminformatics. He is working on challenging
		problems such as genetic expansion, protein engineering and molecular/protein representaion learning.
	\end{IEEEbiographynophoto}
	
	\vspace{-20pt}
	\begin{IEEEbiographynophoto}{Sankha Subhra Mukherjee} is a hands-on research leader who thinks deeply about the hardest problems in machine learning and delivers results on which innovative businesses have been created.
		Dr Mukherjee is a deep learning expert. His doctoral research at Heriot-Watt developed new deep neural network techniques which led to the co-founding of a high growth tech start-up, landmark publications, and patents. 
		He had been the founding EVP of Research and later CSO for a high growth start-up leading breakthroughs in machine learning, recruiting and supervising a team of 20 world-class researchers. In his current role he is a co-founder and Chief Scientific Officer, leading a team of world class researchers to deliver breakthroughs in ML and AI driven cell engineering and synthetic biology.
	\end{IEEEbiographynophoto}
	
	\begin{IEEEbiographynophoto}{David A. Clifton} 
		is a Professor of Clinical Machine Learning and leads the Computational Health Informatics (CHI) Lab. He is Official Fellow in AI \& ML at Reuben College, a Research Fellow of the Royal Academy of Engineering, and a Fellow of Fudan University, China. He studied Information Engineering at Oxford's Department of Engineering Science, supervised by Professor Lionel Tarassenko CBE. His research focuses on `AI for healthcare'.
		In 2018, the CHI Lab opened its second site, in Suzhou (China), with support from the Chinese government. In 2019, the Wellcome Trust's first "Flagship Centre" was announced, which joins CHI Lab to the Oxford University Clinical Research Unit in Vietnam, focused on AI for healthcare in resource-constrained settings.
		He is a Grand Challenge awardee from the UK Engineering and Physical Sciences Research Council, which is an EPSRC Fellowship that provides long-term strategic support for nine "future leaders in healthcare." He was joint winner of the inaugural "Vice-Chancellor's Innovation Prize", which identifies the best interdisciplinary research across the entirety of the University of Oxford.
	\end{IEEEbiographynophoto}

	\vspace{-20pt}
	\begin{IEEEbiographynophoto}{Neil M. Robertson}
		is CEO of Zenith Ai, and a Professor of Electrical Engineering and Computer Science at Queen’s University Belfast. He leads research into underpinning machine learning methods applied to diverse areas including synthetic biology, visual analytics and robotics. He started his career in the UK Scientific Civil Service (2000-2007), was the 1851 Royal Commission Fellow at Oxford University (2003–2006) in the Robotics Research Group and an academic at Edinburgh (2007-2016). His ML/AI research is extensive including UK major research programmes and a national doctoral training centre. He is the co/founder of three successful AI companies.
	\end{IEEEbiographynophoto}
	
	
	

	\newpage

	\onecolumn  
	\appendices

	\section{Proof of propositions}
	\label{appendix_sec:proof_of_propositions}

	\textbf{Proposition 4.} \textit{Compared with CCE, LS and CP penalise entropy minimisation while LC reward it.}
	%
	\\
	\textit{Proof.} We can rewrite CCE, LS, CP, and LC from the viewpoint of KL divergence: 
	%
	\begin{equation}
		\label{eq:KL_CE}
		\begin{aligned}
			\mathrm{L}_\mathrm{{CCE}}(\mathbf{{q}}, \mathbf{p}) = \mathrm{H}(\mathbf{q}, \mathbf{p}) 
			& =\mathrm{D}_\mathrm{KL}(\mathbf{q}||\mathbf{p}) + \mathrm{H}(\mathbf{q}, \mathbf{q}) 
			=\mathrm{D}_\mathrm{KL}(\mathbf{q}||\mathbf{p}), 
		\end{aligned}
	\end{equation}
	{where we have {$\mathrm{H}(\mathbf{q}, \mathbf{q}) = 0$ because $\mathbf{q}$ is a one-hot distribution}}. 
	\begin{equation}
		\label{eq:KL_LS}
		\begin{aligned}
			\mathrm{L}_\mathrm{{CCE+LS}}(\mathbf{{q}}, \mathbf{p}; \epsilon) 
			&
			= (1-\epsilon) \mathrm{D}_\mathrm{KL}(\mathbf{q}||\mathbf{p})
			+
			\epsilon \mathrm{D}_\mathrm{KL}(\mathbf{u}||\mathbf{p}) 
			+ \epsilon\mathrm{H}(\mathbf{u},\mathbf{u})
			\\
			&= (1-\epsilon) \mathrm{D}_\mathrm{KL}(\mathbf{q}||\mathbf{p})
			+
			\epsilon \mathrm{D}_\mathrm{KL}(\mathbf{u}||\mathbf{p}) 
			+ \epsilon \cdot \text{constant},
		\end{aligned}
	\end{equation}
	\begin{equation}
		\label{eq:KL_CP}
		\begin{aligned}
			\mathrm{L}_\mathrm{{CCE+CP}}(\mathbf{{q}}, \mathbf{p}; \epsilon) 
			&= (1-\epsilon) \mathrm{D}_\mathrm{KL}(\mathbf{q}||\mathbf{p})
			-
			\epsilon (\mathrm{H}(\mathbf{p},\mathbf{u}) 
			- \mathrm{D}_\mathrm{KL}(\mathbf{p}||\mathbf{u})
			)\\
			& = (1-\epsilon) \mathrm{D}_\mathrm{KL}(\mathbf{q}||\mathbf{p})
			+
			\epsilon \mathrm{D}_\mathrm{KL}(\mathbf{p}||\mathbf{u})
			- 
			\epsilon \cdot \text{constant},
		\end{aligned}
	\end{equation}
	where $\mathrm{H}(\mathbf{p},\mathbf{u})=\mathrm{H}(\mathbf{u},\mathbf{u}) = \text{constant}$.
	Analogously, 
	LC  in Eq~(\ref{eq:label_correction}) can also be rewritten: 
	\begin{equation}
		\label{eq:KL_LC}
		\begin{aligned}
			\mathrm{L}_\mathrm{{CCE+LC}}(\mathbf{{q}}, \mathbf{p}; \epsilon) 
			& = (1-\epsilon) \mathrm{D}_\mathrm{KL}(\mathbf{q}||\mathbf{p})
			-
			\epsilon \mathrm{D}_\mathrm{KL}(\mathbf{p}||\mathbf{u})
			+ 
			\epsilon \cdot \text{constant}.
		\end{aligned}
	\end{equation}
	In LS and CP, $+\mathrm{D}_\mathrm{KL}(\mathbf{u}||\mathbf{p})$ and $+\mathrm{D}_\mathrm{KL}(\mathbf{p}||\mathbf{u})$ pulls $\mathbf{p}$ towards $\mathbf{u}$. While in LC, the term   $-\mathrm{D}_\mathrm{KL}(\mathbf{p}||\mathbf{u})$ pushes $\mathbf{p}$ away from $\mathbf{u}$.   \hfill\(\Box\)\\

	\noindent
	\textbf{Proposition 5.} \textit{In CCE, LS and CP, a data point $\mathbf{x}$ has the same semantic class. In addition, $\mathbf{x}$ has an identical probability of belonging to other classes except for its semantic class.
	}
	\\
	\textit{Proof.}  
	In LS, the target is $ \mathbf{\tilde{q}_{\mathrm{LS}}}=(1-\epsilon)\mathbf{q}+\epsilon \mathbf{u}$. 
	For any $0 \leq \epsilon < 1$, the semantic class is not changed, because $1-\epsilon + \epsilon * \frac{1}{c} > \epsilon * \frac{1}{c}$. 
	In addition, $j_1 \neq y, j_2 \neq y \Rightarrow \mathbf{\tilde{q}_{\mathrm{LS}}}(j_1|\mathbf{x}) = \mathbf{\tilde{q}_{\mathrm{LS}}}(j_2|\mathbf{x})=\frac{\epsilon}{c}$.
	
	In CP, $ \mathbf{\tilde{q}_{\mathrm{CP}}}=(1-\epsilon)\mathbf{q}-\epsilon \mathbf{p}$.
	In terms of label definition, \textit{CP is against intuition because these zero-value positions in $\mathbf{q}$ are filled with negative values in $\mathbf{\tilde{q}_{\mathrm{CP}}}$.}  
	A probability has to be not smaller than zero. So we rephrase $\mathbf{\tilde{q}_{\mathrm{CP}}}(y|\mathbf{x})=(1-\epsilon)-\epsilon*\mathbf{p}(y|\mathbf{x})$, and $\forall j\neq y, \mathbf{\tilde{q}_{\mathrm{CP}}}(j|\mathbf{x})=0$ by replacing negative values with zeros, as illustrated in Fig. \ref{fig:LS_CP}.     \hfill\(\Box\)

	\section{Discussions on wrongly confident predictions and model calibration}
	\label{appendix:error_and_model_calibration}
	
	1. \textit{It is likely that some highly confident predictions are wrong. Will ProSelfLC suffer from an amplification of those errors? }
	
	First of all, ProSelfLC alleviates this issue a lot and makes a model confident in correct predictions, according to Fig.~\ref{fig:CIFAR100Asy0_4_mean_entropy_NoisySubset} together with \ref{fig:CIFAR100Asy0_4_corruptedFittting} and \ref{fig:CIFAR100Asy0_4_semanticCorrection}. 
	{{Fig.~\ref{fig:CIFAR100Asy0_4_mean_entropy_NoisySubset} shows the confidence of predictions, whose majority are correct according to Fig.~\ref{fig:CIFAR100Asy0_4_corruptedFittting} and \ref{fig:CIFAR100Asy0_4_semanticCorrection}.}} 
	In Fig.~\ref{fig:CIFAR100Asy0_4_corruptedFittting}, ProSelfLC fits noisy labels least, i.e., around 12\% so that the correction rate of noisy labels is about 88\% in Fig.~\ref{fig:CIFAR100Asy0_4_semanticCorrection}.
	Nonetheless, ProSelfLC is non-perfect. A few noisy labels are memorised with high confidence. \\

	\noindent
	2. \textit{How about the results of model calibration using a computational evaluation metric: Expected Calibration Error (ECE) \cite{naeini2015obtaining,guo2017calibration}?}
	
	Following the practice of \cite{guo2017calibration}, on the CIFAR-100 test set, we report the ECE (\%, \#bins=10)  of 
	ProSelfLC versus CCE, as a complement of Fig.~\ref{fig:comprehensive_dynamics}. 
	For a comparison, CCE's results are shown in corresponding brackets.   
	We try several confidence metrics (CMs), including probability, entropy, and their temperature-scaled variants using a parameter $T$.
	Though the ECE metric is sensitive to CM and $T$, ProSelfLC's ECEs are smaller than CCE's. 
	
	\begin{table*}[!h]
		\centering
		\caption{
			ECE results of multiple combinations of logits scaling (logits$/ T $) and confidence metrics (probability and entropy). 
		}
		\setlength{\tabcolsep}{9.9pt} 
		\vspace{-0.35cm}
		\begin{tabular}{lccccc}
			\midrule
			\makecell{Scaling logits with a temperature parameter $T$:  \\logits$/ T $} & $T =1$ & $T =1/4$ & $T =1/8$ \\
			
			\midrule
			CM =  $\mathrm{conf}_{\mathrm{top}}$& 
			15.71 (40.98) & 4.24 (18.27) & \textbf{2.39} (9.94)
			\\
			
			CM = 
			$\mathrm{conf}_{\mathrm{all}}$&
			17.38 (42.83) & 5.22 (17.84) & \textbf{2.66} (9.53)
			\\
			\midrule
		\end{tabular}
		\label{table:ddd}
	\end{table*}

	\section{The changes of entropy statistics and $\epsilon_{\mathrm{ProSelfLC}}$  at training}
	\label{appendix:entropy_and_coef_dynamics}
	
\begin{figure*}[!h]
		\centering
		\begin{subfigure}{0.49\linewidth}
			\captionsetup{width=0.9\textwidth}
			\includegraphics[clip, trim=1.24cm 7.5cm 1.7cm 5.9cm, width=0.98\textwidth]{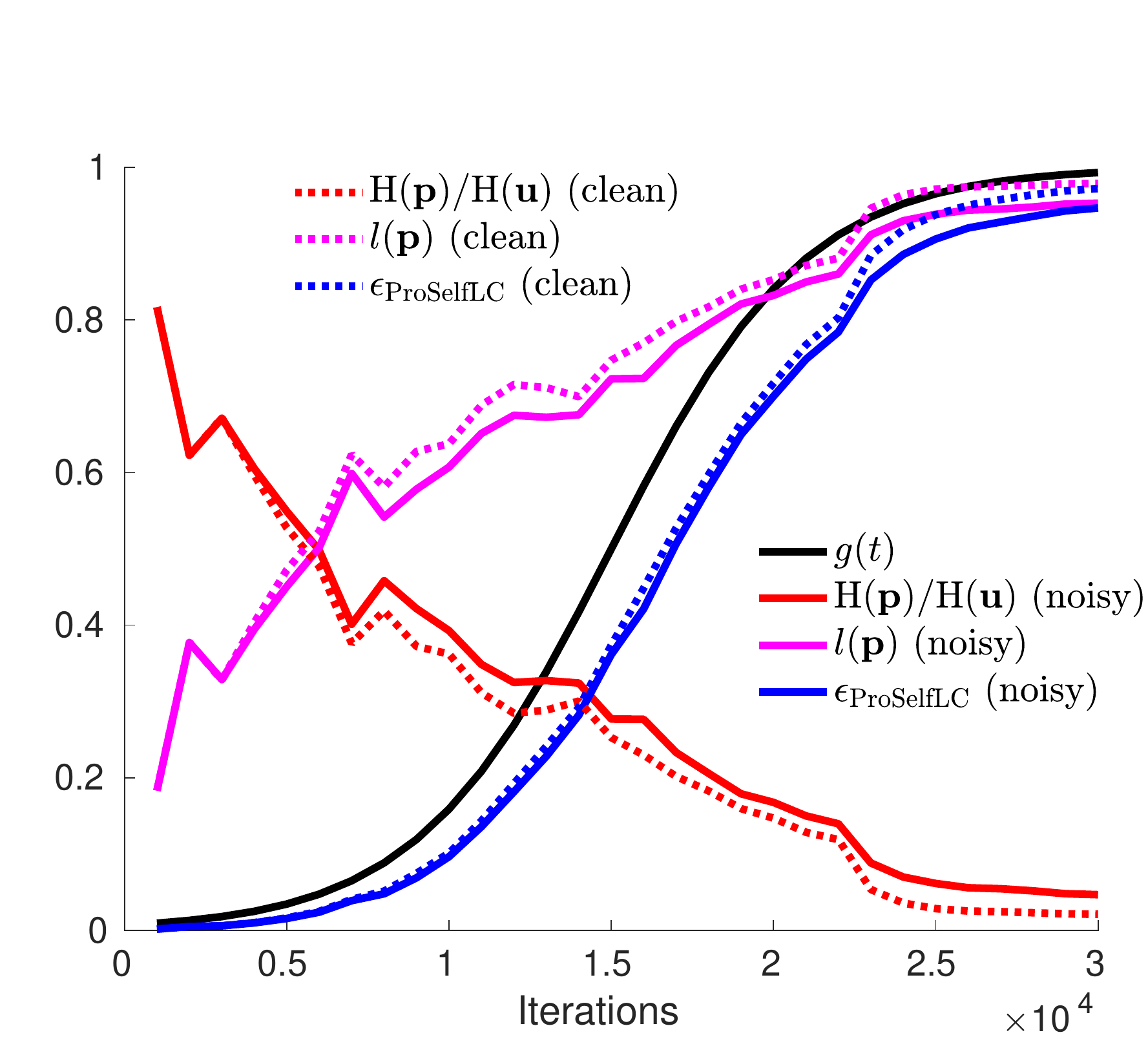}
			\caption{ Asymmetric label noise rate = 20\%.  }
			\label{fig:NoisyCleanSubsets_lcs10_0_2_entropy_episolon}
		\end{subfigure}
		\begin{subfigure}{0.49\linewidth}
			\captionsetup{width=0.9\textwidth}
			\includegraphics[clip, trim=1.24cm 7.5cm 2.3cm 5.9cm, width=0.98\textwidth]{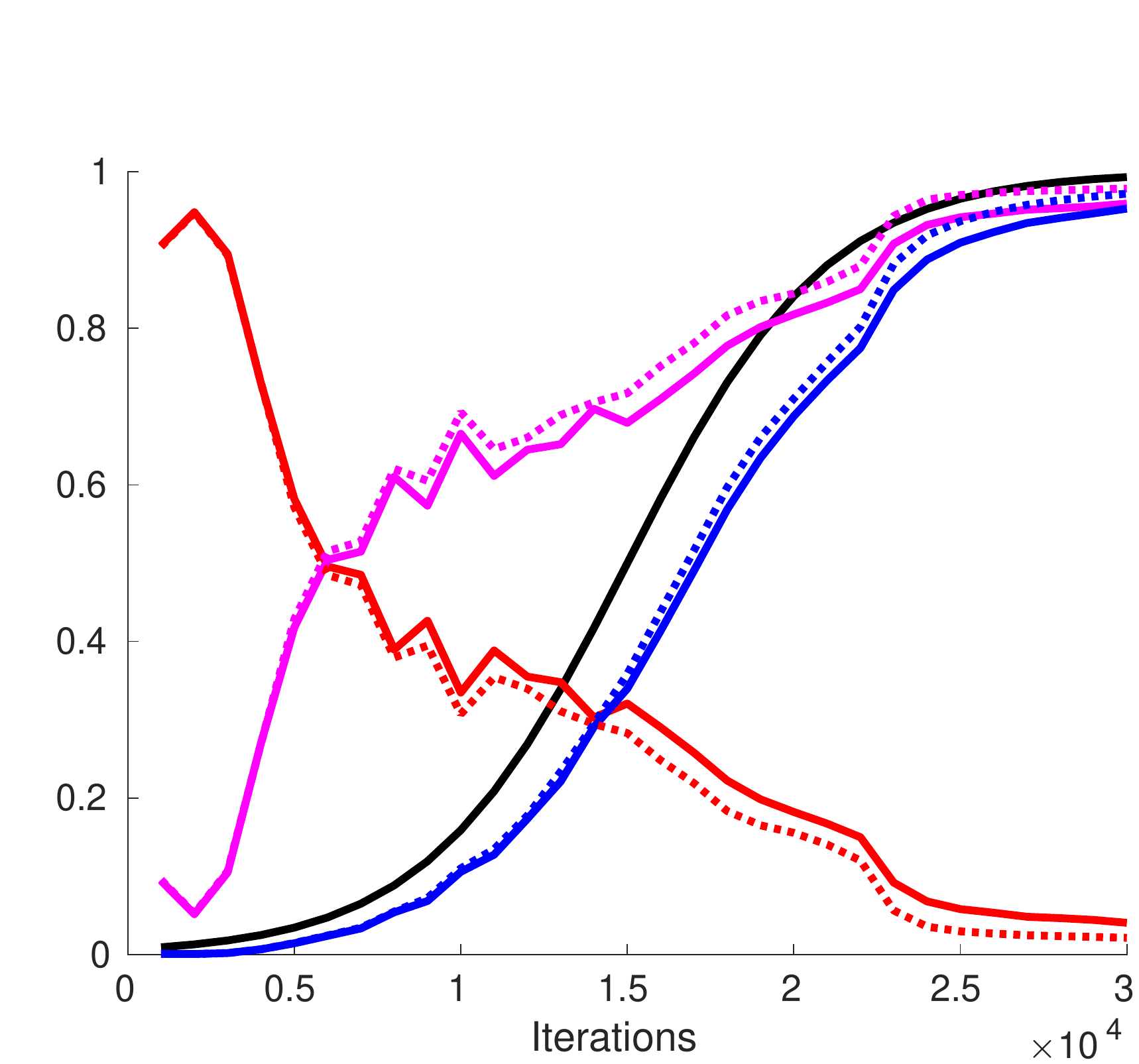}
			\caption{ Asymmetric label noise rate = 40\%. }
			\label{fig:NoisyCleanSubsets_lcs10_0_6_entropy_episolon}
		\end{subfigure}
		\caption{The changes of entropy statistics and $\epsilon_{\mathrm{ProSelfLC}}$  at training.
			We store a model every 1000 iterations to monitor the learning process. 
			For data-dependent metrics, after training, we split the corrupted training data into clean and noisy subsets according to the information about how the training data is corrupted before training.  
			Finally, we report the mean results of each subset.  
		}
		\label{fig:entropy_episoln_dynamics}
	\end{figure*}

	In Fig.~\ref{fig:entropy_episoln_dynamics}, we visualise how the entropies of noisy and clean subsets change at training. 
	
\end{document}